\documentclass{article}

\usepackage[english]{babel}
\usepackage[utf8]{inputenc}
\usepackage{johd}
\usepackage{pifont}
\usepackage{xspace}
\usepackage{graphicx}
\usepackage{multirow}
\usepackage[dvipsnames]{xcolor}
\usepackage{booktabs} 
 \usepackage{amsmath}
 \usepackage{float}
\usepackage{listings}
\usepackage[ruled,vlined]{algorithm2e}
\usepackage{subcaption}

\usepackage{algpseudocode}
\usepackage{amsmath}
\usepackage{doi}
\usepackage{tcolorbox}
\definecolor{violet}{RGB}{120,94,240}
\definecolor{orange}{RGB}{254,97,0}
\definecolor{pink}{RGB}{220,38,127}
\definecolor{myblue}{RGB}{136,34,85}

\SetKwInOut{Input}{\texttt{Input}}
\SetKwInOut{Output}{\texttt{Output}}
\SetKw{KwTo}{\texttt{to}}
\SetKw{KwReturn}{\texttt{return}}
\SetKwFor{For}{\texttt{for}}{\texttt{do}}{\texttt{end for}}
\SetKwIF{If}{ElseIf}{Else}{\texttt{if}}{\texttt{then}}{\texttt{else if}}{\texttt{else}}{\texttt{end if}}

\newcommand{\BPESupervised}{\texttt{\textbf{\textcolor{pink}{Token Supervised }}}}
\newcommand{\BPEDistilled}{\texttt{\textbf{\textcolor{pink}{Token Distilled }}}}
\newcommand{\BPETransformer}{\texttt{\textbf{\textcolor{pink}{Token-1B }}}}
\newcommand{\token}{\texttt{\textbf{\textcolor{pink}{Token }}}}
\newcommand{\tokens}{\texttt{\textbf{\textcolor{pink}{Tokens }}}}
\newcommand{\bpe}{\texttt{\textbf{\textcolor{pink}{Token }}}}

\newcommand{\BytesSupervised}{\texttt{\textbf{\textcolor{orange}{Bytes Supervised }}}}
\newcommand{\byte}{\texttt{\textbf{\textcolor{orange}{Byte }}}}
\newcommand{\MarginalizeDistilled}{\texttt{\textbf{\textcolor{orange}{Marginalize-It Distilled }}}}

\newcommand{\MarginalizeIt}{\texttt{\textbf{\textcolor{orange}{Marginalize-It }}}}
\newcommand{\ByteTransformer}{\texttt{\textbf{\textcolor{orange}{Bytes-1B }}}}
\newcommand{\bytes}{\texttt{\textbf{\textcolor{orange}{Bytes }}}}

\newcommand{\BytesEotSupervised}{\texttt{\textbf{\textcolor{violet}{Bytes w/ <eot> Supervised }}}}
\newcommand{\EndOfTokenDistilled}{\texttt{\textbf{\textcolor{violet}{End-Of-Token Distilled }}}}

\newcommand{\EndOfToken}{\texttt{\textbf{\textcolor{violet}{End-Of-Token }}}}
\newcommand{\EndOfTokenTransformer}{\texttt{\textbf{\textcolor{violet}{End-Of-Token-1B }}}}
\newcommand{\bytesweot}{\texttt{\textbf{\textcolor{violet}{Bytes w/ <eot> }}}}

\newcommand{\twentyfivepercent}{\texttt{\textbf{\textcolor{myblue}{ $\approx$ 30.94\% }}}}

\newcommand{\DownstreamEquation}{
$z(y) = \epsilon - k \cdot e^{-\gamma y}$ }

\usepackage{fontspec}
\definecolor{hypothesisblue}{RGB}{100,143,255}

\newcommand{\FeatherPlot}{\texttt{\textbf{\textcolor[RGB]{ 2, 87, 224}{Feather Plots }}}}

\newcommand{\Featherplot}{\texttt{\textbf{\textcolor[RGB]{ 2, 87, 224}{Feather Plot }}}}

\usepackage{eso-pic}
\usepackage{tikz}
\AddToShipoutPictureBG*{%
  \begin{tikzpicture}[remember picture,overlay]
    \node[opacity=.9,inner sep=0pt,rotate=360] at ([yshift=-2.3cm]current page.north)
      {\includegraphics[width=1.65\paperwidth]{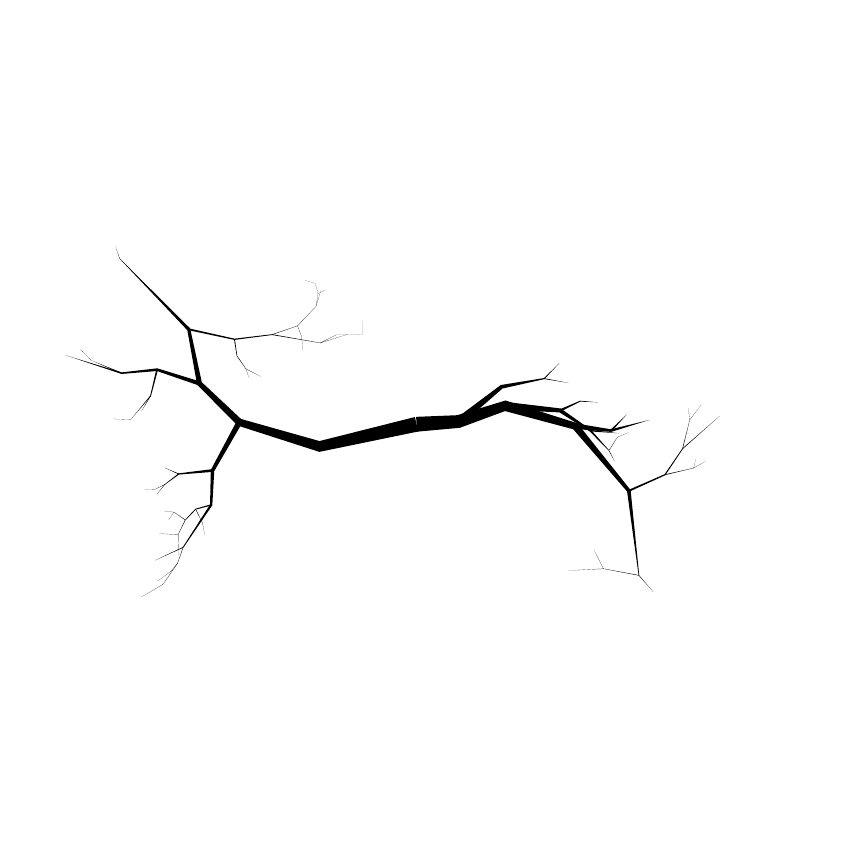}};
  \end{tikzpicture}}

\title{ {Breaking the Token Ceiling}: \\ Distilling Smaller, Stronger Byte Models }

\newcommand{\method}[0]{\texttt{Byte Logits}\xspace}

\usepackage[dvipsnames]{xcolor}

\definecolor{metablue}{RGB}{4, 87, 203}
\definecolor{uwpurple}{RGB}{75, 46, 131}   

\newcommand{\uwm}[1]{\textcolor{uwpurple}{$^{#1}$}}
\newcommand{\metam}[1]{\textcolor{metablue}{$^{#1}$}}

\author{Kalyani Marathe\uwm{1}\metam{\ddagger}$^{\dagger}$$^{*}$,
        Artidoro Pagnoni\metam{2}$^{\dagger}$$^{*}$,
        Tomasz Limisiewicz\uwm{1}\metam{2},
        Margaret Li\uwm{1}, \\
        Mike Lewis\metam{2}$^{\#}$,
        Luke Zettlemoyer\uwm{1}\metam{2}$^{\#}$,
        Srinivasan Iyer\metam{2}$^{\dagger}$$^{\#}$$^{*}$ \\
        \small \textcolor{uwpurple}{$^{1}$University of Washington, Seattle} \\
        \small \textcolor{metablue}{$^{\ddagger}$Work done at Meta FAIR} \\
        \small \textcolor{metablue}{$^{2}$Meta FAIR} \\
        \small $^{\dagger}$Joint first author. $^{\#}$Joint last author.\\
}

\date{} 

\begin{document}

\maketitle
\maketitle

\begingroup
\renewcommand{\thefootnote}{}
\footnotetext{$^{*}$Correspondence: \textcolor{uwpurple}{\texttt{kmarathe@uw.edu}}, \textcolor{metablue}{\texttt{\{artidoro, sviyer\}@meta.com}}}
\addtocounter{footnote}{-1}
\endgroup

\begin{abstract}

\noindent Small models are made more capable through distillation from a larger one that shares their tokenization scheme. However, do distilled byte and token models behave similarly in terms of scaling trends as compute and data increases? To enable this comparison, we introduce two variants to efficiently convert token logits to \texttt{Byte Logits}: 1) approximate: \MarginalizeIt, and 2) exact: \EndOfToken.We then present the first large scale study of overtraining decoder-only dense transformer models varying two dimensions simultaneously: the \texttt{tokenization scheme} ( \tokens, \bytes, \bytesweot) and the \texttt{training objective} (Distillation vs. Cross-Entropy), sweeping layer-parameter-matched models with $\approx$ 1 billion parameters up to 1 trillion bytes of data. Across eight benchmarks spanning three categories: Multiple Choice QA, Language Generation, and Machine Translation, we find that \BPETransformer models outperform byte models ( \EndOfTokenTransformer and \ByteTransformer) in the low-FLOP regime but eventually plateau; byte models start worse yet surpass \BPETransformer models with more compute, reaching a higher downstream task performance ceiling. Extrapolating the average top-1 error vs. validation BPB scaling laws predicts that, asymptotically, distilled \EndOfTokenTransformer outperforms distilled \BPETransformer by up to \texttt{\textcolor{myblue}{4\%}} on averaged downstream task performance. They are also far more data efficient, matching the performance of distilled \BPETransformer using only \texttt{\textcolor{myblue}{one-sixth}} of the training data. Moreover, by operating over a small vocabulary of $\approx 256$ bytes instead of $\approx 100$K tokens, they circumvent the need for top-$k$ truncation during logit dumping, while also reducing logit storage costs to roughly \texttt{\textcolor{myblue}{one-fifth}}. Finally, our downstream performance scaling laws predict that our distilled \EndOfTokenTransformer models asymptotically surpass the \texttt{Llama 3.2-1B} model \citep{meta2024llama32}, \texttt{Gemma-3-1B-pt}   \citep{Kamath2025Gemma3T}, and \texttt{Gemma 2B} \citep{team2024gemma} models on averaged downstream tasks by up to \texttt{\textcolor{myblue}{6.5\%, 8.1\%}}, and \texttt{\textcolor{myblue}{2.1\%}}, respectively.

\end{abstract}

\section{Introduction}
Scaling has become the prevailing approach for improving language model performance \citep{kaplan2020scaling, hoffmann2022training, henighan2020scaling}. However, deploying models in real-world settings demands more than scale alone: the cost of serving can outweigh the benefits of scale \citep{pope2023efficiently, dettmers2023case}. Models are often 1) distilled into smaller yet capable models that, by training against the teacher's next-token distribution rather than ground-truth tokens, outperform on downstream tasks \citep{hinton2015distilling,beyer2022knowledge}, and 2) overtrained to amortize inference costs over a model's lifetime \citep{sardana2023beyond,   gadre2024language}. Yet it remains unclear how the choice of student tokenization interacts with these techniques as compute scales. \\

\noindent In this work we compare how bytes and tokens affect distillation scaling trends as the compute budget grows. Bytes are a compelling alternative for two reasons:   First, by compressing the vocabulary to just $\approx$ 256 values, transformers operating on bytes can preserve the distribution of a large token language model exactly \citep{hayase2025sampling, phan2024exact} and potentially can bypass the need for top-k logit truncation for offline distillation. Second, they also possess interesting properties: they learn relationships between bytes from fewer samples than token models require for tokens \citep{hestness2017deep}, which makes them  data efficient learners \citep{xue2022byt5, evabyte}. Distillation into such models, though, hinges on open questions: 1) how to efficiently convert token logits to byte logits, 2) how does tokenization and training objectives affect the BPB vs FLOPs power laws 3) whether distillation outperforms supervised byte (and token) level training on downstream tasks, 4) at what compute budget byte-level surpasses token-level distillation, 5) how the trained models and the asymptotic predictions compare to the open-weights model of the comparable model sizes and what are the data and storage efficiency implications. This work addresses all five. We summarize our key contributions in section \ref{summary} below: \\ 

  \begin{figure}[H]
    \centering
    \includegraphics[width=1.0\textwidth]{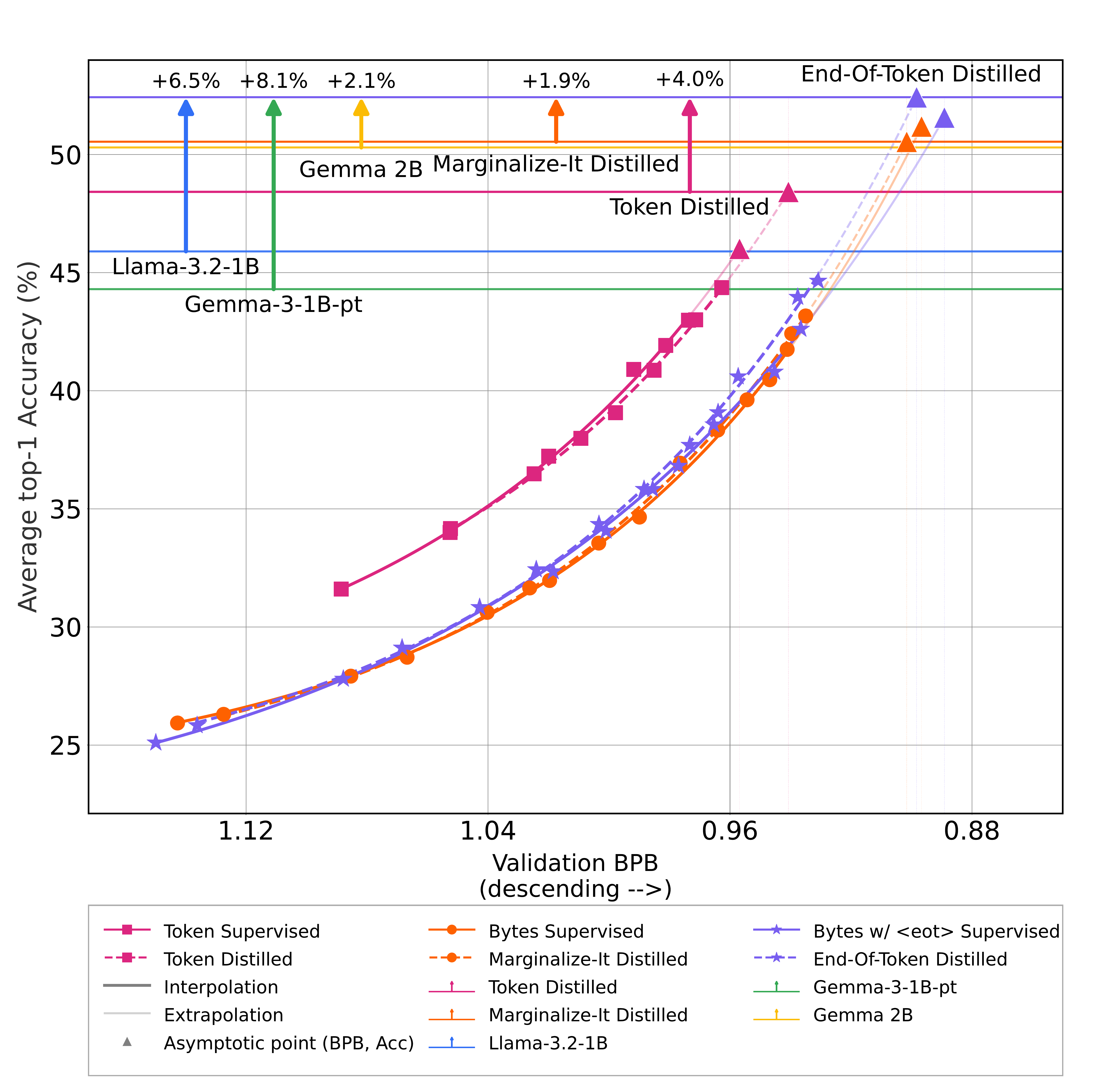}
    \caption{ Downstream task performance scaling laws for lr=4e-3 averaged across six benchmarks: Multiple Choice QA(ARC-Easy, ARC-Challenge, HellaSwag, PIQA) and Language Generation (MBPP, Natural Questions) tasks. \EndOfTokenDistilled model, by preserving the teacher distribution exactly as well as with the highest amount of compute spent on fixed amount of data raises the asymptotic performance ceiling. \EndOfTokenDistilled asymptotically outperforms the \BPEDistilled model by \texttt{\textbf{\textcolor{myblue}{4\%}}}, the \MarginalizeDistilled model by \texttt{\textbf{\textcolor{myblue}{1.9\%}}}, and the three open weights models: Llama 3.2-1B model \citep{meta2024llama32}, Gemma-3-1B-pt   \citep{Kamath2025Gemma3T}, and Gemma 2B \citep{team2024gemma} models on averaged downstream tasks by upto \textbf{\texttt{\textcolor{myblue}{6.5\%, 8.1\%, 2.1\%}}} respectively.  }
    \label{fig:downstream_scaling_law_fitting_notes_4e3}

\end{figure}

\subsection{Summary of Contributions} \label{summary}       
    \noindent \textbf{\textcolor{black}{ Token logits to Byte logits conversion in a single forward pass.}} Distilling a smaller byte model from a large token model first requires converting the teacher's token logits onto byte logits. A token comprises several bytes, so the probability distribution  must be split across byte positions. Prior methods \citep{hayase2025sampling, phan2024exact} to convert token distribution to byte distribution require multiple forward passes which is prohibitive when teacher inference is computationally expensive.   We introduce two single-pass variants for converting token logits to \texttt{Byte Logits} by marginalizing over byte positions: 1) approximate: \MarginalizeIt{} re-normalizes over the prefix-matching tokens, while 2) exact: \EndOfToken{}adds an \texttt{<eot>} token to the vocabulary so no probability is lost.  \\

     \noindent \textbf{\textcolor{black}{Bits-Per-Byte vs Training FLOPs Power Laws Across Tokenization and Training Objectives.}}   With a logit conversion in hand, we ask how the resulting byte models scale compared to token models and their supervised counterparts. We present a large-scale study of overtraining layer-parameter matched decoder-only dense transformers along two axes: the training objective (Distillation loss vs Supervised/Cross-Entropy loss) and the tokenization scheme ( \tokens, \bytes, \bytesweot): sweeping  decoder-only transformer models \citep{vaswani2017attention} with fixed 1.28 billion layer parameters up to 1 trillion bytes data.  We train multiple \BPETransformer (1.81B), \ByteTransformer (1.28B), and \EndOfTokenTransformer (1.28B) models to fit Validation BPB vs Training FLOPs power laws to show that distilled \EndOfTokenTransformer models show better asymptotic behavior compared to distilled \BPETransformer and \ByteTransformer. \\     

     \noindent \textbf{\textcolor{black}{ Average Downstream Task Accuracy vs Training FLOPs Scaling Trends.}} To understand how the scaling trends in validation loss  translate into downstream task performance for \MarginalizeIt (approximate) and \EndOfToken (exact) we now turn to benchmark evaluations. Across three categories and eight benchmarks: Multiple Choice QA (ARC-Easy, ARC-Challenge, HellaSwag, PIQA), Language Generation (MBPP, Natural Questions), and Machine Translation (Flores) we find that \BPETransformer transformers (supervised and distilled) with  show strong performance at low compute regimes but plateau quickly. \ByteTransformer(1.28B) and \EndOfTokenTransformer(1.28B), lag initially behind the \BPETransformer(1.81B) models, continue to improve at a steeper rate with additional compute despite fewer total parameters due to the small vocabulary size of $\approx$ 256 bytes. \\
       
\begin{figure}[H]
    \centering
    \includegraphics[width=1.0\textwidth]{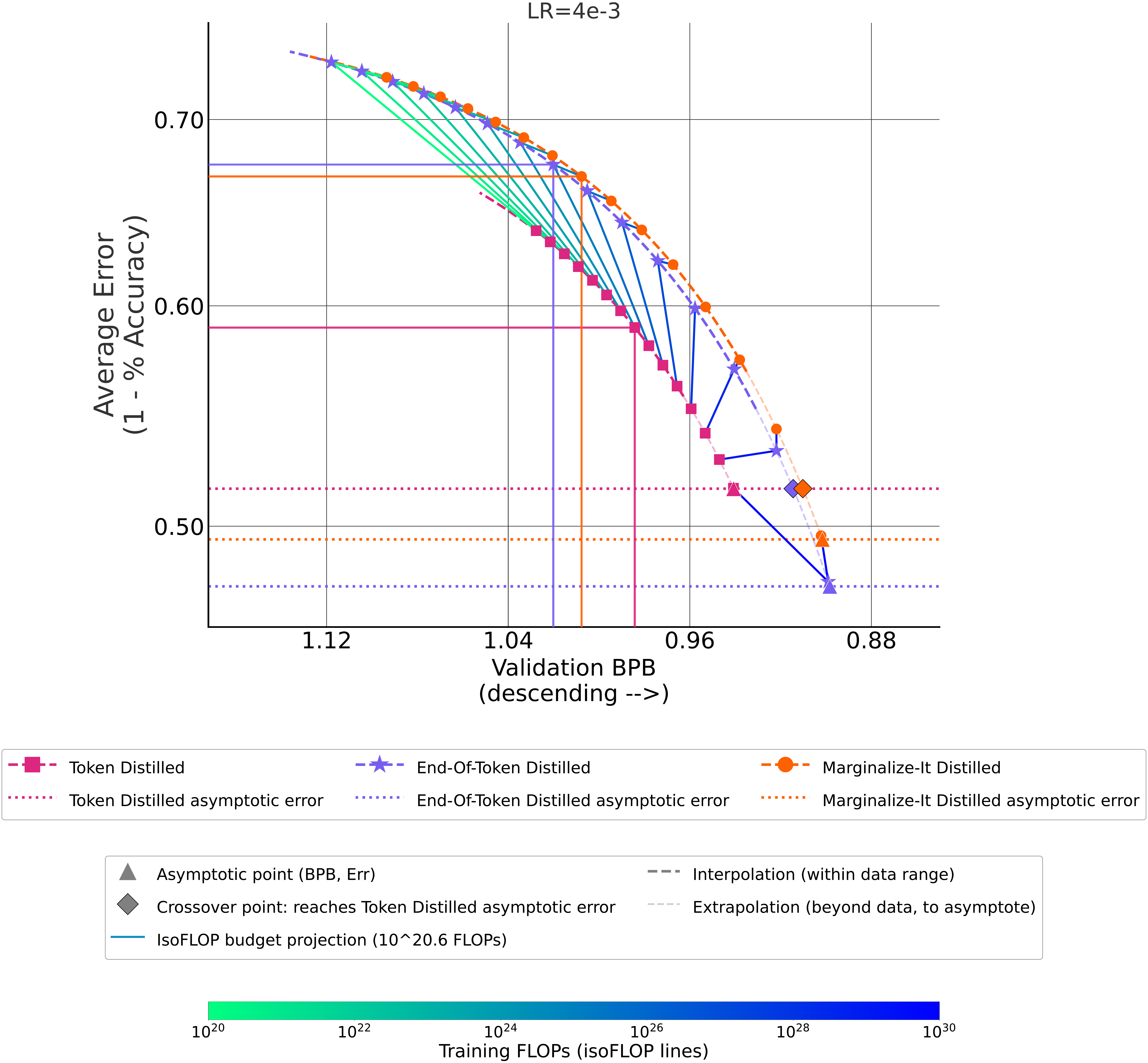}
    \caption{ The figure shows a \Featherplot for a learning rate of 4e-3. We show that \EndOfTokenDistilled transformer model has the lowest Average top-1 downstream task performance error asymptotically compared to the \MarginalizeIt and \token distillation. We also visualize the Validation BPB values and the Average top-1 downstream task error at the compute budget of $10^{20.63}$ FLOPs (a randomly sampled value). We note the differences in observed BPB value and the task performance by projecting iso-FLOP points on the X and Y axes for the three curves. At that compute budget, \token distillation leads significantly, while both \EndOfToken distillation and \MarginalizeIt distillation lag behind.  }
    \label{fig:feather_4e3}

\end{figure}

     \noindent \textbf{\textcolor{black}{Comparing Distillation Scaling Trajectories Across Compute Budgets.}} Downstream task performance vs training FLOPs scaling trend raises the practical question of which method to choose at a given compute budget. We fit Downstream error vs. Validation BPB scaling laws (Figure \ref{fig:downstream_scaling_law_fitting_notes_4e3}) and introduce \FeatherPlot (Figure \ref{fig:feather_4e3}), which connect  IsoFLOP points on the downstream-error vs. Validation-BPB curves. These plots let us compare downstream performance trajectories across compute budgets for each of the tokenization scheme. By extrapolating the downstream error vs validation BPB scaling laws, our experiments predict that distilled \EndOfTokenTransformer transformers raise the asymptotic ceiling of the averaged downstream task performance. 
        We then predict that by both preserving the teacher distribution exactly as well as by spending  \twentyfivepercent more compute on a fixed amount of data compared to \MarginalizeIt (approximate) method, distilled \EndOfTokenTransformer (exact) improves over distilled \BPETransformer by up to \textbf{\texttt{\textcolor{myblue}{4\%}}} and distilled \ByteTransformer by up to \textbf{\texttt{\textcolor{myblue}{1.9\%}}}  asymptotically. \\
       
       \noindent \textbf{\textcolor{black}{\EndOfTokenTransformer: Data and Storage Efficiency, Asymptotic win over Open-Weight models. }}The asymptotic accuracy gains from \EndOfToken method also come with efficiency gains. Distilled \EndOfTokenTransformer model also surpass distilled \BPETransformer models with up to  \textcolor{myblue}{\texttt{one-sixth}} of the training data with sufficient training despite having fewer total parameters. 
        By operating over a small vocabulary of $\approx 256$ bytes instead of the $\approx 100$K tokens they circumvent the need for top-k truncation during
logit dumping,
        reducing logit storage costs to roughly \textcolor{myblue}{\texttt{one-fifth}}. Finally, comparisons with the open-weight models with the asymptotic predictions reveal our distilled \EndOfTokenTransformer model surpass the \cite{meta2024llama32}, \cite{Kamath2025Gemma3T}, and \cite{team2024gemma} models on averaged downstream tasks by upto \textbf{\texttt{\textcolor{myblue}{6.5\% and 8.1\%, 2.1\%}}} respectively. \\

\noindent \textbf{Paper Outline}:  We introduce two logit conversion methods: \MarginalizeIt and \EndOfToken to convert the token logits to byte logits (Section \ref{algorithms}). We then describe the experimental setup, the six scenarios used for our scaling study, and the benchmarks used for evaluations (Section \ref{design}). We then study the validation BPB vs FLOPs scaling laws (Section \ref{bpbvsflops}), the downstream task performance vs training FLOP scaling trends (Section \ref{bench_vs_flops}) and the downstream performance vs validation BPB scaling laws (Section \ref{perf_vs_bpb}). We then use these scaling laws to find out when byte distillation will be  advantageous over token distillation baseline (Section \ref{storage_case}). Through the lens of downstream error scaling laws, with the help of \FeatherPlot  we demonstrate how the relative performance ordering between the \BPEDistilled, \MarginalizeDistilled, and \EndOfTokenDistilled models evolve with the training compute budget.
We then benchmark our trained checkpoints and our asymptotic predictions against the open weight models (Section \ref{llama_gemma}). 
Next, we  contextualize our work within the existing literature (Section \ref{relatedwork}). Lastly we reflect on the findings, limitations and opportunities for future work (Section \ref{discussion}).

\section{ Logit Conversion Methods } \label{algorithms}

In this section we describe two methods: \MarginalizeIt and \EndOfToken, we use in our scaling study to efficiently convert token logits to \method. Figure \ref{algorithms} demonstrate the byte logits formation process with the help of an example. We provide the pseudocode for the two methods in Appendix \ref{dist_pseudocode} and discuss special cases in Appendix \ref{corner_cases}.

\subsection{Marginalize-It Logits}
\label{marginalizeit}

For each token used in teacher inference, we have a $|V|$-sized logit tensor
denoting the next-token distribution, where $V$ is the vocabulary of the BPE
tokenizer. We also have the decoded byte sequence of every BPE token in $V$.
To compute the \emph{first} byte distribution of each BPE token, we marginalize
over the full vocabulary $V$ based on the first byte. For each \emph{subsequent}
byte, we restrict the vocabulary to the tokens whose prefixes match the
ground-truth bytes and compute the byte distribution conditioned on this subset.
We refer to this as the \MarginalizeIt{} method. While this approach always yields an exact distribution for the first byte, the
distributions over subsequent bytes are only \emph{approximate}. Figure~\ref{fig:bpe_to_byte_figure} (column 1) shows the \MarginalizeIt logit conversion process step by step.
To predict the distribution \texttt{B3}, the continuations of token \texttt{is} (whose probability is 0.125) are silently dropped and only the tokens \texttt{isu} and \texttt{isk} are carried forward. While obtaining the next token distribution of token \texttt{is} can fix this problem, it requires running additional teacher inference over the sequence: \texttt{[<bos>, T, iram, is]}. The number of additional inferences required to obtain exact conversions can get intractable for long sequences. To avoid that problem, in \MarginalizeIt method 
remaining probability is redistributed among tokens \texttt{isu} and \texttt{isk}:
\begin{align*}
  P(\texttt{isu}) &= \frac{0.5}{0.5 + 0.125} = 0.8, &
  P(\texttt{isk}) &= \frac{0.125}{0.5 + 0.125} = 0.2,
\end{align*}
During training, the byte sequence used is: \texttt{[\textless{}bos\textgreater{},T,i,r,a,m,i,s,u,\textless{}eos\textgreater{}]}.
 We provide additional details in
Appendix~\ref{corner_cases}. \MarginalizeIt serves as the main baseline throughout the study.

\subsection{End-Of-Token Logits}
\label{endoftoken}

\begin{figure}[H]
    \centering
    \includegraphics[width=0.95\textwidth]{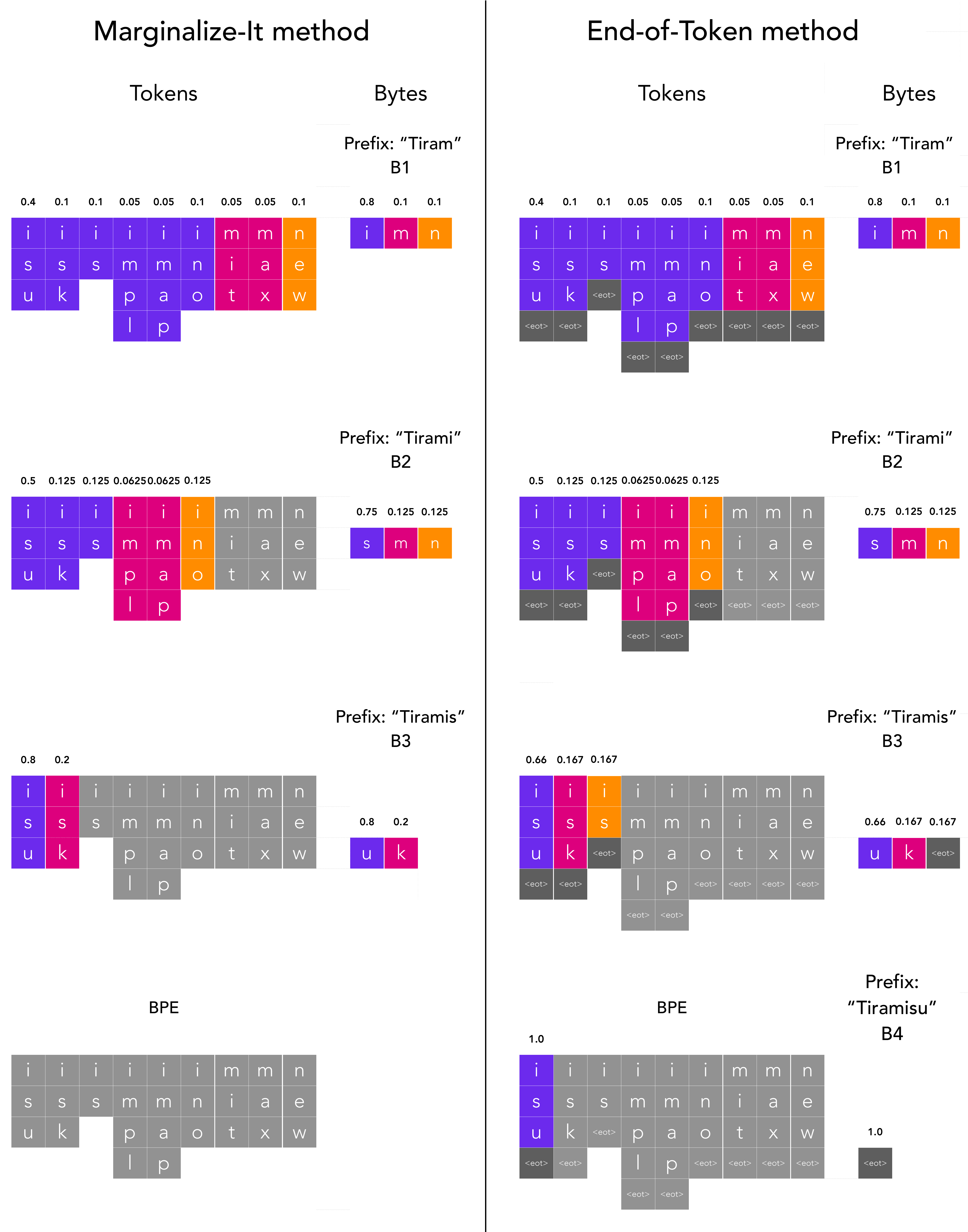}
\caption{A step-by-step illustration of the two logit conversion methods. The first column demonstrates the \MarginalizeIt \texttt{Byte Logits} formation process. Consider the tokenized prompt \texttt{Tiramisu = [<bos>, T, iram, isu, <eos>]}, and in particular the next-token distribution following the token \texttt{iram}, for which the ground truth is \texttt{isu}. Each column represents the decoded byte sequence of a BPE token from the full vocabulary (pruned here for clarity) of the Llama 3-8B model, with the token's probability shown at the top of the column; columns sharing a color share a byte prefix. To compute the first-byte distribution, the columns sharing the byte prefix \texttt{B1} are aggregated. Next, we subset the vocabulary by the ground-truth byte \texttt{i} and aggregate the probabilities over \texttt{B2}. Finally, we repeat this process for \texttt{B3}. A limitation of the \MarginalizeIt method is that it does not account for all possible continuations after the BPE token \texttt{is} while predicting \texttt{B3}: here, for example, only the tokens \texttt{isu} and \texttt{isk} are considered. The \EndOfToken method (second column) addresses this problem by preserving the language modeling distribution exactly, assigning the residual probability mass to the \texttt{<eot>} token, as illustrated for the prediction of byte \texttt{B3}.}
    \label{fig:bpe_to_byte_figure}
    
\end{figure}

The \EndOfToken{} method is essentially the \MarginalizeIt{} method with two
key changes. First, while forming the byte logits after teacher inference, we append
an \texttt{<eot>} token to mark the end of each BPE token as shown in Figure~\ref{fig:bpe_to_byte_figure} (column~2). Second, we expand the
byte vocabulary by one to store the \texttt{<eot>} probabilities and then apply marginalization over byte positions to form the byte logits. During training, we append
\texttt{<eot>} after every BPE token in the pretraining dataset: \texttt{[<bos>, <eot>,T,<eot>,i,r,a,m,<eot>,i,s,u,<eot>,<eos>,<eot>]}. This resolves the
missing-probability problem of \MarginalizeIt{}. To compute \texttt{B3} distribution, unlike \MarginalizeIt{}, \EndOfToken{}
considers \emph{all} tokens with prefix \texttt{is}. Since the total probability of tokens starting with prefix \texttt{is} is 0.5 + 0.125 + 0.125 = 0.75:
\begin{align*}
  P(\texttt{isu<eot>}) &= \frac{0.5}{0.75} = 0.667, &
  P(\texttt{isk<eot>}) &= \frac{0.125}{0.75} = 0.167, &
  P(\texttt{is<eot>})  &= \frac{0.125}{0.75} = 0.167.
\end{align*}
In this way, the probability mass of the possible continuations of \texttt{is}
is absorbed into the \texttt{<eot>} token.  This approach is exact, does not require additional inferences over teacher.
Inserting \texttt{<eot>} tokens also has an unintended but remarkable consequence: the \twentyfivepercent additional compute it spends per unit of data (one \texttt{<eot>} every approx 4.5 bytes) turns out to further improve asymptotic performance relative to \MarginalizeIt{} method.

\section{Scaling Study Design } \label{design}

\subsection{Experimental Setup} 
In this section we describe the Model Architecture, Pretraining Dataset and Training Details. 
We provide specifics of these settings in Section \ref{experimental_setup}. \\

\noindent \textbf{Model Architecture.} We use the transformer architecture \citep{vaswani2017attention} to train all of the layer-parameter-matched models with $\approx$ 1 billion parameters. We use three variants: \BPETransformer, \ByteTransformer, and \EndOfTokenTransformer listed  in Appendix (Table \ref{tab:model_architecture}). While the three model architectures have equal total layer parameters (1.28B), the \BPETransformer (1.81B) is in fact larger due to the larger vocabulary size. We stick to the Llama-3 \citep{grattafiori2024llama} implementation for the model design choices.  \\

\noindent \textbf{Training Dataset.} We use the Llama-2 \citep{meta2024llama32} training mixture for our supervised and distillation experiments. Our study is designed such that both the token logits and byte logits occupy the same amount of memory for the same amount of data. See section \ref{storage_cost_appendix} for the full storage cost breakdown.   \\

\noindent \textbf{Training Details.} We train our models with the AdamW \citep{loshchilov2017decoupled} optimizer and sweep three learning rates $\in$ \{1e-3, 4e-3, 8e-3\}. We use 10\% warmup steps for each data scale. We provide additional details on the loss function in Appendix (\ref{loss_function}) \\

    \subsection{Scaling Along Two Axes: 
    Tokenization and Training Objectives}  \label{scalingaxes}
Our study seeks to explore scaling trends of layer-parameter-matched models with overtraining for six experiments described below. This enables us to understand the complex relationships between the Supervised (Cross-Entropy Loss)  and Distillation pretraining as well as the effects of the tokenization schemes: \tokens, \bytes, \bytesweot on the training dynamics. Table \ref{tab:scalingaxes} details the six experimental settings in our overtraining study. Model size and  FLOPs/unit values are obtained from Table \ref{tab:model_architecture}. For \MarginalizeIt and \EndOfToken distillation we use the logit conversion process described in Section \ref{algorithms}.\\

\begin{table}[H]
\centering
\small
\scalebox{0.75}{\begin{tabular}{@{}llcllcc@{}}
\toprule
Experiment & Tokenization & Vocab. size & Objective & Architecture & Model size & FLOPs/unit \\
\midrule
\BPESupervised        & \tokens (Llama 3-8B)     & 128{,}256 & Cross-Entropy &  \BPETransformer  & 1.81B      & $8.18 \times 10^9$  \\
\BPEDistilled         & \tokens (Llama 3-8B)    & 128{,}256 & Distillation (top-k)  &  \BPETransformer    & 1.81B      & $8.18 \times 10^9$  \\

\midrule

\BytesSupervised      & \bytes (256 + 4 special)      & 260       & Cross-Entropy &  \ByteTransformer    & 1.28B    & $8.81 \times 10^9$  \\
\MarginalizeDistilled & \bytes (256 + 4 special)     & 260       & Distillation (approximate)  &  \ByteTransformer & 1.28B      &  $8.81 \times 10^9$ \\

\midrule

\BytesEotSupervised   & \bytesweot (256 + 5 special)  & 261       & Cross-Entropy &   \EndOfTokenTransformer  & 1.28B  & $9.44 \times 10^9$\\
\EndOfTokenDistilled  & \bytesweot (256 + 5 special) & 261       & Distillation (Exact) &  \EndOfTokenTransformer & 1.28B   & $9.44 \times 10^9$ \\
\bottomrule
\end{tabular}}
\caption{Six experiments in our overtraining study across the training-objective and tokenization axes.}
\label{tab:scalingaxes}
\end{table}

\subsection{Benchmark Evaluations} \label{benchmarks}

\noindent\textbf{Multiple Choice Question Answering.}
We evaluate performance on four benchmarks which evaluate the ability of a model to solve multiple choice questions. Specifically we use ARC-Easy \citep{clark2018think}, ARC-Challenge \citep{clark2018think}, HellaSwag \citep{zellers2019hellaswag}, and PIQA \citep{bisk2020piqa} to benchmark the token and byte checkpoints with and without distillation.  \\

\noindent\textbf{Language Generation Tasks.}
We further evaluate the language modeling capabilities of the byte and token models on MBPP \citep{austin2021program} and Natural Questions \citep{kwiatkowski-etal-2019-natural}.\\

\noindent\textbf{Machine Translation.}
Lastly we also evaluate our trained checkpoints on English to German and German to English translation using the Flores \citep{goyal2022flores} benchmark.\\

\section{ Scaling Trends I: Validation BPB vs Training FLOPS} \label{bpbvsflops}
In this experiment we analyze the overtraining behavior of the six experiments described in \ref{scalingaxes}. We plot the  BPB of the model on a held out validation dataset.

\subsection{Validation BPB vs Training FLOP Plots } \noindent We fit the power law for validation BPB with respect to FLOPs using equation (\ref{eq:bpb_flops_scaling_law}) where $x$ is the number of FLOPs and y is the BPB at a given compute budget (Table \ref{tab:scaling_laws_eqs_4e3}). The $y$, asymptotically approaches c.:

\begin{equation}
    y = b \cdot x^a + c
    \label{eq:bpb_flops_scaling_law}
\end{equation}

\begin{figure}[H]
    \centering
    \includegraphics[width=1.0\textwidth]{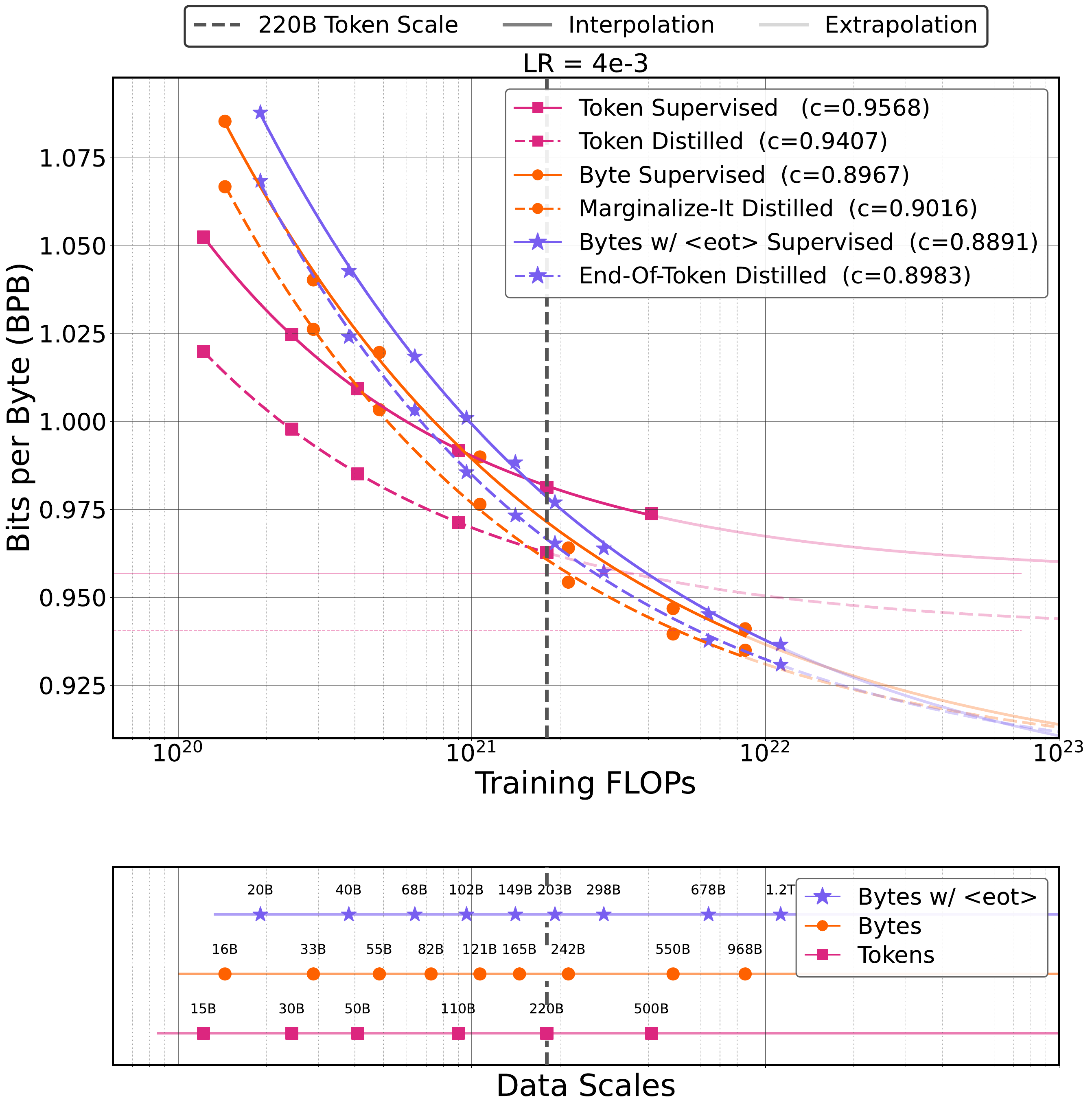}
    \caption{  The figure shows our scaling curves for different data scales, for a learning rate of 4e-3 and a fixed parameter count of approximately 1 billion parameter transformer models(see Table \ref{tab:model_architecture} for details). Note that data scales plot serves as an alternative x axis to training flops. Since the FLOPs per unit changes for the three models, we specify the number of units used for training. The distilled models are trained using Llama 3-8B as a teacher model. The validation BPB of the teacher model is 0.85511. 
    \label{fig:bpb_vs_flops_4e3}
    }
\end{figure}

\begin{table}[H]
\centering
\caption{Power-law scaling equations $y = b \cdot x^{a} + c$ fitted to BPB vs.\ training FLOPs for LR=4e-3.}
\label{tab:scaling_laws_eqs_4e3}
\vspace{6pt}
\begin{tabular}{lll}
\toprule
 Scenario & Equation & $R^2$ \\
\midrule

 \BPESupervised & $y = 1.075 \times 10^{9} \cdot x^{-0.500369} + 0.9568$ & 0.9999 \\
  \BPEDistilled & $y = 2.803 \times 10^{8} \cdot x^{-0.475375} + 0.9407$ & 1.0000 \\
   \midrule
  \BytesSupervised & $y = 4.467 \times 10^{6} \cdot x^{-0.365844} + 0.8967$  & 0.9980  \\

  \MarginalizeDistilled & $y = 2.706 \times 10^{7} \cdot x^{-0.407427} + 0.9016$ & 0.9974 \\
   \midrule
  \BytesEotSupervised & $y = 3.080 \times 10^{6} \cdot x^{-0.354594} + 0.8891$ & 0.9982 \\
  \EndOfTokenDistilled & $y = 2.734 \times 10^{7} \cdot x^{-0.404740} + 0.8983$ & 0.9990 \\

\bottomrule
\end{tabular}
\end{table}

\begin{table}[H]
\centering
\caption{
 We observe that the amount of FLOPs spent on a fixed amount of data (one BPE token in this case) affects the asymptotic BPB. Both supervised and distilled models asymptotically follow descending BPB trend as the Total FLOPs Per Token increase. The FLOPs/Unit values are obtained from Table \ref{tab:model_architecture}. Depending on the model architecture, a unit can be a token or a byte. We note that \EndOfTokenTransformer uses \twentyfivepercent extra FLOPs per token compared to \ByteTransformer. 
}

\vspace{6pt}
\label{tab:asymptotic_bpb_comparison_4e3}
\scalebox{0.86}{\begin{tabular}{llcccccc}

\toprule
\multirow{2}{*}{Method} & \multirow{2}{*}{Architecture} & \multirow{2}{*}{FLOPs / Unit} & \multirow{2}{*}{Units} & \multirow{2}{*}{Total FLOPs } & \multicolumn{1}{c}{Asymptotic BPB} \\
\cmidrule(lr){6-6}
& & & Per Token & Per Token  & 4e-3  \\
\midrule

\BPESupervised        &  \BPETransformer  & $8.18 \times 10^9$ & 1.0 & $8.18 \times 10^9$  &  0.9568 \\

\BytesSupervised       &      \ByteTransformer            & $8.81 \times 10^9$ & 4.5 & $39.65 \times 10^9$   & 0.8967 \\

\BytesEotSupervised   &  \EndOfTokenTransformer  & $9.44 \times 10^9$ & 5.5 & $51.92 \times 10^9$  & 0.8891 \\

\midrule

\BPEDistilled      &   \BPETransformer   & $8.18 \times 10^9$ & 1.0 & $8.18 \times 10^9$   & 0.9407  \\

\MarginalizeDistilled     &   \ByteTransformer      & $8.81 \times 10^9$ & 4.5 & $39.65 \times 10^9$  &  0.9016 \\

\EndOfTokenDistilled   & \EndOfTokenTransformer  & $9.44 \times 10^9$ & 5.5 & $51.92 \times 10^9$  &  0.8983 \\
 \bottomrule
\end{tabular}}
\end{table}

We notice several trends from these plots:
\begin{enumerate}
    \item First, both \MarginalizeDistilled (approximate) and \EndOfTokenDistilled (exact) methods demonstrate improvements over their supervised counterparts. 
    \item \MarginalizeDistilled training, despite being an approximation works very well in practice. \EndOfTokenDistilled method is exact and yields consistent improvements over the supervised training.
    \item Second, we also observe that \BPETransformer models have the lowest BPB at lower compute budgets, and appear to saturate faster. The \EndOfTokenTransformer have the worst BPB to begin with yet yield the best asymptote in the infinite compute regime. 
    \ByteTransformer training runs, appear to be sandwiched between the other two scenarios at the beginning and the end of the training runs. 

    \item From Table \ref{tab:asymptotic_bpb_comparison_4e3}, we observe that asymptotically for both supervised and distilled models: 1) \EndOfTokenTransformer BPB $<$ \ByteTransformer BPB $<$ \BPETransformer BPB. We also observe that the supervised and distilled models with the same tokenization scheme exhibits small differences in the asymptotic value.

\item We also observe interesting trends across all training runs. At around $1.8 \times 10^{21}$ FLOPs, the \BPESupervised training run appears to have the highest BPB of all, with all other five curves demonstrating lower BPB values. Does this mean that all other five models at that compute budget are better than \BPESupervised? To find out, we turn to benchmark evaluations in Section \ref{bench_vs_flops}.  \label{lowerbpbhyp}

\end{enumerate}

\section{Scaling Trends II: Task Performance vs Training FLOPs} \label{bench_vs_flops}

In this experiment we seek to understand if the observations from the validation BPB curves from section \ref{bpbvsflops} are consistent on a variety of benchmarks. We evaluate the capabilities of our trained models on a total of eight benchmarks spanning three categories: Multiple choice question answering, Generative, and Machine Translation tasks (Refer to section \ref{benchmarks} for more details).

\subsection{Results on Multiple Choice Question Answering Benchmarks} \label{mcqs}
\begin{figure}[H]
    \centering
    \includegraphics[width=1.0\textwidth]{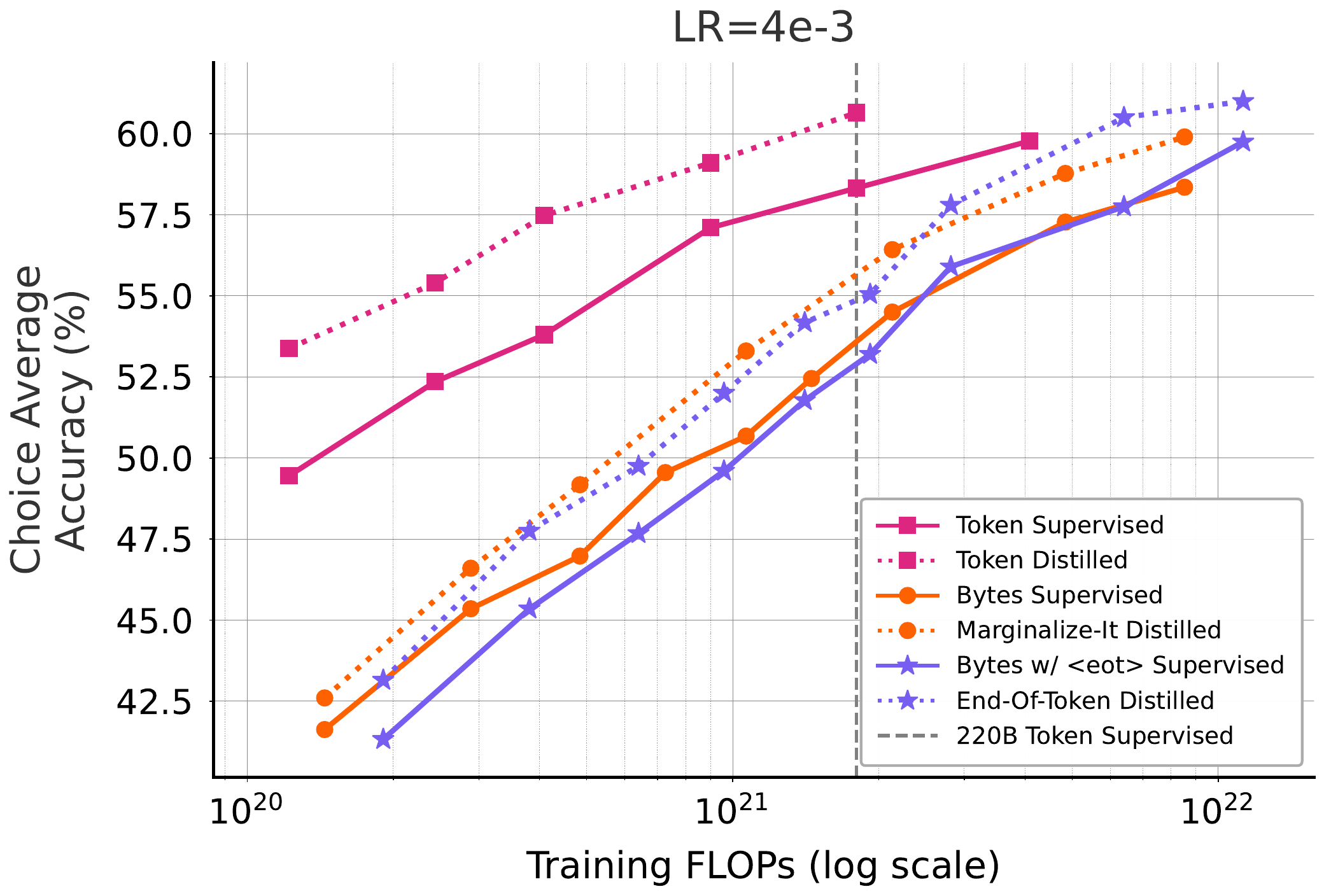}
    \caption{ We plot Accuracy vs training FLOP curves for all the six scenarios described in section \ref{scalingaxes}. For all of the Multiple Choice Question Answering Benchmarks, we observe that the token models show better performance at low compute budgets and eventually plateau. \ByteTransformer and \EndOfTokenTransformer models: both supervised and distilled, start worse yet improve at a much faster rate.   
    \label{fig:choice_benchmark_4e3_avg}
    }
\end{figure}

To evaluate the capabilities of trained models on four multiple choice question answering tasks we plot the \% Accuracy vs Training FLOP trends for our six experiments in Figure \ref{fig:choice_benchmark_4e3_avg}. In addition we also report the averaged accuracy across four benchmarks. We observe that:
\begin{enumerate}
    \item The averaged choice accuracy improves for all of the 6 experiments, and the distilled models consistently show better performance compared to the supervised models. This trend also holds for the benchmarks individually.

    \item \BPETransformer models start best, outperforming all of the byte curves at lower compute budgets. All of the byte curves start worse but appear to be improving faster than the token models, eventually catching up with them.

    \item If we compare within the four byte experiments,both supervised and distilled \EndOfTokenTransformer start worse, and eventually surpass the \ByteTransformer models  methods respectively.

    \item  \BPETransformer models, both supervised and distilled have the best performance at the beginning and appear to be eventually saturating, a pattern also observed in Validation BPB vs Training FLOPs curves.

    \item Lastly we observe that at 220B token supervised budget only the distilled \BPETransformer outperforms its supervised counterpart curve but the byte curves, both supervised and distilled \ByteTransformer and \EndOfTokenTransformer models are much worse, which is counterintuitive to the key observation in \ref{lowerbpbhyp}.  
\end{enumerate}

\subsection{Results on Language Generation Benchmarks} \label{langgen}
\begin{figure}[H]
    \centering
    \includegraphics[width=1.0\textwidth]{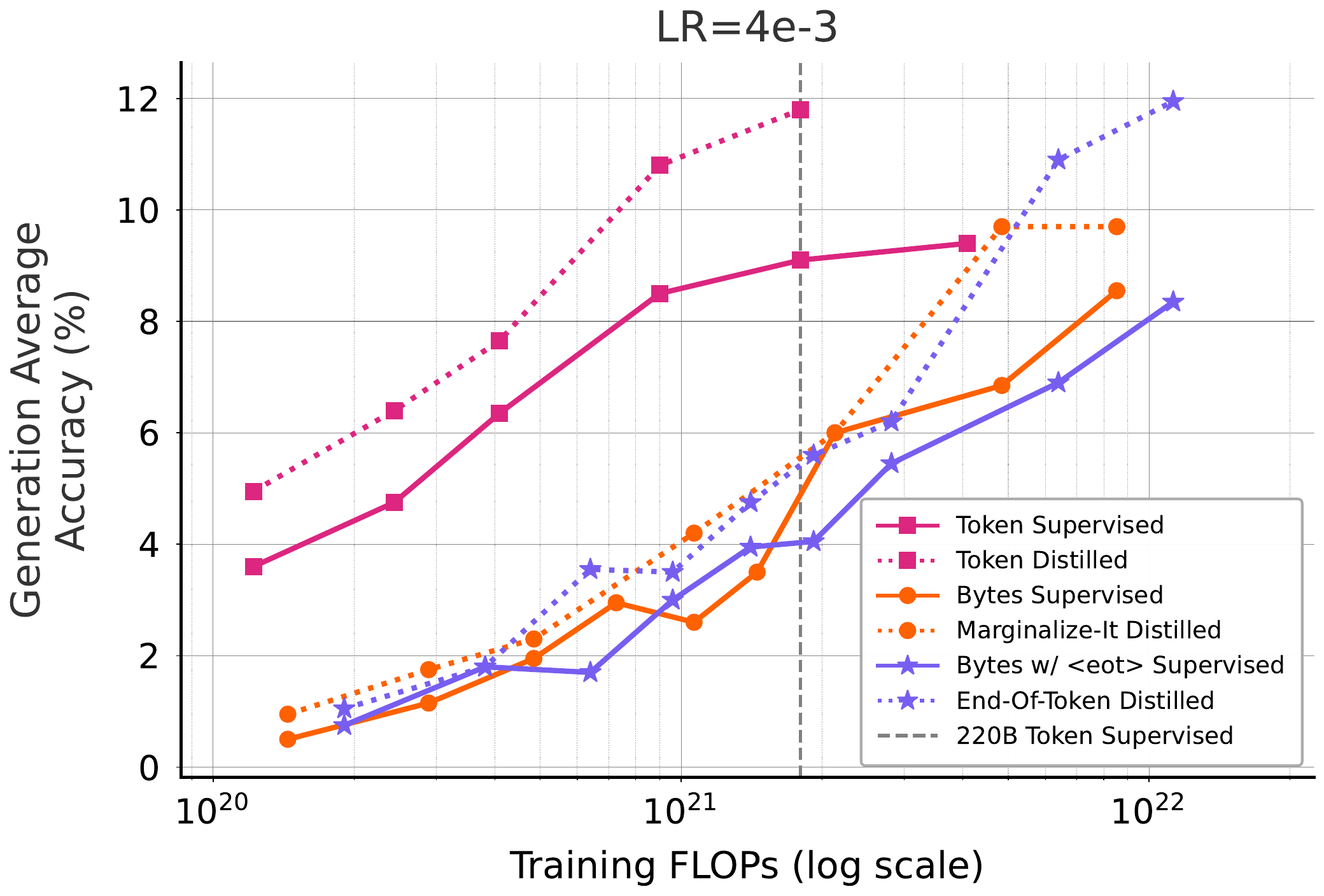}
    \caption{ We visualize $\%$ Accuracy vs training FLOPs curves for language generation tasks. On generative tasks, token models achieve better performance at lower compute budgets and appear to plateau with the training FLOPs. \byte models and \bytesweot models show a different scaling behavior, with much worse performance at a lower compute budget which continues to improve with FLOPs with no sign of saturation.
    \label{fig:gen_benchmark_4e3_avg}
    }
\end{figure}

In this experiment we study the language generation capabilities of the trained models on two benchmarks ( Figure \ref{fig:gen_benchmark_4e3_avg}) (more details in section \ref{benchmarks}).  We observe that:
\begin{enumerate}
    \item The trends are similar to Multiple Choice benchmarks. However the saturation effect of \BPETransformer models seems to be less pronounced, potentially because the tasks are much harder. 
    \item In all cases, distilled models appear to be better for the same compute budget and the  result about observation \ref{lowerbpbhyp} at $1.8 \times 10^{21}$ FLOPs budget is observed in this case as well. Only the \BPEDistilled model has better performance while all other four byte curves are much worse in most cases. 
\end{enumerate}

\subsection{Results on Machine Translation Benchmarks}\label{translation}
\begin{figure}[H]
    \centering
    \includegraphics[width=1.0\textwidth]{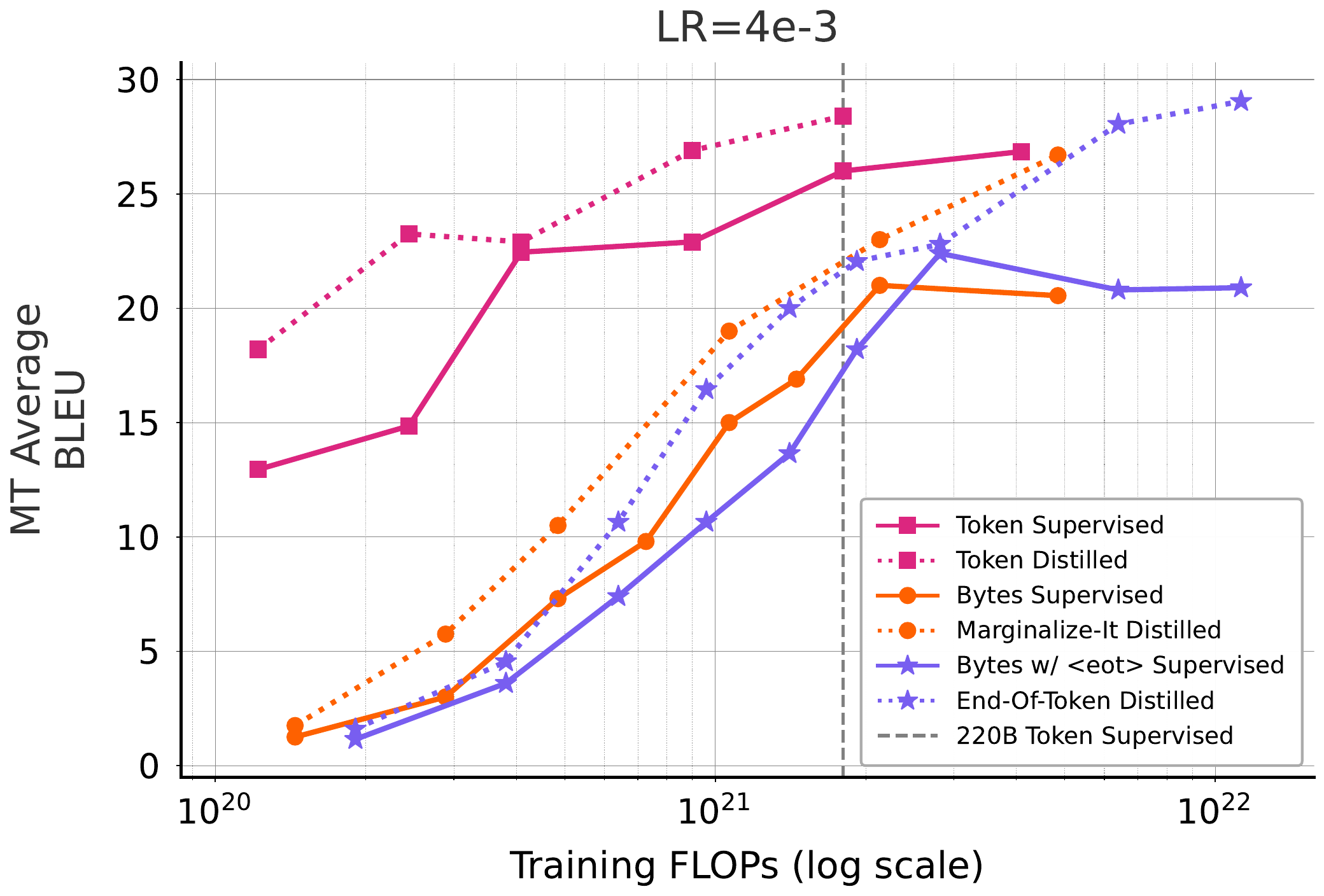}
    \caption{  We plot BLEU score for the translation tasks for different training budgets. We observe clearly different scaling behaviors for token and byte models. Token models can achieve significantly better performance at smaller compute budgets. The BLEU score appears to saturate with overtraining. Byte models both with and without the \texttt{<eot>} demonstrate extremely low BLEU scores at lower compute budgets yet the rate of improvement with overtraining is better compared to token models. 
    \label{fig:mt_benchmark_4e3_avg}
    }
\end{figure}

We also evaluate models on machine translation benchmarks and plot BLEU score \citep{10.3115/1073083.1073135} vs Training FLOP curves. The plots (Figure \ref{fig:mt_benchmark_4e3_avg}) show that:
\begin{enumerate}

    \item Byte model BLEU scores appear to be following an S-shaped trend with training FLOPs, improving slowly at low compute regimes, followed by a steeper improvement regions and then finally saturating at high compute budgets. 
    \item Next, the results on these benchmarks contradict the observation \ref{lowerbpbhyp} here as well. 

    \item \EndOfTokenDistilled models by preserving the teacher distribution exactly show significant improvements over \BytesEotSupervised models.
    
\end{enumerate}

\subsection{Concluding Remarks}

In section \ref{mcqs}, \ref{langgen}, and \ref{translation}, we consistently saw that despite achieving better validation BPB at $\approx 1.8 \times 10^{21}$ FLOPs, all four byte models, both distilled and supervised continue to be much worse compared to \BPESupervised model. What could explain this behavior?   In section \ref{perf_vs_bpb} we set out to explain this phenomenon. 

\begin{tcolorbox}[colback=hypothesisblue!5, colframe=hypothesisblue!80, arc=8pt, boxrule=1pt]

For the data points we observe, the distilled models achieve lower BPB and better overall downstream task performance than their supervised counterparts. However, this does not always hold when comparing the \BPETransformer against the \ByteTransformer and \EndOfTokenTransformer models: a \BPETransformer can have worse BPB yet better downstream performance, as observed at the 220B token data scale.
\end{tcolorbox} \label{when_it_holds_true}

\section{Scaling Trends III: Task Performance vs Validation BPB} \label{perf_vs_bpb}

We now turn to explaining how the observations in sections \ref{bpbvsflops} and \ref{bench_vs_flops} fit together. For all of the three downstream task categories we study the relationship between the downstream task performance and the validation BPB. To do so, we use the formulation of scaling laws for downstream tasks in \citep{gadre2024language}. We plot the Average top-1 error, essentially, $(1-\% \text{ Accuracy})$ against the validation BPB, for Multiple Choice Question Answering (section \ref{mcq_err_vs_bpb}) and Language Generation (section \ref{lang_gen_vs_bpb}). Lastly, since Machine Translation (section \ref{machine_translation_vs_bpb}) uses BLEU as an evaluation metric, we plot the BLEU score vs the validation BPB.

\subsection{Results on Multiple Choice Question Answering Benchmarks} \label{mcq_err_vs_bpb}

\begin{figure}[H]
    \centering
    \includegraphics[width=1.0\textwidth]{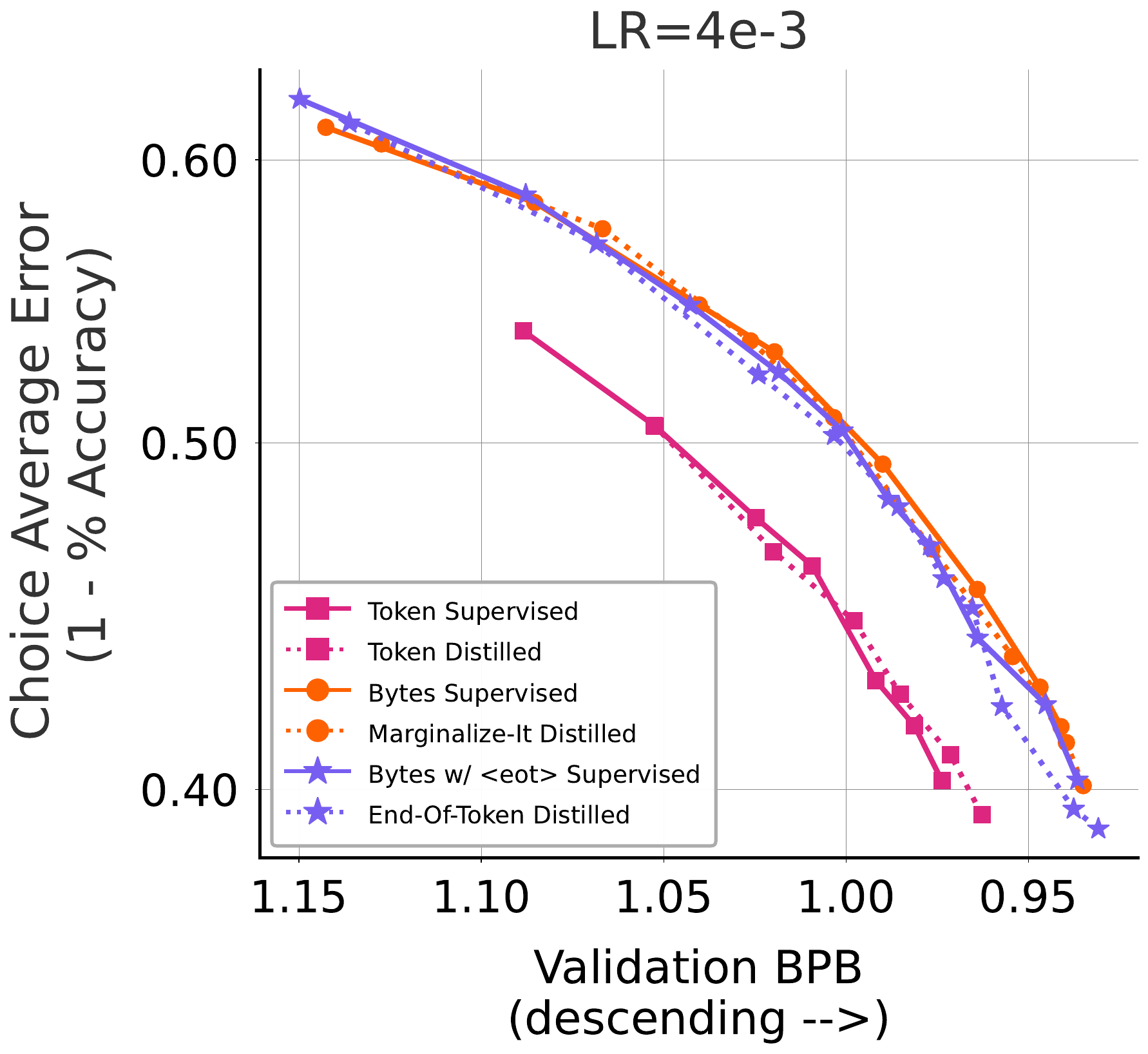}
    \caption{ We visualize Average-top-1 error vs the validation BPB curves for the six scenarios for LR=4e-3. All of these curves including the averaged metric appear to follow an exponential decay. \BPETransformer shows significantly different trend compared to \ByteTransformer and \EndOfTokenTransformer. Both the training objective (Cross-Entropy vs Distillation) as well as the tokenization scheme: \tokens, \bytes, \bytesweot appear to influence the Average Error vs Validation BPB trends.   }
    \label{fig:choice_benchmark_error_vs_bpb}
\end{figure}

\begin{enumerate}
    \item Across all benchmarks as well the averaged performance (Figure \ref{fig:choice_benchmark_error_vs_bpb}), all of the six scaling trends demonstrate significantly different behavior. First, if we compare the performances for constant BPB, they imply wildly different error rates. Second, the iso-error lines intersect with the \EndOfToken and \token curves at very different compute budgets.

    \item Second: Changes in the tokenization scheme i.e. the amount of FLOPs spent to model a fixed amount of data can dramatically change the scaling behavior. For instance, \ByteTransformer and \EndOfTokenTransformer models do not show significantly different scaling behaviors compared to \BPETransformer models.

    \item Lastly, the training objective (supervised as well as distillation) also appear to alter the scaling behavior.
\end{enumerate}

\subsection{Results on Language Generation Benchmarks}\label{lang_gen_vs_bpb}

We now visualize the Average top-1 error vs Validation BPB curves for the language generation tasks described in section \ref{benchmarks}. In addition, we also plot the averaged generation performance (in row 1).

\begin{enumerate}
    \item We observe that for language generation as well, the same BPB can mean different averaged downstream performance for the three experiment categories - (\token, \bytes, \bytesweot) as well as training objectives - supervised and distillation.  
\end{enumerate}

  \begin{figure}[H]
    \centering
    \includegraphics[width=1.0\textwidth]{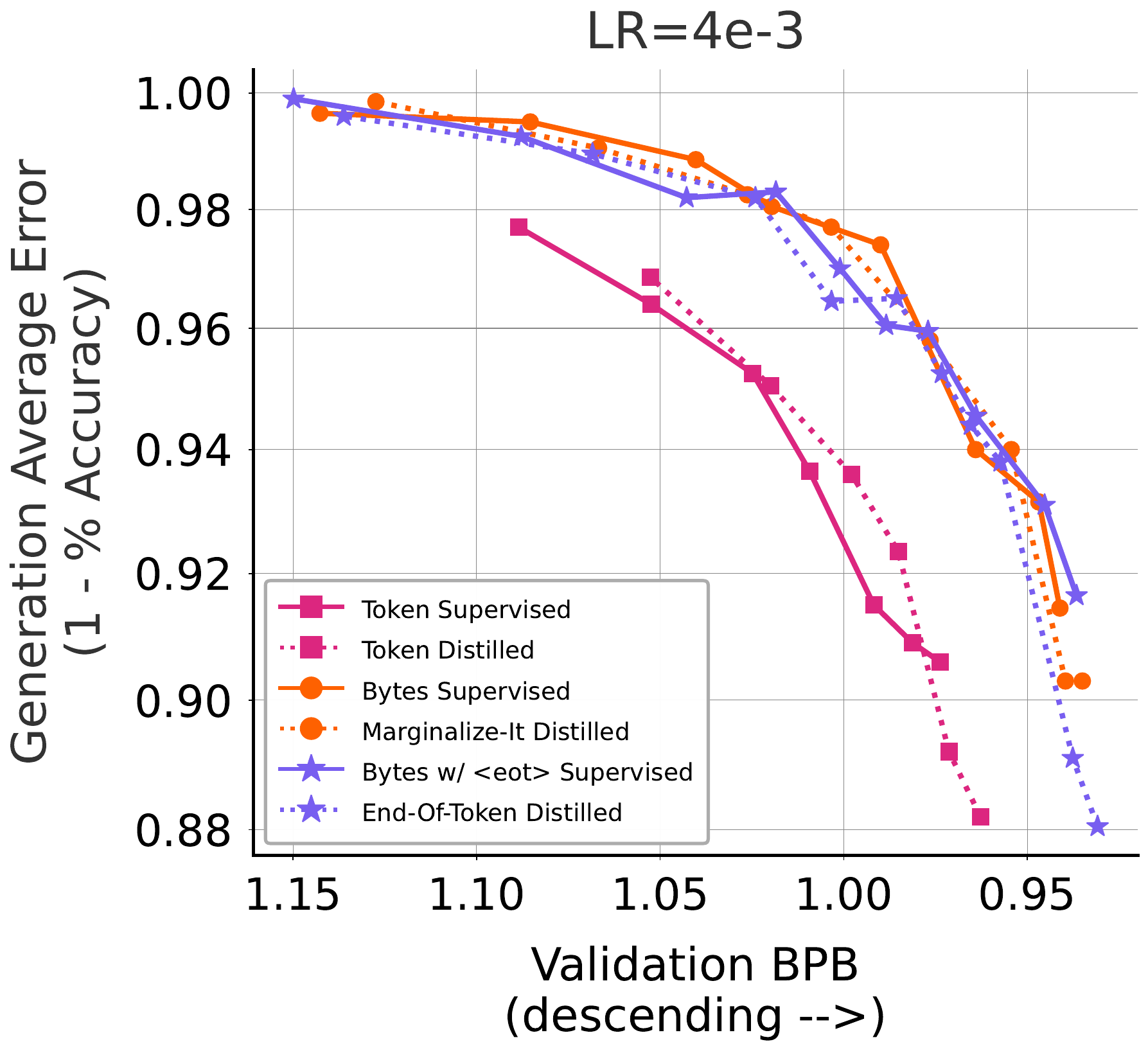}
    \caption{ We plot the Average-top-1 error vs BPB curves for the MBPP and the NQ benchmarks. We notice an exponential decay for all three scenarios, with significantly different trends for the \BPETransformer vs \ByteTransformer and \EndOfTokenTransformer models.  Distilled models as well demonstrate slightly different scaling behavior compared to the supervised counterparts. }
    \label{fig:gen_benchmark_error_vs_bpb_4e3_avg}
 \end{figure}

\subsection{Results on Machine Translation Benchmarks}\label{machine_translation_vs_bpb}
\begin{figure}[H]
    \centering
    \includegraphics[width=1.0\textwidth]{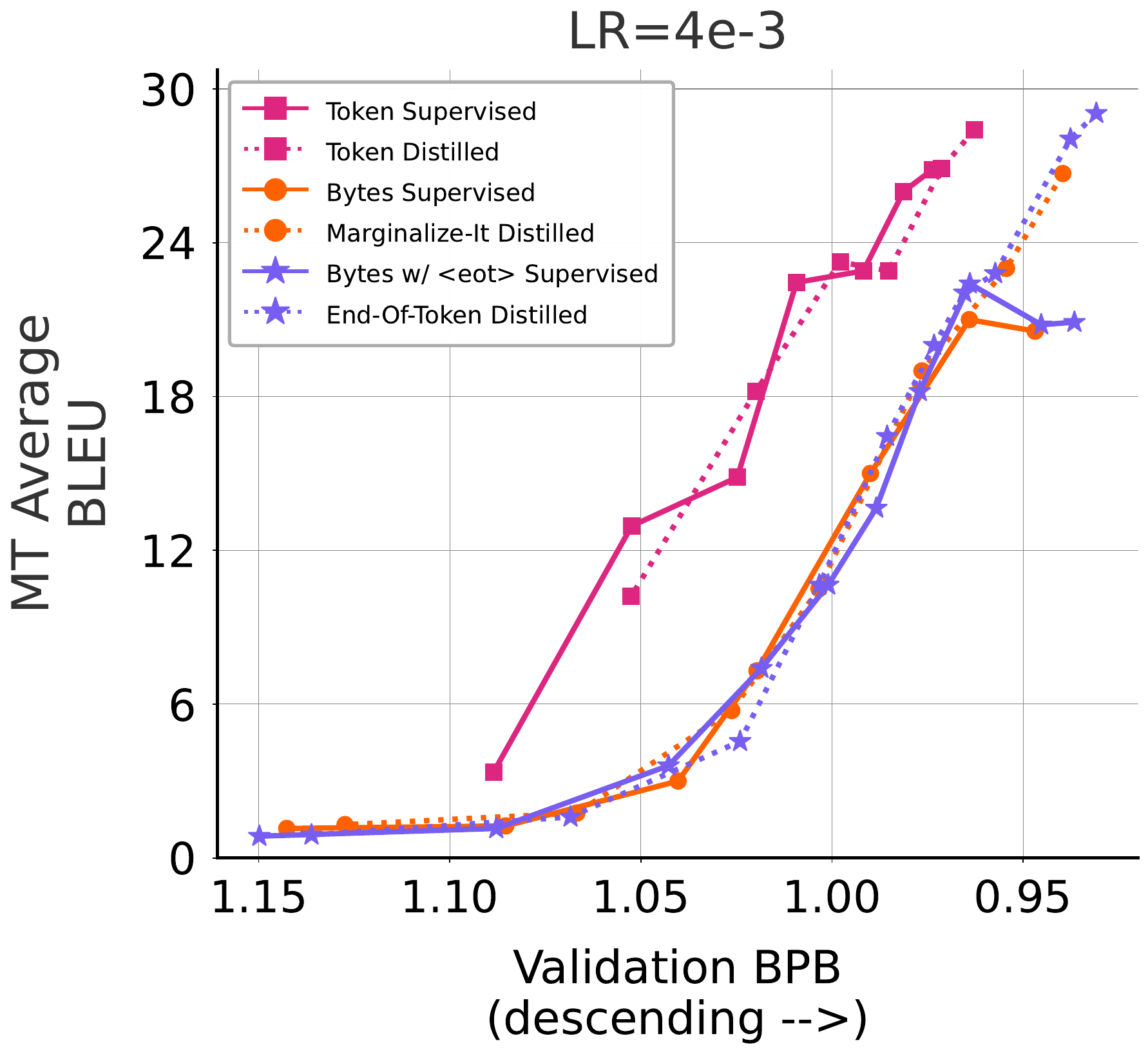}
    \caption{ We plot averaged BLEU score vs Validation BPB performance, and on the averaged BLEU performance for the six scenarios for LR=4e-3. We notice the differences in the scaling behavior across tokenization schemes as well as the training objectives. \EndOfToken distillation appears to exhibit monotonic trend compared to the noisy supervised curves. }
    \label{fig:mt_benchmark_vs_bpb_4e3_avg}

\end{figure}

Lastly we plot BLEU score vs Validation BPB curves for Machine Translation tasks (Figure \ref{fig:mt_benchmark_vs_bpb_4e3_avg}). We observe that:
\begin{enumerate}
    \item The \BPETransformer models, \ByteTransformer models and \EndOfTokenTransformer models show different scaling behaviors.
    \item On average, \EndOfToken distillation appears to exhibit monotonic trend compared to the noisy \BytesEotSupervised curves.
\end{enumerate}

\subsection{Fitting the Downstream Task Error vs Validation BPB Scaling Law}\label{section:downstream_scaling}

\begin{figure}[H]
    \centering
    \includegraphics[width=0.8\textwidth]{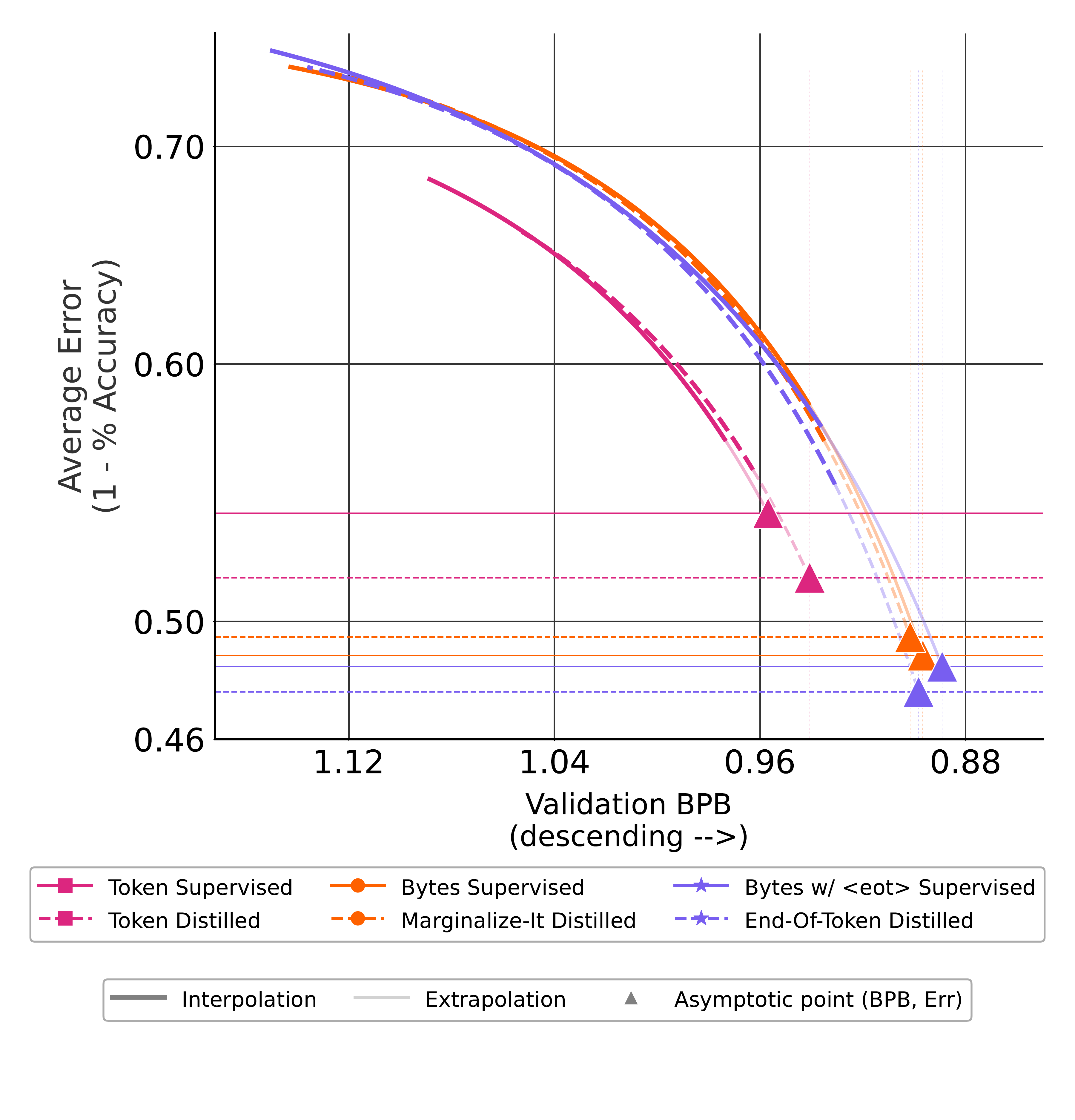}
    \caption{ We fit the downstream error scaling laws for six benchmarks with respect to Validation BPB. All six experiments follow a smooth trend. \BPETransformer curves show significantly different behavior compared to the \ByteTransformer and \EndOfTokenTransformer curves. The training objective (distillation vs supervised (Cross Entropy) also affects the scaling laws. \EndOfTokenDistilled, by preserving the teacher distribution exactly as well as with the highest amount of compute spent  on fixed amount of data outperforms all of the other models asymptotically. \byte and \bytesweot models, despite having fewer total parameters asymptotically surpass the \token models.  }
    \label{fig:downstream_scaling_law_fitting_notes_scatter}

\end{figure}

We fit the equation \ref{eq:downstream_error} for our six experiments on averaged top-1 error calculated over total 4 Multiple Choice and 2 language generation benchmarks (from \ref{benchmarks}) where $y$ denotes the BPB value, $z(y)$ is the Averaged Top-1 Downstream Task Error, and $\epsilon$, k and $\gamma$ are fit using data. The scaling law equations are detailed in Table \ref{tab:downstream_scaling_law_eqs_4e3}. The interpolations and extrapolations are visualized in Figure \ref{fig:downstream_scaling_law_fitting_notes_scatter}. The curves are bounded by the asymptotic BPB c, obtained from the power law equations in Table \ref{tab:scaling_laws_eqs_4e3}.
\begin{equation}
    z(y) = \epsilon - k \cdot e^{-\gamma y}
    \label{eq:downstream_error}
\end{equation} 

\begin{table}[H]
  \centering
  \caption{Fitted downstream error scaling laws \DownstreamEquation (equation \ref{eq:downstream_error}) for each model variant and learning rate of 4e-3, where $y$ is the validation BPB.}
  \vspace{6pt}
  \label{tab:downstream_scaling_law_eqs_4e3}
  \scalebox{1}{
  \begin{tabular}{@{}llll@{}}
    \toprule
    {Method} & \multicolumn{1}{c}{Model Architecture} & \multicolumn{1}{c}{Scaling law equation $z(y)$}\\
    \midrule
    \BPESupervised &\BPETransformer &  $0.7501 - 934.3212\,e^{-8.7782\,y}$  \\
    \BPEDistilled &\BPETransformer & $0.7383 - 1292.0124\,e^{-9.2132\,y}$  \\
    \midrule
    \BytesSupervised &\ByteTransformer  & $0.7686 - 1231.1241\,e^{-9.3532\,y}$  \\
    \MarginalizeDistilled & \ByteTransformer  & $0.7721 - 1137.3004\,e^{-9.2265\,y}$ \\
    \midrule
   \BytesEotSupervised & \EndOfTokenTransformer  & $0.7972 - 185.6303\,e^{-7.1822\,y}$  \\
    \EndOfTokenDistilled & \EndOfTokenTransformer  & $0.7776 - 827.6076\,e^{-8.8125\,y}$  \\
    \bottomrule
  \end{tabular}}
\end{table}

\subsection{Downstream Task Performance Upper Bound}
To calculate the asymptotic downstream task performance upper bound, we use the Average top-1 error at asymptotic validation BPB c, for each curve obtained from equation \ref{tab:scaling_laws_eqs_4e3}. We then compute the asymptotic performance at that BPB as shown in Figure \ref{fig:downstream_scaling_law_fitting_notes_scatter}. 
 Table \ref{tab:downstream_scaling_law_eqs_4e3} shows the asymptotic BPB as well as the task performance of the six experiments. 

We observe that:
\begin{enumerate}
    \item We notice that distilled \BPETransformer asymptotically outperforms over its supervised counterpart.
    \item \EndOfTokenDistilled method outperforms both \BPEDistilled and \BytesEotSupervised asymptotically. 
    \item Distilled \ByteTransformer performs on par or suboptimally compared to its supervised model potentially because it uses an approximation of the teacher distribution. It however still outperforms asymptotic distilled \BPETransformer accuracy. 

    \item Lastly, we note that by spending around \twentyfivepercent additional compute on a fixed amount of data, \BytesEotSupervised shows better asymptotic behavior compared to the \BytesSupervised model. \item \EndOfTokenDistilled shows better asymptotic downstream performance (lowest Average top-1 error) compared to the \BytesEotSupervised model.

\end{enumerate}

\begin{table}[H]
  \centering
  \caption{Asymptotic validation BPB ($\mathrm{c}$) and the corresponding downstream average accuracy $a^\star = 1-\%z^\star$ where $z^\star$ is the Asymptotic Average top-1 error
  predicted by the scaling law equation: \ref{eq:downstream_error}, $z^\star = \epsilon - k \cdot e^{-\gamma c}$ , where $c$ is obtained from the power law equations in Table \ref{tab:downstream_scaling_law_eqs_4e3} for each model variant and learning rate.}
\vspace{6pt}
  
  \label{tab:asymptotic_perf_4e3}
 \scalebox{1.2}{ \begin{tabular}{@{}lcccccc@{}}
    \toprule
    {Method} &  $\mathrm{c}$ & $a^\star$ (\%)  \\
    \midrule
    \BPESupervised  & 0.9568 & 46.0 & \\
    \BPEDistilled & 0.9407 & 48.4  \\
    \midrule
    \BytesSupervised & 0.8967 & 51.2 \\
    \MarginalizeDistilled  & 0.9016 & 50.5  \\
    \midrule
    \BytesEotSupervised  & 0.8891 & 51.6 \\
    \EndOfTokenDistilled  & 0.8983 & 52.4 \\
    \bottomrule
  \end{tabular}}
\end{table}

\subsection{When Lower BPB Does Not (Always) Imply Better Models}
Our observations show that BPB (Bits-Per-Byte) does not allow fair comparison across models with different training objectives (Distillation and Cross-Entropy) and tokenization schemes: \tokens, \bytes, \bytesweot. We show that the relationship between the BPB across \tokens, \bytes, \bytesweot is nuanced in the context of benchmark evaluations. These disparities potentially arise from the differences in probability distributions learned by spending more compute per unit data on a small amount of data vs small amount of compute spent per unit data on larger amount of data under a fixed compute budget. Although all of the \ByteTransformer and \EndOfTokenTransformer models surpass the \BPESupervised BPB training curve, they exhibit worse downstream task performance across benchmarks, falsifying observation \ref{lowerbpbhyp}  when the models being compared have different training objectives and tokenization schemes, emphasizing the importance of this final calibration step while determining the quality of a language model before real-world deployment. See appendix \ref{different_downstream} and section \ref{discussion} for additional discussion.

\section{ The Case for \EndOfToken Logits} \label{storage_case}
To understand if \EndOfToken logits are advantageous, it is essential to find out when they will if at all surpass the distilled token models. To achieve that we use the downstream error prediction scaling laws from \ref{section:downstream_scaling} as shown below:

\subsection{\Featherplot: Visualizing Iso-FLOP Point Connections }

 We visualize lines connecting Iso-FLOP points which we call \FeatherPlot to understand how the BPB(Bits-Per-Byte) value as well as the downstream task performance evolves with compute budget. To plot them we sample points from FLOP range $f1$ to $f2$. For each sampled point $x$ we compute $y$ using equation \ref{eq:bpb_flops_scaling_law}. Next we use the obtained validation loss in terms of BPB $y$ to compute the downstream error $z(y)$ using equation \ref{eq:downstream_error}. We connect the points obtained for three methods: \token, \MarginalizeIt, \EndOfToken distillation to analyze the trend.
We observe for example the learnig rate of 4e-3:
\begin{enumerate}
    \item At low compute budgets, distilled \BPETransformer performance leads significantly compared to \ByteTransformer and \EndOfTokenTransformer models. 
    \item As the compute budget progresses, distilled \EndOfTokenTransformer overtakes the \ByteTransformer models. 
    \item Eventually, distilled \EndOfTokenTransformer catches up with the \ByteTransformer too and continues to improve until the asymptotic value of the performance.  
\end{enumerate}

\subsection{Calculating Downstream Error Cross-Over Point with Token Distillation}
Given the validation BPB prediction and downstream error prediction scaling laws for distilled \BPETransformer, \ByteTransformer, \EndOfTokenTransformer models we want to find out at what compute budget \MarginalizeDistilled and \EndOfTokenDistilled methods will give the same asymptotic downstream error prediction as that of a \BPEDistilled model. 

\begin{table}[H]
  \centering
  \caption{Training compute at which distilled \ByteTransformer{} and \EndOfTokenTransformer{} match the asymptotic downstream performance of \BPETransformer. We take \BPEDistilled asymptotic BPB (the $c$ term of its fitted BPB--FLOPs power law; column Token Dist. BPB) and map it through \BPEDistilled benchmark scaling law to obtain its asymptotic downstream accuracy (column Accuracy). For each byte method we then invert its own benchmark scaling law at that accuracy to get the BPB it must attain (column Method BPB), and invert its BPB--FLOPs power law at that BPB to get the required training compute (column Training FLOPs).}
  \label{tab:crossover_compute}
  \renewcommand{\arraystretch}{1.3}
  \resizebox{\textwidth}{!}{
  \begin{tabular}{cccccc}
    \toprule
      \textbf{Method} & \textbf{Token Dist. BPB} & \textbf{Accuracy (\%)} & \textbf{Method BPB} & \textbf{Method Training FLOPs} \\
\midrule
      \MarginalizeDistilled & $0.9407$ & $48.4\%$ & $0.9102$ & $2.04 \times 10^{23}$ \\
      \EndOfTokenDistilled & $0.9407$ & $48.4\%$ & $0.9145$ & $6.33 \times 10^{22}$ \\

    \bottomrule
  \end{tabular}
  }
\end{table}

\begin{table}[H]
  \centering
  \caption{Effective training-data volume at the compute budget required to achieve the \token distillation asymptotic performance from Table \ref{tab:asymptotic_perf_4e3} for \MarginalizeIt and \EndOfToken methods. \textbf{Bytes} $=\text{Byte Units}\times\text{Multiplier}$ converts byte units into actual UTF-8 bytes of unique text, accounting for the fraction of each unit that is genuine content (Multiplier $=1$ for \MarginalizeDistilled; $=4.5/5.5$ for \EndOfTokenDistilled, whose sequences carry an extra \texttt{<eot>} per $4.5$ content bytes). \textbf{Data Savings} $=\text{Token to Byte}/\text{Bytes}$ is how much less unique text the byte method needs to reach the same performance. To calculate storage savings we use our storage cost calculations from Table \ref{tab:storage_cost_analysis} where we equate the storage cost of one \token with 4.5 \bytes to obtain k (for top-k).  \textbf{Storage Savings} $=(\text{Tokens}\times4.5)/\text{Byte Units}$ compares the byte footprint of \BPEDistilled approach against the raw byte-unit count. }
  \label{tab:crossover_data_volume_4e3}
  \renewcommand{\arraystretch}{1.3}
  \resizebox{1.0\textwidth}{!}{
  \begin{tabular}{llcc}
    \toprule
    \textbf{Metric} & \textbf{Formula} & \textbf{\MarginalizeDistilled} & \textbf{\EndOfTokenDistilled} \\
    \midrule
    Training FLOPs &  Compute to achieve \token distillation perf. (Table \ref{tab:asymptotic_perf_4e3}) & $2.04 \times 10^{23}$ & $6.33 \times 10^{22}$ \\
    FPU$_{\text{bytes}}$ & FLOPs per byte unit (Table \ref{tab:model_architecture}) & $8.81 \times 10^{9}$ & $9.44 \times 10^{9}$ \\
    FPU$_{\text{tokens}}$ &  FLOPs per token (Table \ref{tab:model_architecture})  & $8.18 \times 10^{9}$ & $8.18 \times 10^{9}$ \\
    Byte Units & $\text{ Training FLOPs}/\text{FPU}_{\text{bytes}}$ & $2.32 \times 10^{13}$ & $6.70 \times 10^{12}$ \\
    Multiplier & (for actual data) & 1 & 4.5/5.5
    \\
    Bytes & $\text{Byte Units}\times\text{Multiplier}$  & $2.32 \times 10^{13}$ & $5.48 \times 10^{12}$ \\
    Tokens & $\text{Training FLOPs}/\text{FPU}_{\text{tokens}}$ & $2.50 \times 10^{13}$ & $7.73 \times 10^{12}$ \\
    Token to Byte & $\text{Tokens}\times4.5$ & $1.12 \times 10^{14}$ & $3.48 \times 10^{13}$ \\
    Data Savings & $\text{Token to Byte}/\text{Bytes}$ & $4.85\times$ & $6.35\times$ \\
    Storage Savings & $\text{Tokens}/(\text{Byte Units}/4.5)$ & $4.85\times$ & $5.19\times$ \\
    \bottomrule
  \end{tabular}
  }
\end{table}

\subsection{Results}
\begin{enumerate}
    \item 
    
    We summarize our results in Table \ref{tab:crossover_data_volume_4e3}. Distilled \EndOfTokenTransformer training runs asymptotically outperform \BPETransformer model, hence we can compute the crossover points of \ByteTransformer and \EndOfTokenTransformer models with \BPETransformer model. 

    \item  Since \EndOfToken uses the \texttt{<eot>} tokens every 4.5 bytes, the actual text data is less than the byte units used for training. Taking this into account, \EndOfToken \texttt{Byte Logits} can reduce the storage cost to approximately \texttt{one-fifth} compared to token distillation when trained for upto 6.7 Trillion byte units or data equivalent to 1.2 Trillion Llama 3-8B tokens.

\end{enumerate}

\begin{tcolorbox}[colback=hypothesisblue!5, colframe=hypothesisblue!80, arc=8pt, boxrule=1pt]

\EndOfTokenTransformer demonstrates better asymptotic downstream performance compared to distilled \BPETransformer. It requires \texttt{one-sixth} text data and it can save storage costs upto \texttt{one-fifth}.

\end{tcolorbox} \label{bpbparadox}

\section{Comparisons with Open-Weight models} \label{llama_gemma}
We now compare our asymptotic predictions for each of the six experiments described in section \ref{scalingaxes} with the open weight models: \texttt{Llama 3.2-1B} model \citep{meta2024llama32}, \texttt{Gemma-3-1B-pt}   \citep{Kamath2025Gemma3T}, and \texttt{Gemma 2B} \citep{team2024gemma} in Table \ref{tab:bench}. We evaluate these models using \citep{eval-harness}.

\begin{table}[H]
\centering
\caption{Predicted asymptotic downstream accuracy (\%) in the for our
LR=4e-3 obtained from table \ref{tab:asymptotic_perf}, obtained by extrapolating the scaling trend. Comparisons with the open-weights model with our asymptotic predictions reveal our distilled \EndOfTokenTransformer model surpass the \cite{meta2024llama32}, \cite{Kamath2025Gemma3T}, and \cite{team2024gemma} models on averaged downstream tasks by upto $6.5\%$ and $8.1\%$, 2.1\% respectively. Additional details for llama and gemma models are provided in table \ref{tab:llama_gemma_bench_comparison}}
\vspace{2mm}
\label{tab:bench}
\begin{tabular}{l l c}
\toprule
Model & Method & Asymptotic Avg. Accuracy(\%) \\

\midrule
\BPETransformer   & \BPESupervised        & 46.0 \\
\BPETransformer   & \BPEDistilled         & 48.4 \\
\midrule
\ByteTransformer  & \BytesSupervised      & 51.2 \\
\ByteTransformer  & \MarginalizeDistilled & 50.5 \\
\midrule
\EndOfTokenTransformer & \BytesEotSupervised   & 51.6 \\
\EndOfTokenTransformer & \EndOfTokenDistilled  & \textbf{52.4} \\
\midrule

\texttt{Llama-3.2-1B}  & \cite{meta2024llama32}   & 45.9 ($\approx$ 9T BPE tokens) \\
\texttt{Gemma-3-1B-pt} & \cite{Kamath2025Gemma3T} & 44.3 ($\approx$ 2T BPE tokens) \\
\texttt{Gemma 2b} & \cite{team2024gemma}   &	50.3 ($\approx$ 3T BPE tokens) \\
\midrule

\end{tabular}
\end{table}

\section{Related Work} \label{relatedwork}
\noindent\textbf{Scaling Laws with Overtraining.} Which factors affect the power law exponent with overtraining has been a topic of great interest in deep learning. \cite{kaplan2020scaling} show that changing the architecture from transformers to LSTMs does not change the power law exponent for shorter contexts. \citep{bahri2024explaining} show that the power law exponent changes with data distribution. \citep{henighan2020scaling} show that the power law exponents change across modalities and also across image resolutions. \citep{bansal2022data, bahri2024explaining} show that the noise level in data can also significantly affect the exponents. In the context of language modeling \citep{hestness2017deep} show that character models learn relationships between characters with fewer samples compared to word level language models learn from words. Several recent works focus on the overtraining of language models. \citep{sardana2023beyond} demonstrate that when the inference budget is large enough, it is better to train a small model for longer. \cite{gadre2024language} derive scaling laws with overtraining for different model sizes and training datasets. \cite{limisiewicz2026compute} analyze the impact of data compression on scaling laws. \cite{lee2026efficiency} show that the autoregressive byte models approach performance parity with their BPE counterparts as compute scales. Distinct from prior work, we study power laws for $\approx$ fixed model size with overtraining along two axes: training objective and the tokenization scheme. We observe that both axes affect the exponent as seen from table (\ref{tab:scaling_laws}) and table (\ref{tab:asymptotic_perf}). We show that a simple addition of an \texttt{<eot>} token every $\approx$ 4.5 bytes can improve a model's asymptotic downstream performance: an effect explained by the additional FLOPs spent for a fixed amount of data from the increased sequence length at fixed context. \\

\noindent\textbf{Scaling Laws for Downstream Tasks.} \cite{schaeffer2023emergent} argue that the capabilities that appear to be emergent in language models are due to the researcher's choice of metric. In the context of transfer learning \cite{isik2025scaling} study how the choice of the pretraining data affects downstream cross entropy and Machine Translation performance (BLEU) score.
\cite{gadre2024language} establish the relationship between the Average top-1 downstream error across a number of benchmarks and the validation Cross-Entropy Loss. They demonstrate that using  validation loss alone while comparing models trained on different data distributions can be misleading.  \cite{owen2024predictable} show that the aggregate benchmark performance is decently predictable, though predicting individual performance is harder.   \cite{bhagia2024establishing} establish a two stage approach where they use N (\# of parameters)  and D (\# of tokens) to predict the intermediate task loss which is then mapped to the task performance. \cite{lourie2025scaling} conduct a meta-analysis of the downstream tasks considered in \cite{gadre2024language} to demonstrate that the task-specific scaling trends may vary. They further compare them with the findings from \cite{magnusson2025datadecide} to show that the scaling behavior depends on the experimental setup and that the choice pretraining corpus, validation corpus, or downstream task can affect the scaling laws. Our work studies how the training objective and the tokenization scheme can affect the downstream performance with Training FLOPs. \\

\noindent\textbf{Pretraining Distillation.}
\cite{hinton2015distilling} introduced distillation and showed that a small model can be improved by using the output probabilities of a large model. \cite{busbridge2025distillation} build scaling laws to determine the optimal compute allocation between student and the teacher.\cite{peng2025pre} investigate the effect of different design decisions in token to token distillation. Beyond language modeling \cite{beyer2022knowledge} proposed the ``patient and consistent teacher'' hypothesis for ResNet-50 models \citep{he2016deep} trained on Imagenet \citep{5206848}. In this work we study pretraining distillation for both token and byte models. \\

\noindent\textbf{Token to Bytes transfer.} \cite{minixhofer2025bolmo} present an approach for adapting pretrained token-level models to operate on raw bytes. \cite{minixhofer2026universal} and \cite{bao2026distilling} propose methods for distilling token models into byte models in the fine-tuning stage. \cite{hayase2025sampling}  and \cite{phan2024exact} study the exact BPE to bytes transfer to primarily tackle the \texttt{prompt boundary problem} where the prompt, when ends in the middle of a BPE token, results into distorted next token distributions at inference time. These methods can be used to convert token logits to byte logits. However they remain computationally expensive and require multiple inference passes over the teacher to obtain exact conversions. We circumvent this problem by simply marginalizing the prefix-matched token distributions in \MarginalizeIt method and introducing an additional \texttt{<eot>} token in the vocabulary and the sequences during training to preserve the distribution in the \EndOfToken method. \\

\noindent \textbf{Byte Level Models.} \noindent\cite{AlRfou2018CharacterLevelLM}, \cite{choe2019bridging},  \cite{el2020characterbert} laid the groundwork for byte level language modeling.  \cite{clark2022canine} introduce CANINE, an architecture with downsampling, followed by a deep transformer stack and upsampling layers for efficient character level modeling.  \cite{xue2022byt5} train an encoder-decoder model operating on UTF-8 bytes and show that byte level transformers are data efficient learners, though take longer to train and are much expensive during inference time. \cite{yu2023megabyte} introduce the concept of patching to enable sub-quadratic self attention. \cite{nawrot-etal-2023-efficient} propose a hierarchical architecture for byte level modeling jointly performs language modeling and token segmentation. \cite{pagnoni2025byte} propose a new architecture with a local encoder and decoder that operates on patches of raw bytes. \cite{hwang2026dynamic} propose dynamic chunking for end to end byte level sequence modeling. Very recently EvaByte \cite{evabyte} trained byte level transformers with various tricks like multi-byte prediction and linearized attention. 
We study three decoder-only transformer variants \citep{vaswani2017attention} with vocabulary sizes of 128256 ( \BPETransformer), 260 ( \ByteTransformer), and 261 ( \EndOfTokenTransformer). They differ in both parameter count (from the embedding layers) and FLOPs (by varying sequence lengths). By preserving the teacher distribution exactly while spending \twentyfivepercent additional compute on a fixed amount of data, the \EndOfTokenTransformer model asymptotically outperforms the distilled \BPETransformer. We also note that \EndOfTokenTransformer model despite having fewer parameters can asymptotically surpass \BPETransformer on downstream tasks.\\

 \section{Discussion} \label{discussion}

\textbf{Marginalize-It vs End Of Token.} Our findings show that \MarginalizeIt, despite being an approximation asymptotically outperforms the \bpe distillation baseline. \EndOfToken offers two advantages: 1) it preserves the teacher distribution exactly, and 2) it improves asymptotic performance of the supervised \EndOfTokenTransformer baseline itself at the cost of an extra \twentyfivepercent compute per unit of data. Due to these two properties, \EndOfToken distillation asymptotically outperforms all of the other five models on downstream tasks. \\

\noindent\textbf{The role of \texttt{<eot>} tokens on power laws.} Our work shows that simply adding \texttt{<eot>} token every $\approx 4.5$ bytes on average affects the power law coefficients \ref{fig:bpb_vs_flops_4e3}.  We also observe that supervised  \EndOfTokenTransformer yields better asymptotic downstream performance compared to the \ByteTransformer baseline. This finding opens many exciting avenues for future work.\\

\noindent \textbf{Inference Costs.} While our scaling laws show that, distilling Llama 3-8B \citep{grattafiori2024llama} model into \EndOfTokenTransformer models with 1.28B total parameters is asymptotically better than distilling into \BPETransformer models with 1.81B total parameters, \EndOfTokenTransformer models remain significantly expensive at inference time. Comparing the asymptotic performance of a larger \BPETransformer model against that of a smaller \EndOfTokenTransformer transformers with matched inference cost could reveal whether distilled \EndOfTokenTransformer transformers are also efficient at inference time; this is beyond the scope of this study. \\

\noindent \textbf{BPB vs.\ Benchmark Performance Discrepancy.} 
Our observations reveal that BPB is not a metric that enables fair comparison across models with different training objectives and tokenization schemes. Many other works observe the same phenomenon for different optimizers \citep{zhang2026double}, data distributions \citep{gadre2024language}, long-context setting, 
\citep{fang2025wrong}, byte modeling \citep{lee2026efficiency}, and in theoretical analyses \citep{liu2023same, velivckovic2026perplexity}. We identify additional settings in which both the training objective (Cross Entropy Supervised vs Distillation) and the tokenization scheme (\token, \bytes, \bytesweot) can alter the downstream-task-performance vs.\ validation-BPB scaling trajectories, implying that a model with a lower BPB may not be necessarily a better model. \\

\subsection{Future work}
Our scaling study focuses on dense transformer models. It would be interesting to explore how introducing a third axis of sparsity, by training Mixture-of-Experts models affects these power laws. Next, our study examines scaling trends across varying data budgets with approximately fixed model size of 1.28 billion layer parameters, ranging up to 1 trillion bytes. We also note that some of our larger training runs epoch. Future work should explore training with larger model sizes and more data.
Next we report our findings based on eight benchmarks spanning three categories: Multiple Choice Question Answering, Language Generation, and Machine Translation. Future work should explore evaluating models on more benchmarks. Lastly, understanding the training dynamics of distilling large byte transformers into smaller ones remains an open question.

\section*{Acknowledgments}
We thank Jonathan Hayase, Alisa Liu, and Benjamin Minixhofer for thoughtful discussions and feedback on this work.

\newpage
\tableofcontents
\newpage
\appendix
\section{Experimental Setup Details} \label{experimental_setup}

In this section we provide additional details on the design decisions for our experiments. 

\subsection{Model Architecture and FLOP Calculations.}  To train both the token and byte student models, we use the transformer \citep{vaswani2017attention} architecture with 1 billion parameters. We use SwiGLU \citep{shazeer2020glu} activation function. \textcolor{PineGreen}{} The differences in the number of parameters and training FLOPs arise from the differences in the vocabulary size -- byte models use fewer parameters and FLOPs per byte. 
 
We use Llama3-8B \citep{grattafiori2024llama} as our teacher model. \\

\begin{table}[H]
\centering
\scalebox{0.8}{
\begin{tabular}{l l r r r}
\toprule
Property & Formula & \BPETransformer \label{bpe1b} & \ByteTransformer \label{byte1b} & \EndOfTokenTransformer \label{byteot1b} \\
\midrule
Model Dimension & $d$ & 2048 & 2048 & 2048 \\
FFN dimension & $ d_{ffn} = \lceil 4 \cdot \frac{2}{3}\cdot \frac{d}{256}    \rceil \cdot 256  $  & 5632 & 5632 & 5632 \\
Number of Layers & $n$ & 25 & 25 & 25 \\
Number of Heads & $n_{heads}$ & 16 & 16 & 16 \\
Vocabulary size & $V$ & 128256 & 260 & 261 \\

\midrule
Attention Parameters per Layer & $N_{attn} = 4 \cdot d^{2} $ & 16777216 & 16777216 & 16777216 \\
FFN Parameters per Layer & $ N_{ffn} =  3 \cdot d \cdot d_{ffn}$ & 34603008 & 34603008 & 34603008 \\
RMSNorm Parameters per Layer & $N_{norm} = 2\cdot d$& 4096 & 4096 & 4096 \\
Total Layer Parameters &  $N_{layers} = n \cdot (N_{attn} + N_{ffn} + N_{norm}) $  & 1284608000 & 1284608000 & 1284608000 \\
Pre Output Norm Parameters & $N_{preoutputnorm} = d$ & 2048 & 2048 & 2048 \\
Total Non-Embedding Parameters & $N_{nonembed} = N_{layers} + N_{preoutputnorm} + d \cdot V   $ & 1547278336 & 1285142528 & 1285144576 \\
Total Parameters & $N_{total} = N_{nonembed} + d \cdot V  $ & 1809946624 & 1285675008 & 1285679104 \\

\midrule

Sequence Length & $l$ & 2048 & 9216 & 11264 \\

Self Attention FLOPs &  $ F_{{attn}} = 4 \cdot n \cdot (d // n_{heads}) \cdot n_{\text{heads}} \cdot \frac{l+1}{2}$ & 209817600 & 943820800 & 1153536000 \\
Self Attention QKVO FLOPs &  $ F_{{qkvo}} = (2 + 2) \cdot 2 \cdot n \cdot d^2$ & 838860800 & 838860800 & 838860800 \\
Feed Forward FLOPs & $ F_{{ffn}} =  2 \cdot n \cdot 2 \cdot d \cdot d_{ffn}$ & 1153433600 & 1153433600 & 1153433600 \\
De Embedding FLOPs & $ F_{{deemb}} = 2 \cdot d \cdot |V|$ & 525336576 & 1064960 & 1069056 \\
Total FLOPs per unit & $ F_{total} =  3 \cdot (F_{{attn}} + F_{{qkvo}} + F_{{ffn}} + F_{{deemb}})$ & 8182345728 & 8811540480 & 9440698368 \\
\bottomrule
\end{tabular}
}

\caption{Model Architecture Comparison}
\label{tab:model_architecture}
\end{table}

\subsection{Pretraining Dataset, Tokenization, and Data Scales } We use the Llama-2 \citep{touvron2023llama} pretraining mixture for all of our experiments.  We tokenize the data using two tokenizers: BPE-Tiktoken (Llama3-8B) tokenizer and the byte tokenizer. Our byte tokenizer uses a vocabulary size of 260: 256 UTF-8 bytes and 4 special tokens. We train multiple 1B models with varying D where D is the number of tokens/byte units. \\

\subsection{Loss Function and Optimization Settings} \label{loss_function} A typical knowledge distillation setup involves two components: $\alpha {L_{KL}}$, a loss term computed using the teacher's logits and $(1-\alpha){L_{CE}}$, the next token prediction loss computed against the ground truth data. When $\alpha = 0$, the objective reduces to supervised  training (Cross-Entropy loss with respect to data) and when $\alpha = 1$, the model is trained purely via teacher distillation. We use  equation \ref{eq:loss} as our loss function and use $\alpha$, the mixing coefficient to control our supervised and distillation runs.

\begin{equation}
{L}_{\text{KD}} = \alpha \cdot {L}_{\text{KL}} + (1 - \alpha) \cdot {L}_{\text{CE}}
\label{eq:loss}
\end{equation}

\begin{equation}
L_{\text{KL}} = \frac{1}{n}\sum_{t=1}^{n} \sum_{c} p_T(c \mid x_{<t}) \log \frac{p_T(c \mid x_{<t})}{p_S(c \mid x_{<t})}
\end{equation}

\begin{equation}
L_{\text{CE}} = -\frac{1}{n}\sum_{t=1}^{n} \sum_{c} y_{t,c} \log p_S(c \mid x_{<t})
\end{equation}

\noindent where:
\begin{itemize}
    \item $t$ is the position index in the sequence
    \item $n$ is the total number of tokens in the sequence
    \item $c$ is the token index over the vocabulary ${V}$
    \item $y_{t,c} \in \{0,1\}$ is a one-hot indicator that the ground-truth token at position $t$ is $c$
    \item $p_T(c \mid x_{<t})$ is the teacher model's predicted probability distribution over the next token given all preceding tokens $x_{<t}$
    \item $p_S(c \mid x_{<t})$ is the student model's predicted probability distribution over the next token given all preceding tokens $x_{<t}$
    \item $\alpha \in [0,1]$ is a weighting factor balancing the KL divergence and cross-entropy losses
\end{itemize}

\noindent We train our models using the AdamW\citep{loshchilov2017decoupled} optimizer with $\beta_{1} = 0.9 $ and $\beta_{2} = 0.95 $ and a cosine learning rate schedule. We set warmup to 10\% of the maximum number of steps for each data scale use and three peak learning rates $\in$ \textcolor{PineGreen}\{1e-3, 4e-3, 8e-3\}. We use independent weight decay \citep{kosson2025weight} of 1e-4 and a grad clip norm of 1.0.    \\

\subsection{Sequence Length and Global Batch Size.} We set sequence length to 2048 for our \BPETransformer models. We train models using 64 H200s per training run and use a global batch size is 0.5M tokens. For training byte transformers, our initial explorations with the exact same context and data (8K $\times$ 4 $\times$ 64 = 2M bytes) per optimization step resulted in loss spikes. To stabilize the runs we use qk-norm and set the sequence length to 9216 for \ByteTransformer and  11264 for \EndOfTokenTransformer, and maintain the global batch size as that of the token models. If the byte sequence formed by the 2048 tokens exceeds the byte sequence length, we truncate it; if the sequence is shorter, we pad it with padding tokens to preserve the context length. We use 64 H200s for each \ByteTransformer and \EndOfTokenTransformer as well, maintaining the global batch size of $\approx$ 0.5M byte units as that of the token models. Hence, overall our byte models use much less data compared to token based models for the same compute budget. \\

\subsection{Bits Per Byte Calculation.} We use BPB (Bits per Byte) (equation ~\ref{eq:bpb}), a smooth tokenizer independent metric on a held out validation set to plot the scaling curves. The BPB is defined as the total cross-entropy loss $\mathcal{L}_{\text{CE}}$ summed over the validation data $x$ divided by the total number of bytes $B$ in $x$ and scaled by a constant. Intuitively, it captures the average number of bits a model uses to represent the data. 

\begin{equation}
\text{BPB} = \frac{{L}_{\text{CE}}(x)\cdot \log_2 e}{B}
\label{eq:bpb}
\end{equation}

\noindent Finally, We use Llama 3-8B \citep{grattafiori2024llama} and Llama3.2-1B \citep{meta2024llama32} models as reference points to compare with our token and byte asymptotes. Since these models are overtrained on several trillion tokens, they simulate the infinite data regimes. \\


\subsection{Storage Cost Calculations of Token and Byte Logits}\label{storage_cost_appendix}

The biggest pain point of using offline distillation to train small language models is the storage cost of the teacher logits. Each BPE token of a Llama3-8B model, for instance, produces 128256 logit values.  A billion parameter model overtrained to 2T tokens, for instance, would require and $2T \times 4 \times 128256 $ bytes = 1.026048 Exabytes assuming \texttt{float32} datatype. To circumvent this storage problem, researchers store only the \textit{top-k} logits per token, where $k$ typically is of the order of several hundreds. On the contrary, byte transformers use a fixed vocabulary of only $\approx$ 260 bytes, trading vocabulary size for increased sequence length. Our controlled study is designed to characterize this tradeoff. \\

\begin{table*}[th!]
\centering
\caption{Storage cost analysis for top-k logits per token. We solve for k and found it to be $\approx$ 600} 
\vspace{6pt}
\label{tab:storage_cost_analysis}
\scalebox{0.9}{
\begin{tabular}{ l l c c c c } 
    \toprule
    Token type & Data Attribute & Datatype & Bytes per value & \# Values  & Total Cost \\
    \midrule	

    \multirow{3}{*}{BPE-tiktoken} & Token Logit values & \texttt{float32} &4& $k$ & \multirow{3}{*}{   $ 4k + 4k +4 $       }  \\
                               & Token Logit indices & \texttt{int32}  &4 & $k$ \\
                               & Tokens & \texttt{int32} &4 & $1$ \\
    \midrule	

    \multirow{2}{*}{Bytes} & Byte Logits & \texttt{float32} & 4 & $260 \times  4.5$ & \multirow{2}{*}{   $ 4680 + 9 $       }   \\
                           & Bytes & \texttt{int16}  & 2 & $ 4.5$  \\
    \midrule

\end{tabular}}
\end{table*}

\noindent In table \ref{tab:storage_cost_analysis} we detail the storage cost of storing one BPE token  and their \textit{top-k} logits. Note that for BPE tokens, we use the sparse representation of logits while for bytes logits we store the full distribution. By equating the total cost of BPE-tiktoken with bytes, we solve for $k = 4685/8$, which turns out to be 585.625 $\approx$ 600. Therefore for our controlled comparison we store 600 logits per BPE-tiktoken token and all of the byte logits i.e. 260 values for distillation.

\section{Additional Results using Validation Dataset 1}
In this section we provide additional results for the learning rates of 1e-3 an 8e-3 for validation dataset 1 (which is used for analysis in the paper).

\begin{figure}[H]
    \centering
    \includegraphics[width=0.8\textwidth]{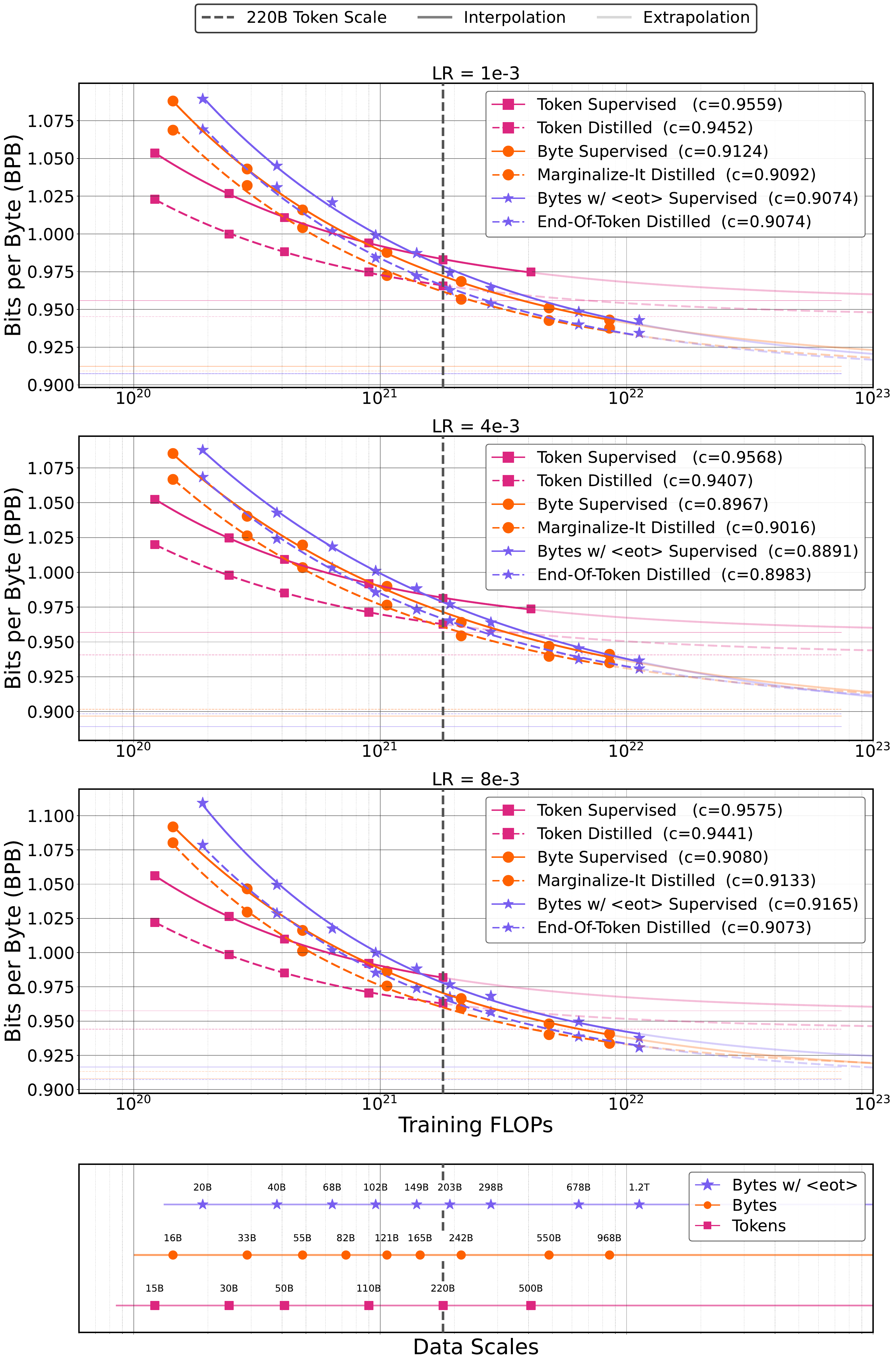}
    \caption{  The figure shows our scaling curves for different data scales and learning rates for a fixed parameter count of approximately 1 billion parameter transformer models(see Table \ref{tab:model_architecture} for details). The distilled models are trained using Llama 3-8B as a teacher model. The validation BPB of the teacher model is 0.85511. 
    \label{fig:bpb_vs_flops_lrwise}
    }
\end{figure}

\begin{table}[H]
\centering
\caption{Power-law equations $y = b \cdot x^{a} + c$ fitted to BPB vs.\ training FLOPs.}
\label{tab:scaling_laws}
\begin{tabular}{lll}
\toprule
LR & Scenario & Equation \\
\midrule
\multirow{6}{*}{$1 \times 10^{-3}$} & \BPESupervised & $y = 3.013 \times 10^{8} \cdot x^{-0.4724} + 0.9559$ \\
 & \BPEDistilled & $y = 5.677 \times 10^{8} \cdot x^{-0.4911} + 0.9452$ \\
 & \BytesSupervised & $y = 7.253 \times 10^{7} \cdot x^{-0.4274} + 0.9124$ \\
 & \MarginalizeDistilled & $y = 1.503 \times 10^{8} \cdot x^{-0.4449} + 0.9092$ \\
 & \BytesEotSupervised & $y = 6.329 \times 10^{7} \cdot x^{-0.4210} + 0.9074$ \\
 & \EndOfTokenDistilled & $y = 3.383 \times 10^{8} \cdot x^{-0.4593} + 0.9074$ \\
\midrule
\multirow{6}{*}{$4 \times 10^{-3}$} & \BPESupervised & $y = 1.075 \times 10^{9} \cdot x^{-0.5004} + 0.9568$ \\
 & \BPEDistilled & $y = 2.803 \times 10^{8} \cdot x^{-0.4754} + 0.9407$ \\
 & \BytesSupervised & $y = 4.467 \times 10^{6} \cdot x^{-0.3658} + 0.8967$ \\
 & \MarginalizeDistilled & $y = 2.706 \times 10^{7} \cdot x^{-0.4074} + 0.9016$ \\
 & \BytesEotSupervised & $y = 3.080 \times 10^{6} \cdot x^{-0.3546} + 0.8891$ \\
 & \EndOfTokenDistilled & $y = 2.734 \times 10^{7} \cdot x^{-0.4047} + 0.8983$ \\
\midrule
\multirow{6}{*}{$8 \times 10^{-3}$} & \BPESupervised & $y = 2.927 \times 10^{9} \cdot x^{-0.5214} + 0.9575$ \\
 & \BPEDistilled & $y = 3.692 \times 10^{9} \cdot x^{-0.5314} + 0.9441$ \\
 & \BytesSupervised & $y = 8.176 \times 10^{7} \cdot x^{-0.4289} + 0.9080$ \\
 & \MarginalizeDistilled & $y = 2.271 \times 10^{9} \cdot x^{-0.5028} + 0.9133$ \\
 & \BytesEotSupervised & $y = 3.627 \times 10^{9} \cdot x^{-0.5068} + 0.9165$ \\
 & \EndOfTokenDistilled & $y = 6.619 \times 10^{8} \cdot x^{-0.4729} + 0.9073$ \\
\bottomrule
\end{tabular}
\end{table}

\begin{table}[H]
\centering
\caption{
 We demonstrate that the amount of FLOPs spent on a fixed amount of data (one BPE token in this case) affects the asymptotic BPB. Overall: \BytesEotSupervised BPB $<$ \BytesSupervised BPB $<$ \BPESupervised BPB. and \EndOfTokenDistilled BPB $<$ \MarginalizeDistilled) BPB $<$  \BPEDistilled BPB.  We also note that for all learning rates, within the groups (based on training objective), the supervised and distilled models exhibit small differences in the asymptotic value.  The FLOPs/Unit values are obtained from Table \ref{tab:model_architecture}.
}

\vspace{6pt}
\label{tab:asymptotic_bpb_comparison_lrwise}
\begin{tabular}{llcccccc}
\toprule
\multirow{2}{*}{Method} & \multirow{2}{*}{FLOPs / Unit} & \multirow{2}{*}{Units} & \multirow{2}{*}{Total FLOPs } & \multicolumn{3}{c}{Asymptotic BPB} \\
\cmidrule(lr){5-7}
& & & Per Token & 1e-3 & 4e-3 & 8e-3 \\
\midrule

\BPESupervised           & $8.18 \times 10^9$ & 1.0 & $8.18 \times 10^9$  & 0.9559& 0.9568 & 0.9575\\

\BytesSupervised                        & $8.81 \times 10^9$ & 4.5 & $39.65 \times 10^9$   & 0.9124 & 0.8967 & 0.9080 \\

\BytesEotSupervised      & $9.44 \times 10^9$ & 5.5 & $51.92 \times 10^9$  & 0.9074& 0.8891& 0.9165 \\

\midrule

\BPEDistilled           & $8.18 \times 10^9$ & 1.0 & $8.18 \times 10^9$   &0.9452 & 0.9407 & 0.9441 \\

\MarginalizeDistilled              & $8.81 \times 10^9$ & 4.5 & $39.65 \times 10^9$  & 0.9092 & 0.9016 & 0.9133\\

\EndOfTokenDistilled      & $9.44 \times 10^9$ & 5.5 & $51.92 \times 10^9$  & 0.9074 & 0.8983 & 0.9073 \\
\bottomrule
\end{tabular}
\end{table}

\begin{table}[H]
  \centering
  \caption{Scaling-law goodness of fit. Each entry is the coefficient of determination $R^2 = 1 - \mathrm{SS_{res}}/\mathrm{SS_{tot}}$ for the fit $y = b \cdot x^{a} + c$  of BPB versus training FLOPs.}
  \label{tab:scaling_r2}
  \begin{tabular}{lccc}
    \toprule
    Scenario & $R^2$ (LR=1e-3) & $R^2$ (LR=4e-3) & $R^2$ (LR=8e-3) \\
    \midrule
    \BPESupervised & 0.9999 & 0.9999 & 1.0000 \\
    \BPEDistilled & 0.9996 & 1.0000 & 0.9997 \\
    \midrule
    \BytesSupervised & 0.9999 & 0.9980 & 0.9996 \\
    \MarginalizeDistilled & 0.9974 & 0.9989 & 0.9985 \\
    \midrule
   \BytesEotSupervised & 0.9982 & 0.9995 & 0.9984 \\
    \EndOfTokenDistilled & 0.9980 & 0.9990 & 0.9987 \\
    \bottomrule
  \end{tabular}
\end{table}

  \begin{figure}[H]
    \centering
    \includegraphics[width=0.9\textwidth]{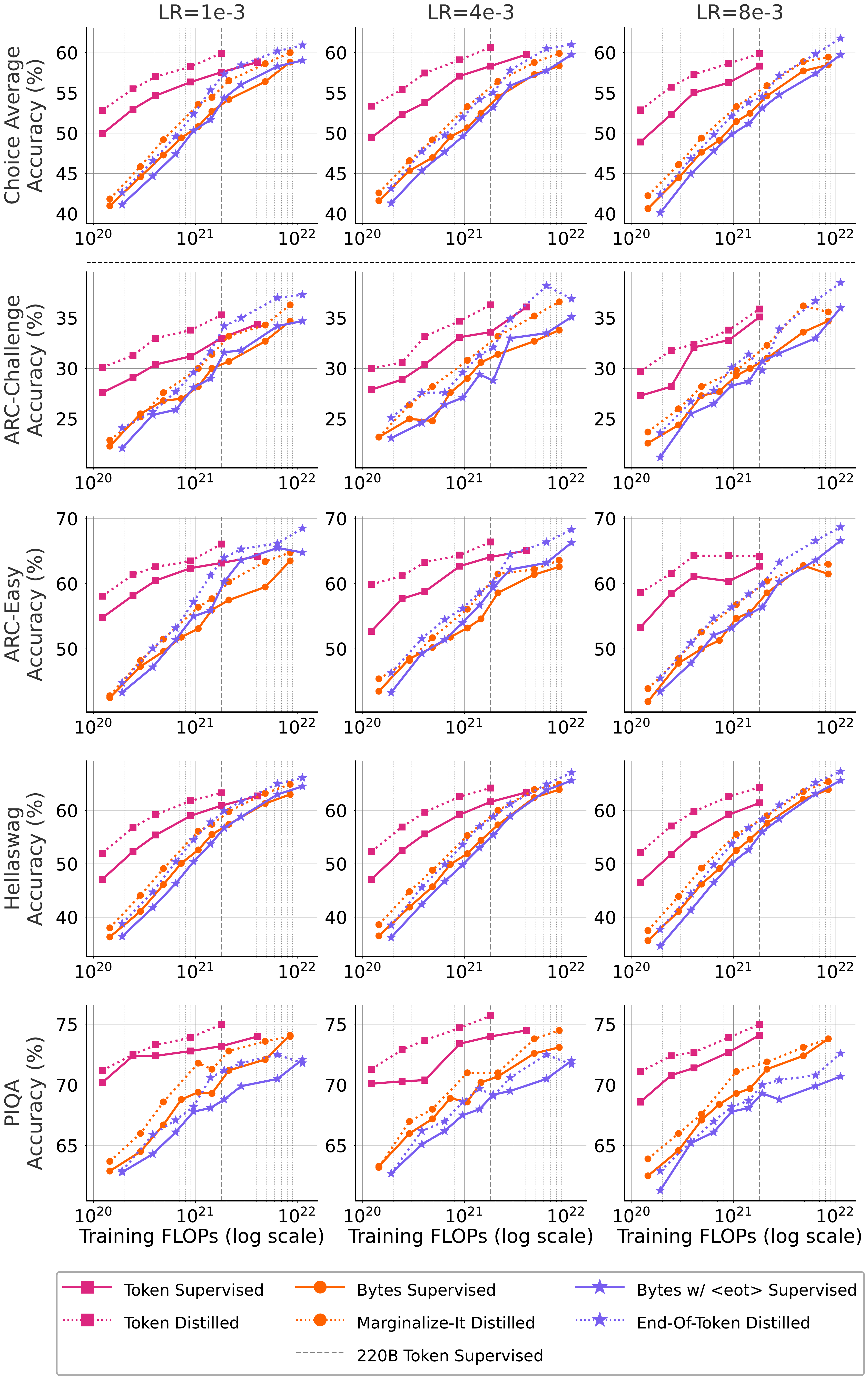}
    \caption{ We plot Accuracy vs training FLOP curves for all the six scenarios from section \ref{scalingaxes}. For all of the Multiple Choice Question Answering Benchmarks, we observe that the token models show better performance at low compute budgets and eventually plateau. Byte models, start worse yet improve at a much faster rate.   
    \label{fig:choice_benchmark_lrwise}
    }
\end{figure}

  \begin{figure}[H]
    \centering
    \includegraphics[width=1.0\textwidth]{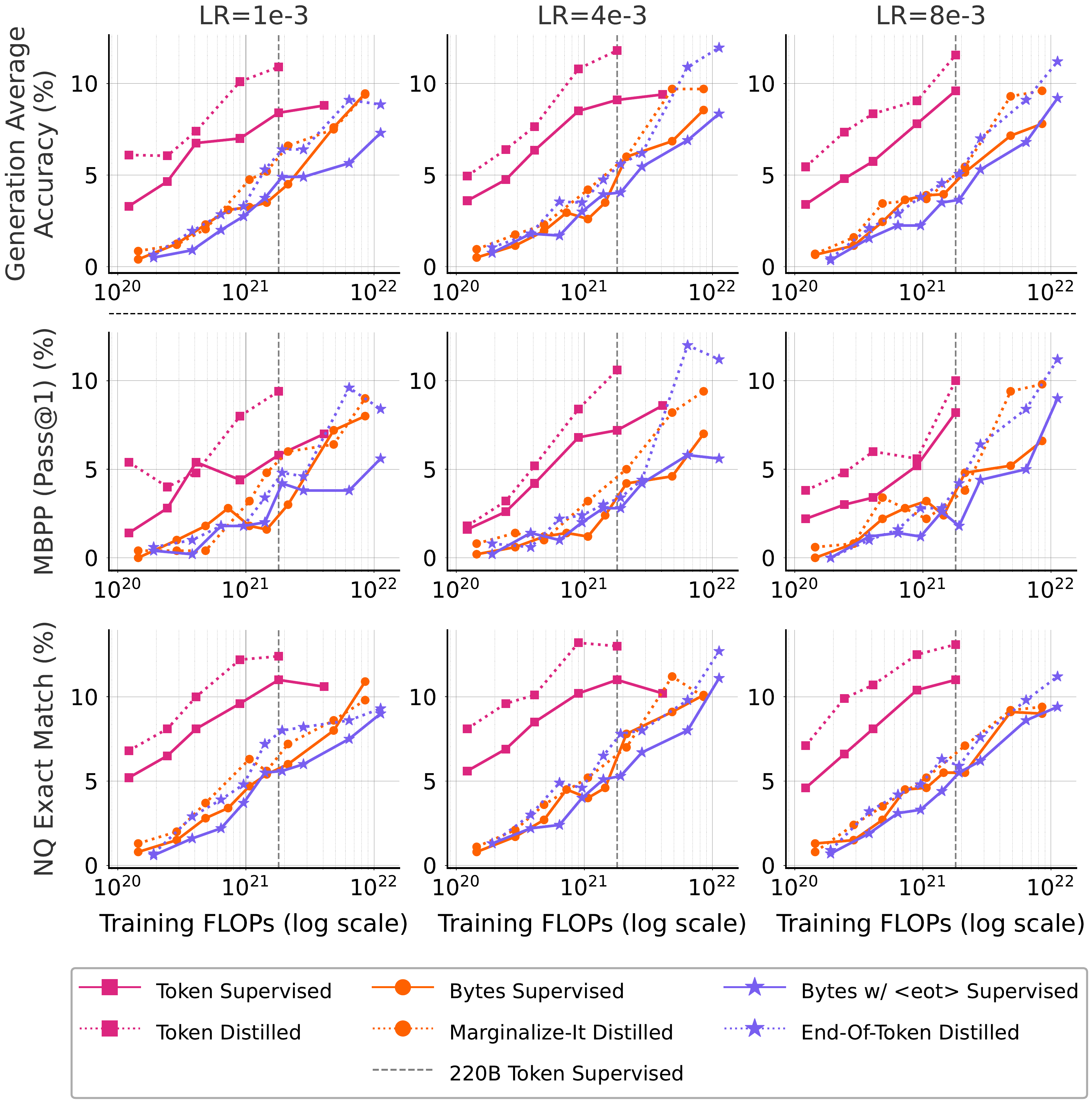}
    \caption{ We visualize $\%$ Accuracy vs training FLOPs curves for language generation tasks. On generative tasks, token models achieve better performance at lower compute budgets and appear to plateau with the training FLOPs. \ByteTransformer and \EndOfTokenTransformer models show a different scaling behavior, with much worse performance at a lower compute budget which continues to improve with FLOPs with no sign of saturation.
    \label{fig:gen_benchmark}
    }
\end{figure}

  \begin{figure}[H]
    \centering
    \includegraphics[width=1.00\textwidth]{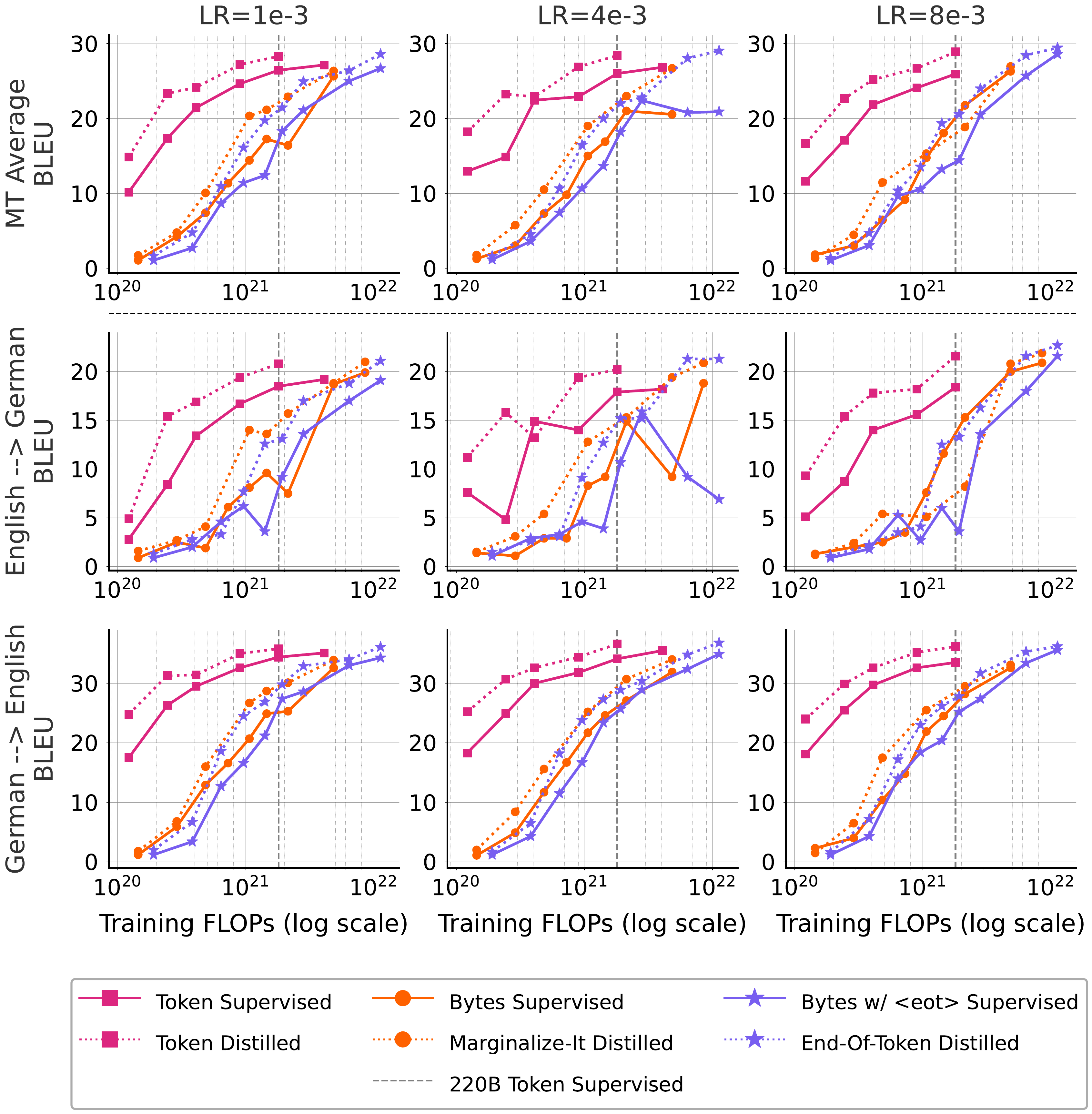}
    \caption{  We plot BLEU score for the translation tasks for different training budgets. We observe clearly different scaling behaviors for token and byte models. Token models can achieve significantly better performance at a smaller compute budgets. The BLEU score appears to saturate with overtraining. Byte models both with and without the \texttt{<eot>} demonstrate extremely low BLEU scores at lower compute budgets yet the rate of improvement with overtraining is better compared to token models. 
    \label{fig:mt_benchmark}
    }
\end{figure}

  \begin{figure}[H]
    \centering
    \includegraphics[width=0.88\textwidth]{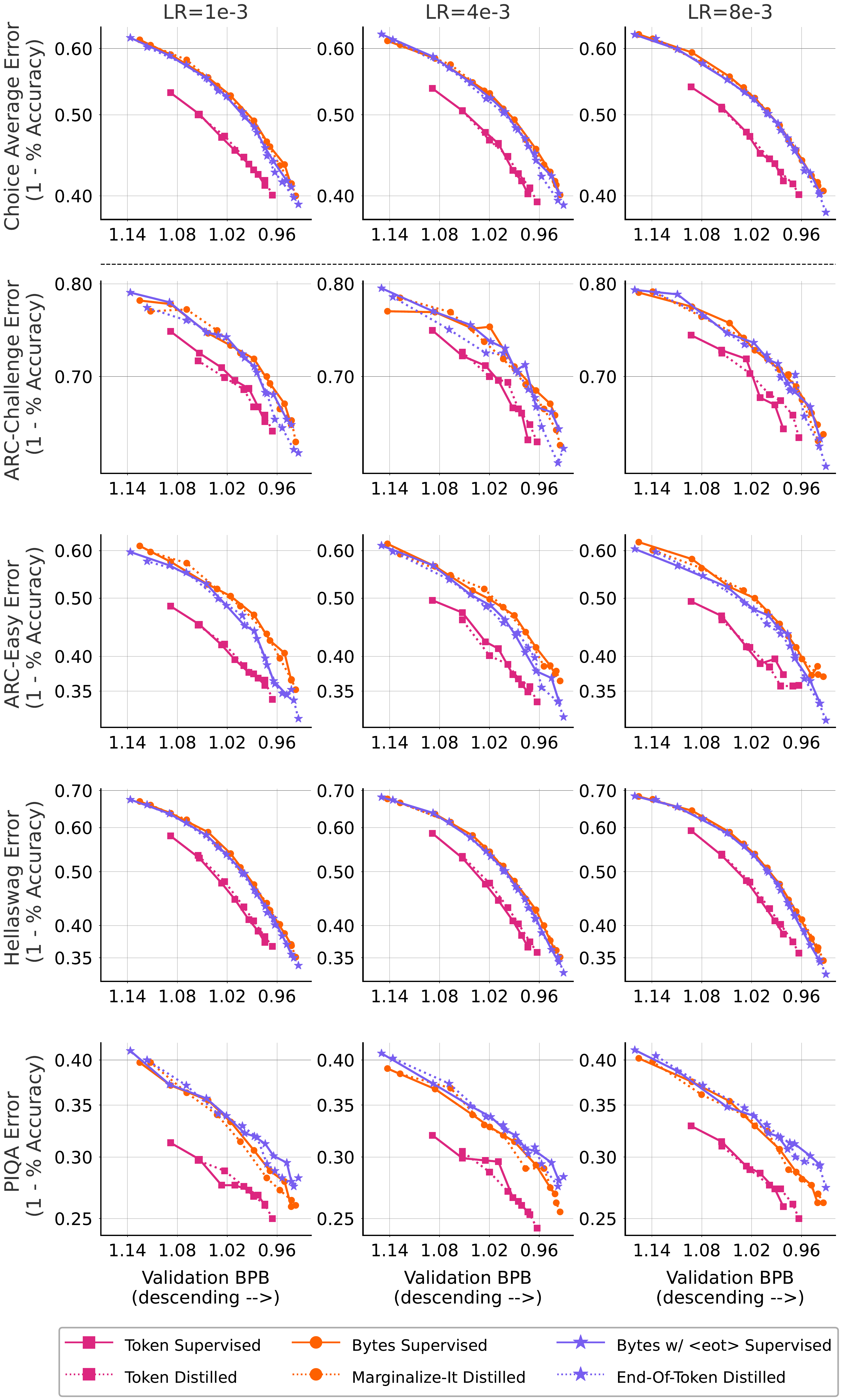}
    \caption{ We visualize Average-top-1 error vs the validation BPB curves for the six scenarios for different learning rates and Multiple Choice Question Answering benchmarks. All of these curves including the averaged metric appear to follow an exponential decay. HellaSwag demonstrates smoothest scaling behavior.   }
    \label{fig:choice_benchmark_error_vs_bpb_lrwise}
\end{figure}

  \begin{figure}[H]
    \centering
    \includegraphics[width=1.0\textwidth]{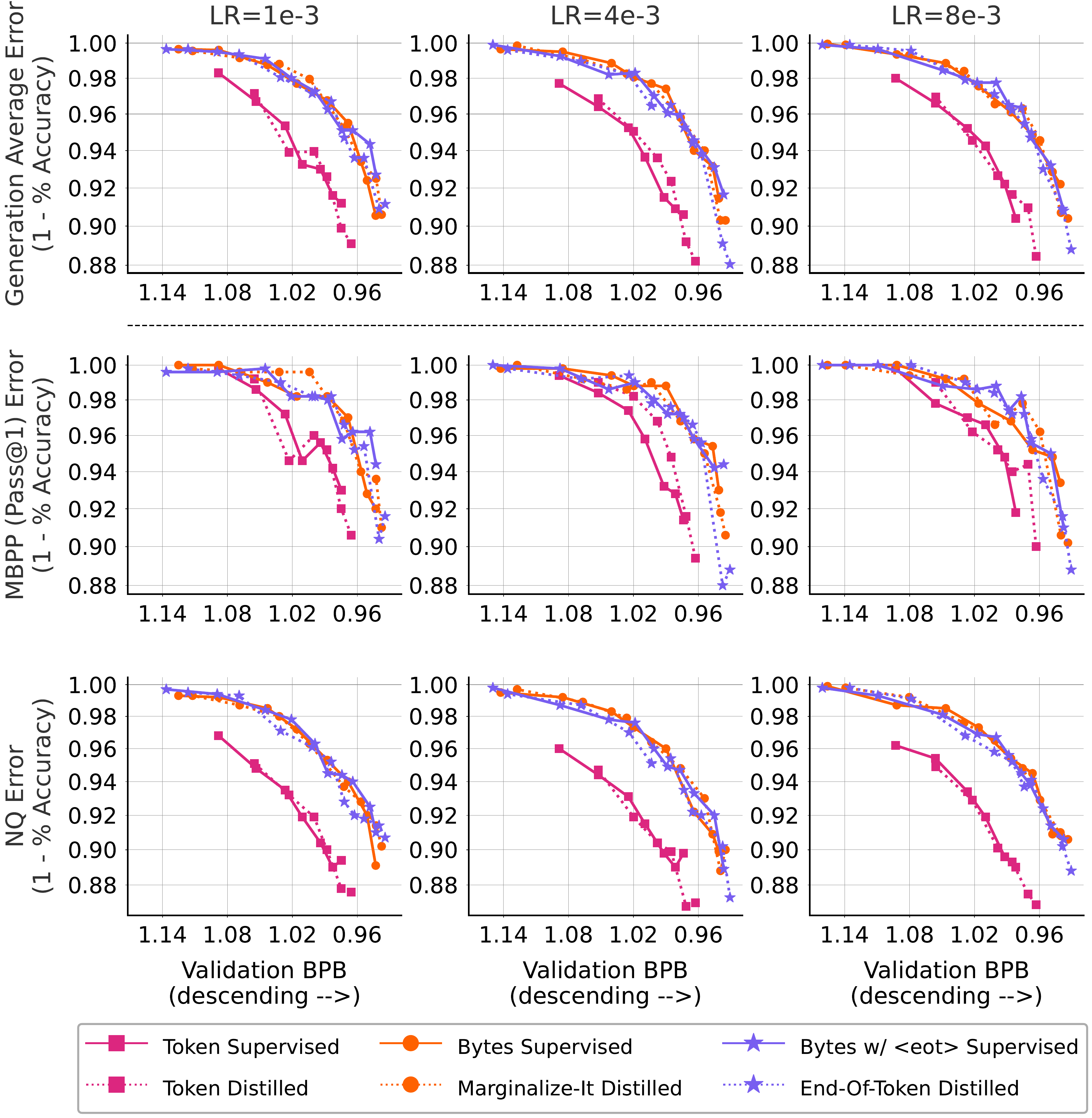}
    \caption{ We plot the Average-top-1 error vs BPB curves for the MBPP and the NQ benchmarks. We notice an exponential decay for all three scenarios, with significantly different trends for the token and byte models. MBPP shows noisier behavior compared to NQ. We deduce that the same validation Bits-Per-Byte imply different downstream performace for different training objectives(distillation vs cross-entropy) tokenization schemes(tokens vs bytes). Distilled models as well demonstrate slightly different scaling behavior compared to the supervised counterparts. }
    \label{fig:gen_benchmark_error_vs_bpb_lrwise}

\end{figure}

  \begin{figure}[H]
    \centering
    \includegraphics[width=1.0\textwidth]{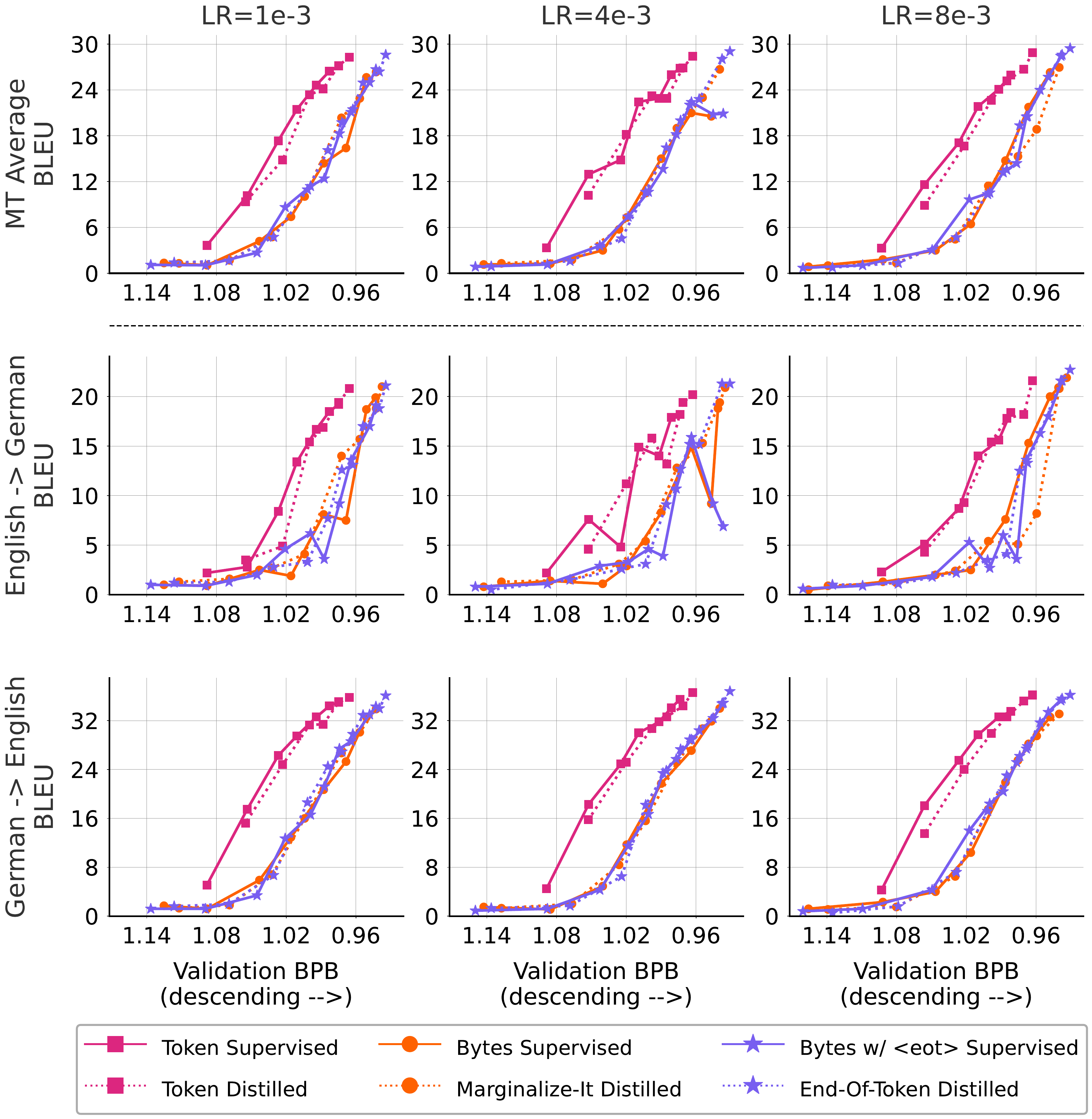}
    \caption{ We plot BLEU score vs Validation BPB performance for English to German translation, German to English translation, and on the averaged BLEU performance for the six scenarios for three learning rates. We notice the differences in the scaling behavior across tokenization schemes as well as the training objectives.  In some cases distillation appears to exhibit monotonic trend compared to the noisy supervised curves especially for English to German translation. }
    \label{fig:mt_benchmark_vs_bpb}

\end{figure}

  \begin{figure}[H]
    \centering
    \includegraphics[width=1.0\textwidth]{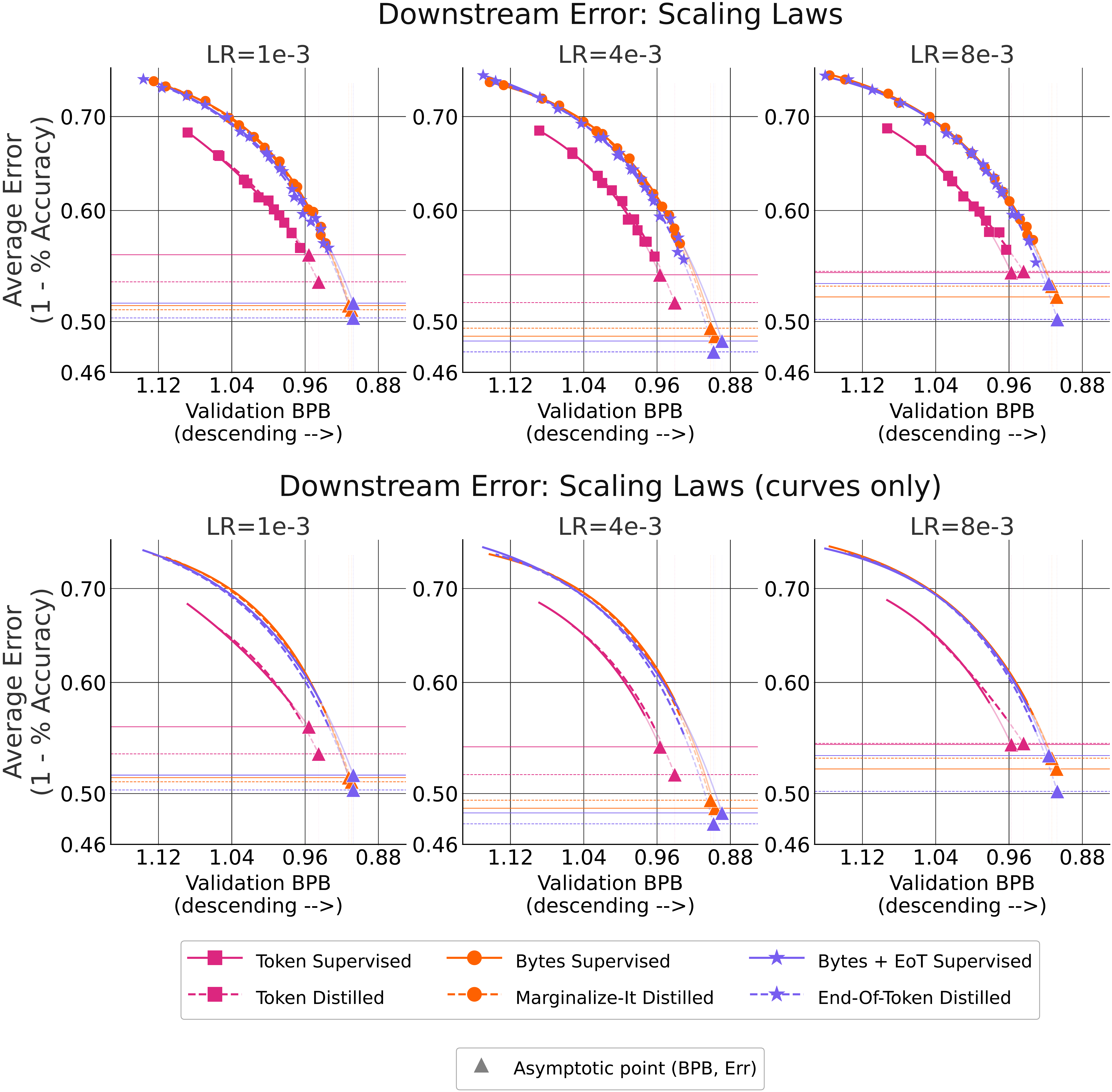}
    \caption{ Downstream Error Scaling laws }
    \label{fig:downstream_scaling_law_notes}

\end{figure}

\begin{table}[h]
  \centering
  \caption{Fitted downstream error scaling laws \DownstreamEquation (equation \ref{eq:downstream_error}) for each model variant and learning rate, where $y$ is the validation BPB.}
  \vspace{6pt}
  \label{tab:downstream_scaling_law_eqs}
  \scalebox{0.88}{
  \begin{tabular}{@{}llll@{}}
    \toprule
    \textbf{Model} & \multicolumn{1}{c}{LR$=1\times10^{-3}$} & \multicolumn{1}{c}{LR$=4\times10^{-3}$} & \multicolumn{1}{c}{LR$=8\times10^{-3}$} \\
    \midrule
    \BPESupervised & $0.9239 - 7.3780\,e^{-3.1423\,y}$ & $0.7501 - 934.3212\,e^{-8.7782\,y}$ & $0.7600 - 492.4601\,e^{-8.0652\,y}$ \\
    \BPEDistilled & $0.7430 - 398.4377\,e^{-7.9888\,y}$ & $0.7383 - 1292.0124\,e^{-9.2132\,y}$ & $1.0076 - 5.6152\,e^{-2.6394\,y}$ \\
    \midrule
    \BytesSupervised & $0.7726 - 1950.1906\,e^{-9.7822\,y}$ & $0.7686 - 1231.1241\,e^{-9.3532\,y}$ & $0.7872 - 367.2256\,e^{-7.9602\,y}$ \\
    \MarginalizeDistilled & $0.7805 - 958.5881\,e^{-8.9880\,y}$ & $0.7721 - 1137.3004\,e^{-9.2265\,y}$ & $0.7825 - 454.9215\,e^{-8.2080\,y}$ \\
    \midrule
   \BytesEotSupervised & $0.7947 - 267.8536\,e^{-7.5664\,y}$ & $0.7972 - 185.6303\,e^{-7.1822\,y}$ & $0.7776 - 650.0335\,e^{-8.5998\,y}$ \\
    \EndOfTokenDistilled & $0.7915 - 378.9535\,e^{-7.9129\,y}$ & $0.7776 - 827.6076\,e^{-8.8125\,y}$ & $0.7799 - 773.4793\,e^{-8.7411\,y}$ \\
    \bottomrule
  \end{tabular}}
\end{table}

\begin{table}[H]
  \centering
  \caption{Asymptotic validation BPB ($\mathrm{c}$) and the corresponding downstream average accuracy $a^\star = 1-\%z^\star$ where $z^\star$ is the Asymptotic Average top-1 error
  predicted by the scaling law equation: \ref{eq:downstream_error}, $z^\star = \epsilon - k \cdot e^{-\gamma c}$ , where $c$ is obtained from the power law equations in Table \ref{tab:downstream_scaling_law_eqs_4e3} for each model variant and learning rate.}
\vspace{6pt}
  
  \label{tab:asymptotic_perf}
  \begin{tabular}{@{}lcccccc@{}}
    \toprule
     & \multicolumn{2}{c}{LR$=1\times10^{-3}$} & \multicolumn{2}{c}{LR$=4\times10^{-3}$} & \multicolumn{2}{c}{LR$=8\times10^{-3}$} \\
    \cmidrule(lr){2-3} \cmidrule(lr){4-5} \cmidrule(lr){6-7}
    {Method} & $\mathrm{c}$ & $a^\star$ (\%) & $\mathrm{c}$ & $a^\star$ (\%) & $\mathrm{c}$ & $a^\star$ (\%) \\
    \midrule
    \BPESupervised & 0.9559 & 44.2 & 0.9568 & 46.0 & 0.9575 & 45.8 \\
    \BPEDistilled & 0.9452 & 46.6 & 0.9407 & 48.4 & 0.9441 & 45.7 \\
    \midrule
    \BytesSupervised & 0.9124 & 48.7 & 0.8967 & 51.2 & 0.9080 & 47.9 \\
    \MarginalizeDistilled & 0.9092 & 49.0 & 0.9016 & 50.5 & 0.9133 & 47.0 \\
    \midrule
    \BytesEotSupervised & 0.9074 & 48.5 & 0.8891 & 51.6 & 0.9165 & 46.8 \\
    \EndOfTokenDistilled & 0.9074 & 49.7 & 0.8983 & 52.4 & 0.9073 & 49.8 \\
    \bottomrule
  \end{tabular}
\end{table}

  \begin{figure}[H]
    \centering
    \includegraphics[width=1.0\textwidth]{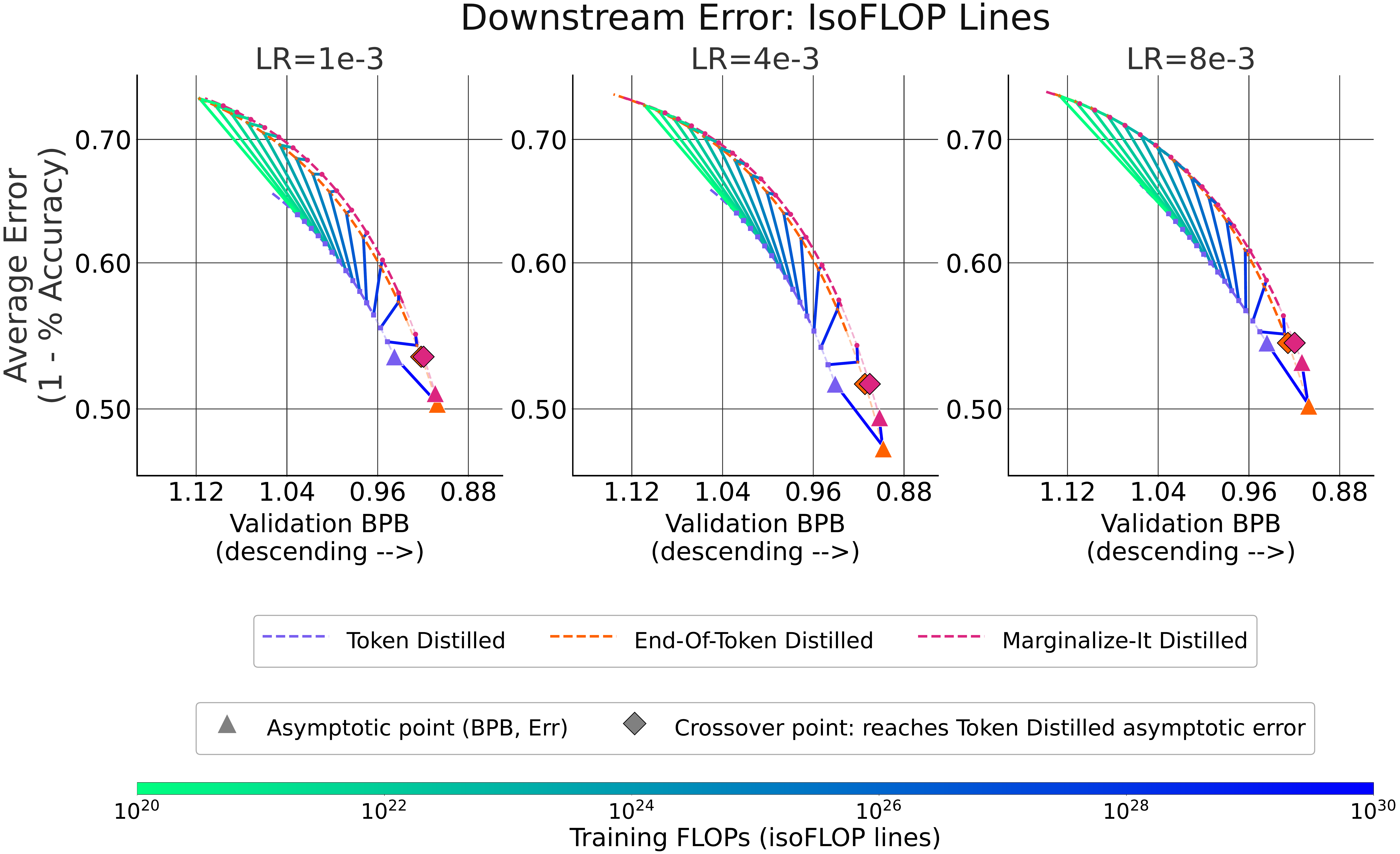}
    \caption{ Feather plots for three learning rates on validation dataset 1  }
    \label{fig:feather_plot_notes}

\end{figure}

\begin{table}[H]
  \centering
  \caption{Training compute at which \MarginalizeDistilled{} and \EndOfTokenDistilled{} match the asymptotic downstream performance of Token Distilled. We take Token Distilled\'s asymptotic BPB (the $c$ term of its fitted BPB--FLOPs power law; column \textbf{Token Dist. BPB}) and map it through Token Distilled\'s benchmark scaling law to obtain its asymptotic downstream accuracy (column \textbf{Accuracy}). For each byte method we then invert its own benchmark scaling law at that accuracy to get the BPB it must attain (column \textbf{Method BPB}), and invert its BPB--FLOPs power law at that BPB to get the required training compute (column \textbf{Training FLOPs}). The accuracy column is shared within each learning rate because both methods are evaluated against the same Token Distilled asymptote.}
  \label{tab:crossover_compute}
  \renewcommand{\arraystretch}{1.3}
  \resizebox{\textwidth}{!}{
  \begin{tabular}{cccccc}
    \toprule
    \textbf{LR} & \textbf{Method} & \textbf{Token Dist. BPB} & \textbf{Accuracy (\%)} & \textbf{Method BPB} & \textbf{Method Training FLOPs} \\
    \midrule
    \multirow{2}{*}{$10^{-3}$} & \MarginalizeDistilled & $0.9452$ & $46.6\%$ & $0.9195$ & $7.05 \times 10^{22}$ \\
     & \EndOfTokenDistilled & $0.9452$ & $46.6\%$ & $0.9216$ & $3.91 \times 10^{22}$ \\
    \cmidrule{1-6}
    \multirow{2}{*}{$4\times10^{-3}$} & \MarginalizeDistilled & $0.9407$ & $48.4\%$ & $0.9102$ & $2.04 \times 10^{23}$ \\
     & \EndOfTokenDistilled & $0.9407$ & $48.4\%$ & $0.9145$ & $6.33 \times 10^{22}$ \\
    \cmidrule{1-6}
    \multirow{2}{*}{$8\times10^{-3}$} & \MarginalizeDistilled & $0.9441$ & $45.7\%$ & $0.9197$ & $9.27 \times 10^{22}$ \\
     & \EndOfTokenDistilled & $0.9441$ & $45.7\%$ & $0.9256$ & $2.13 \times 10^{22}$ \\
    \bottomrule
  \end{tabular}
  }
\end{table}

\begin{table}[H]
  \centering
  \caption{Effective training-data volume at the compute budget required to achieve the \token distillation asymptotic performance from Table \ref{tab:asymptotic_perf_4e3}. Here $\mathrm{FPU}$ (\emph{FLOPs Per Unit}) is the training FLOPs consumed per sequence unit, where a unit is one token for \BPEDistilled and one byte for the byte-level methods; it is obtained from Table \ref{tab:model_architecture}. \textbf{Byte Units} $=\text{FLOPs}/\mathrm{FPU}_{\text{bytes}}$ is the number of byte units the method processes at its crossover compute. \textbf{Bytes} $=\text{Byte Units}\times\text{Multiplier}$ converts byte units into actual UTF-8 bytes of unique text, accounting for the fraction of each unit that is genuine content (Multiplier $=1$ for \MarginalizeDistilled; $=4.5/5.5$ for \EndOfTokenDistilled, whose sequences carry an extra \texttt{<eot>} per $4.5$ content bytes). \textbf{Tokens} $=\text{FLOPs}/\mathrm{FPU}_{\text{tokens}}$ is the token count \BPEDistilled would process at the same compute. \textbf{Token to Byte} $=\text{Tokens}\times4.5$ expresses that token count in bytes using the average of $4.5$ bytes per token. \textbf{Data Savings} $=\text{Token to Byte}/\text{Bytes}$ is how much less unique text the byte method needs to reach the same performance. To calculate storage savings we use our storage cost calculations from Table \ref{tab:storage_cost_analysis} where we equate the storage cost of one \token with 4.5 \bytes to obtain k (for top-k).  \textbf{Storage Savings} $=(\text{Tokens}\times4.5)/\text{Byte Units}$ compares the byte footprint of \BPEDistilled approach against the raw byte-unit count. }
  \label{tab:crossover_data_volume_lrwise}
  \renewcommand{\arraystretch}{1.3}
  \resizebox{\textwidth}{!}{
  \begin{tabular}{ccccccccccc}
    \toprule
    \textbf{LR} & \textbf{Method} & \textbf{Training FLOPs} & \textbf{FPU$_{\text{bytes}}$} & \textbf{FPU$_{\text{tokens}}$} & \textbf{Byte Units} & \textbf{Bytes} & \textbf{Tokens} & \textbf{Token to Byte} & \textbf{Data Savings} & \textbf{Storage Savings} \\
    \midrule
    \multirow{2}{*}{$10^{-3}$} & \MarginalizeDistilled & $7.05 \times 10^{22}$ & $8.81 \times 10^{9}$ & $8.18 \times 10^{9}$ & $8.00 \times 10^{12}$ & $8.00 \times 10^{12}$ & $8.61 \times 10^{12}$ & $3.87 \times 10^{13}$ & $4.85\times$ & $4.85\times$ \\
     & \EndOfTokenDistilled & $3.91 \times 10^{22}$ & $9.44 \times 10^{9}$ & $8.18 \times 10^{9}$ & $4.14 \times 10^{12}$ & $3.39 \times 10^{12}$ & $4.77 \times 10^{12}$ & $2.15 \times 10^{13}$ & $6.35\times$ & $5.19\times$ \\
    \cmidrule{1-11}
    \multirow{2}{*}{$4\times10^{-3}$} & \MarginalizeDistilled & $2.04 \times 10^{23}$ & $8.81 \times 10^{9}$ & $8.18 \times 10^{9}$ & $2.32 \times 10^{13}$ & $2.32 \times 10^{13}$ & $2.50 \times 10^{13}$ & $1.12 \times 10^{14}$ & $4.85\times$ & $4.85\times$ \\
     & \EndOfTokenDistilled & $6.33 \times 10^{22}$ & $9.44 \times 10^{9}$ & $8.18 \times 10^{9}$ & $6.70 \times 10^{12}$ & $5.48 \times 10^{12}$ & $7.73 \times 10^{12}$ & $3.48 \times 10^{13}$ & $6.35\times$ & $5.19\times$ \\
    \cmidrule{1-11}
    \multirow{2}{*}{$8\times10^{-3}$} & \MarginalizeDistilled & $9.27 \times 10^{22}$ & $8.81 \times 10^{9}$ & $8.18 \times 10^{9}$ & $1.05 \times 10^{13}$ & $1.05 \times 10^{13}$ & $1.13 \times 10^{13}$ & $5.10 \times 10^{13}$ & $4.85\times$ & $4.85\times$ \\
     & \EndOfTokenDistilled & $2.13 \times 10^{22}$ & $9.44 \times 10^{9}$ & $8.18 \times 10^{9}$ & $2.25 \times 10^{12}$ & $1.84 \times 10^{12}$ & $2.60 \times 10^{12}$ & $1.17 \times 10^{13}$ & $6.35\times$ & $5.19\times$ \\
    \bottomrule
  \end{tabular}
  }
\end{table}

\section{Comparisons with open-weight models: Additional details}

\begin{table}[H]
\centering
\caption{Downstream benchmark accuracy (\%) for the $4\text{e}{-}3$ learning-rate runs at their maximum available data scale, compared against the Llama and Gemma 1B baselines. Avg.\ / Asympt.\ Avg.\ is the mean over the six benchmarks.}
\label{tab:llama_gemma_bench_comparison}
\scalebox{0.8}{\begin{tabular}{l c cccccc c c}
\toprule
Model & Data Scale &HellaSwag & PIQA & ARC-Easy & ARC-Challenge & MBPP & NQ & Avg.  \\
\midrule

\texttt{Llama-3.2-1B} & 9T  & 64.3 & 75.3 & 66.1 & 36.6 & 27.0 & 5.7 & 45.9 &  \\
\texttt{Gemma-3-1B-pt} & 2T  & 62.1 & 74.7 & 72.2 & 38.4 & 9.0 & 9.5 & 44.3  \\
\texttt{Gemma 2b} & 3T &	71.44	&	78.07	&	72.18	&	41.89 &	28.0 &	10.53	&	50.3 \\

\midrule

\BPESupervised & 500B & 63.4 & 74.5 & 65.1 & 36.1 & 8.6 & 10.2 & 43.0 &  \\
\BPEDistilled & 220B & 64.2 & 75.7 & 66.4 & 36.3 & 10.6 & 13.0 & 44.4 &  \\
\BytesSupervised & 880B & 63.9 & 73.1 & 62.6 & 33.8 & 7.0 & 10.1 & 41.8 &  \\
\MarginalizeDistilled & 880B & 64.9 & 74.5 & 63.6 & 36.6 & 9.4 & 10.0 & 43.2 &  \\
\BytesEotSupervised & 880B & 65.6 & 72.0 & 66.3 & 35.1 & 5.6 & 11.1 & 42.6 &  \\
\EndOfTokenDistilled & 880B & 67.1 & 71.7 & 68.3 & 36.9 & 11.2 & 12.7 & 44.6 &  \\

\bottomrule
\end{tabular}}
\end{table}

\newpage

\section{On  Scaling Laws Sensitivity} 
\label{sensitivity}
We now characterize the sensitivity of downstream error scaling laws.  

\subsection{Leave One Out Error Analysis }
We analyze the Leave-One-Out sensitivity of the power laws.

  \begin{figure}[H]
    \centering
    \includegraphics[width=1.0\textwidth]{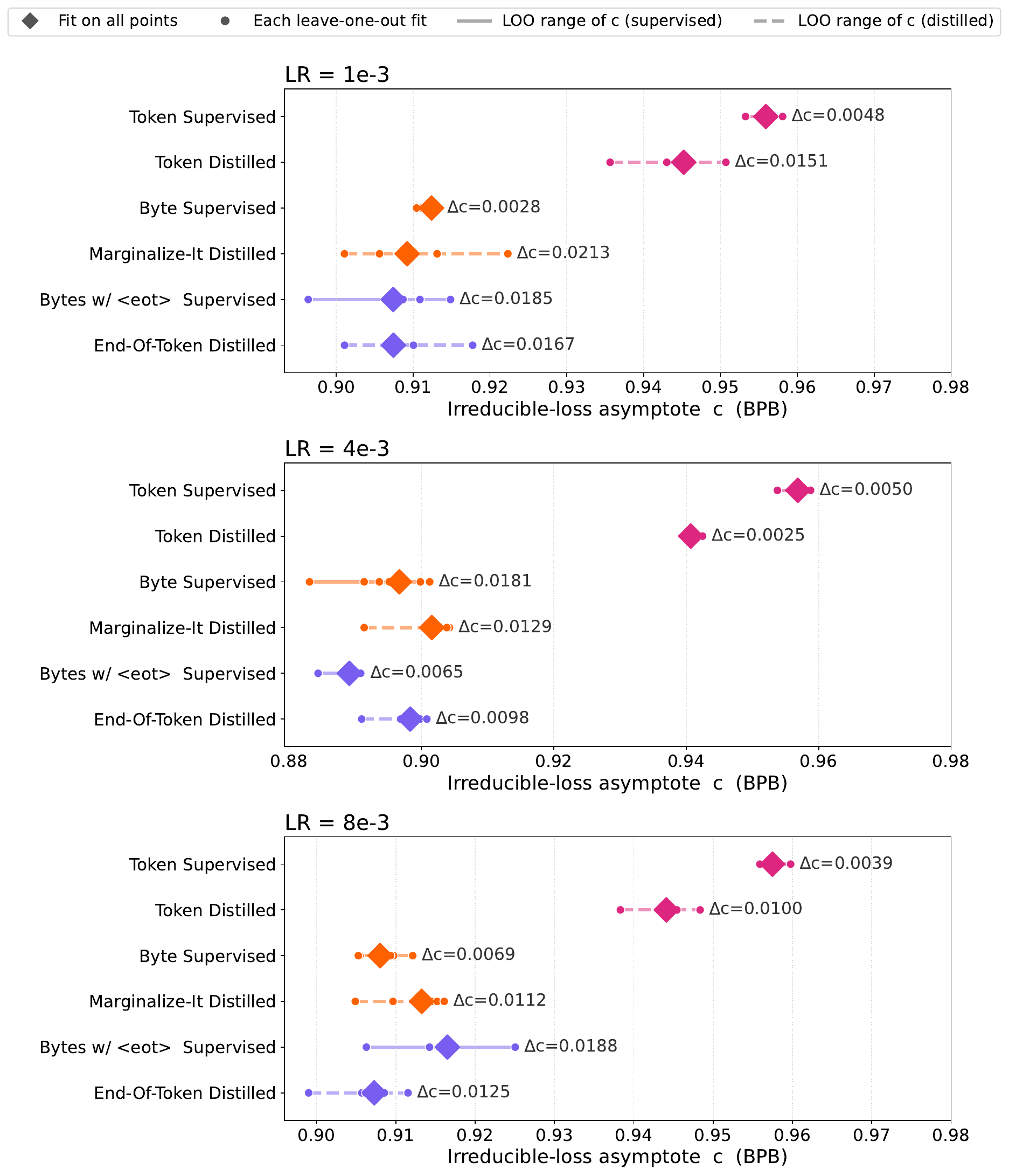}
    \caption{ We visualize the sensitivity of the scaling law fits via leave-one-out analysis. For each learning rate and for each experiment, we drop one of the datapoints and fit the power law using the rest and visualize the range of c. \token curves on average demonstrate little spread compared to \bytes and \bytesweot methods.  }
    \label{fig:loo_sensitivity_summary}

\end{figure}

\subsection{Residual Weighting Scheme Sensitivity Analysis }

  \begin{figure}[H]
    \centering
    \includegraphics[width=1.0\textwidth]{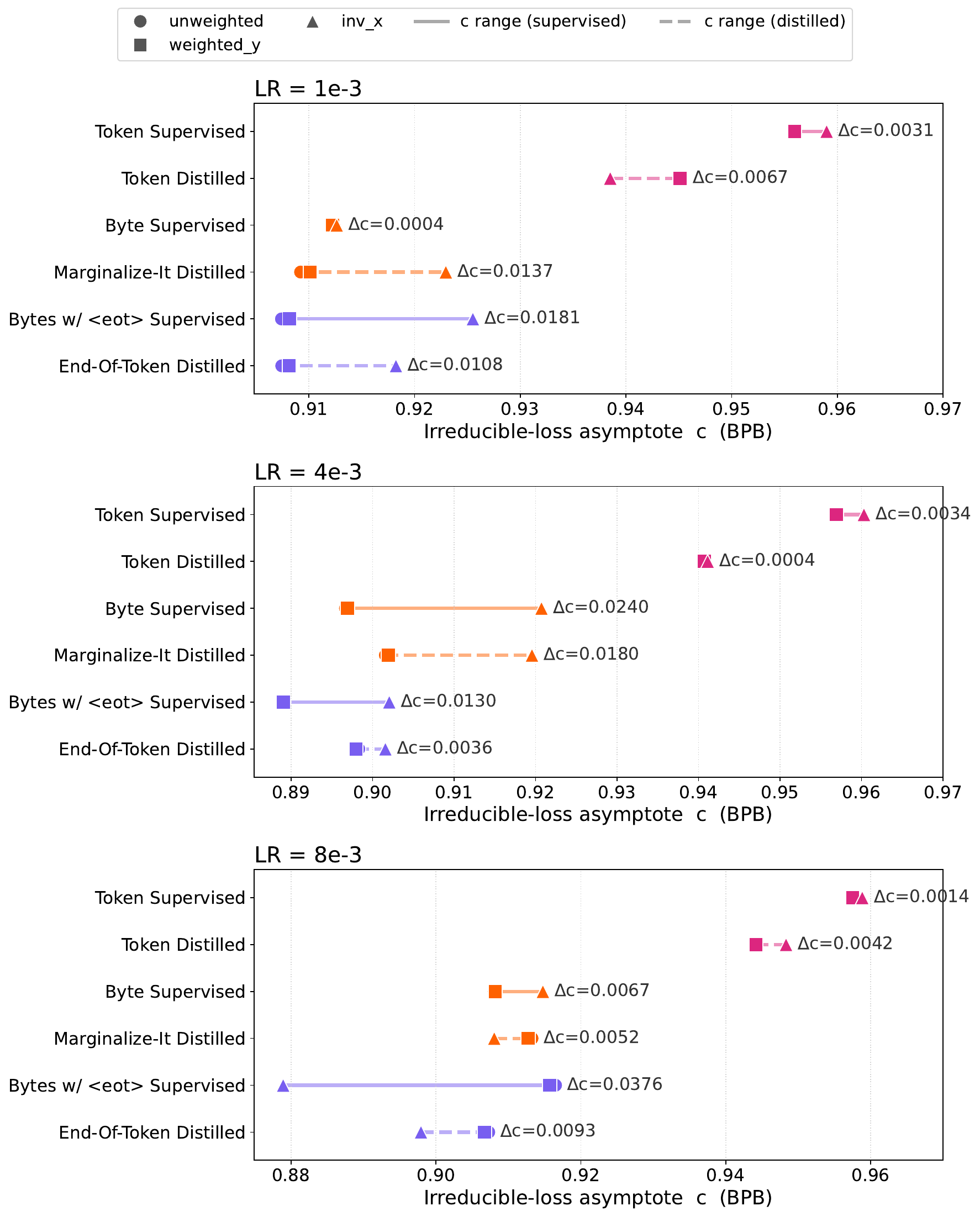}
    \caption{ We visualize the sensitivity of the residual weighting scheme on the power laws. 
Three weighting schemes, each a choice of the $\sigma$ vector in
\texttt{curve\_fit}'s objective $\sum_i\big((\hat{y}(x_i)-y_i)/\sigma_i\big)^2$:\newline
\textbullet \texttt{ unweighted}: ordinary least squares.\newline
\textbullet \texttt{ weighted y} : minimizes relative error $(\hat{y}-y)/y$, down-weighting high-BPB low-compute runs.\newline
\textbullet \texttt{ inverse x}: shrinks $\sigma$ for large-compute points so they dominate the extrapolation to high compute points. \newline 
 We use \texttt{unweighted} method for all of our plots. We note that other weighting schemes are more sensitive to \bytes and \bytesweot tokenizations. }  
    \label{fig:residual_weighting_scheme_sensitivity_sensitivity_summary}

\end{figure}

\subsection{Choice of the validation data}

\begin{figure}[H]
    \centering
    \includegraphics[width=1.0\textwidth]{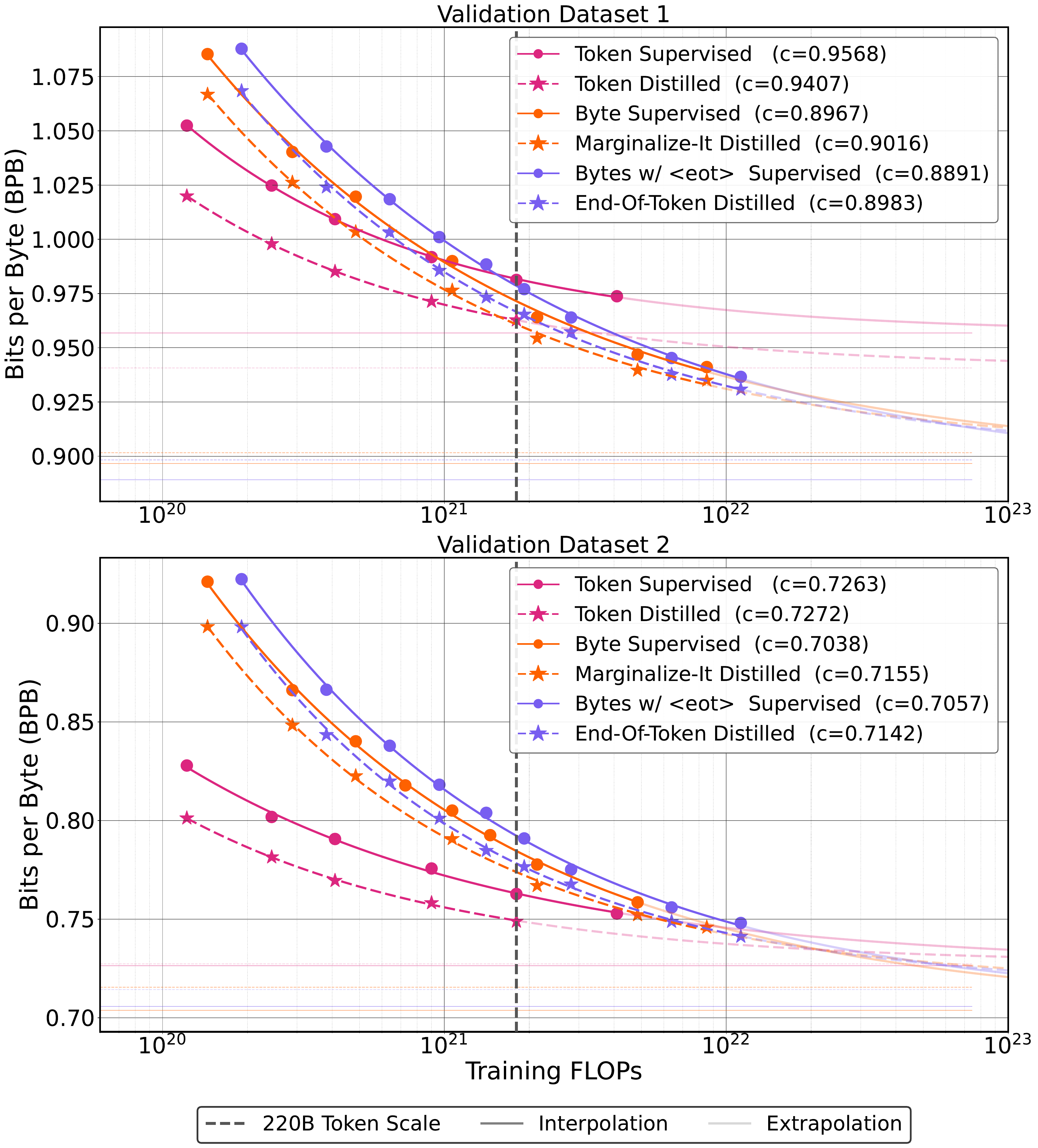}
    \caption{ We demonstrate how changes in the validation dataset can affect the validation BPB vs training FLOPs scaling plots by comparing power laws for LR=4e-3. At the 220B \BPESupervised FLOP budget, all of the other five models have a lower BPB for validation 1. On the other hand, for validation dataset 2, only the distilled token model has a lower BPB. The two datasets also have different asymptotic Bits-Per-Byte values for all six curves. }
    \label{fig:news_vs_notes}

\end{figure}

 We observe that the choice of the validation dataset can alter the cross-over points of the training curves and as a result the perception of a better model.  Consider for example in Figure \ref{fig:news_vs_notes}. In Validation Dataset 2, the token curves have not intersected the byte curves yet, while in Validation Dataset 1 they cross them quickly. The nature of the \FeatherPlot also change with changes in the validation data. Moreover, the amount of overtraining required to reliably predict the asymptotic behavior could also change with change in the validation dataset. Hence, taking performance on multiple validation sets into account, when fitting scaling laws might be helpful in predicting the final model performance.

  \begin{figure}[H]
    \centering
    \includegraphics[width=1.0\textwidth]{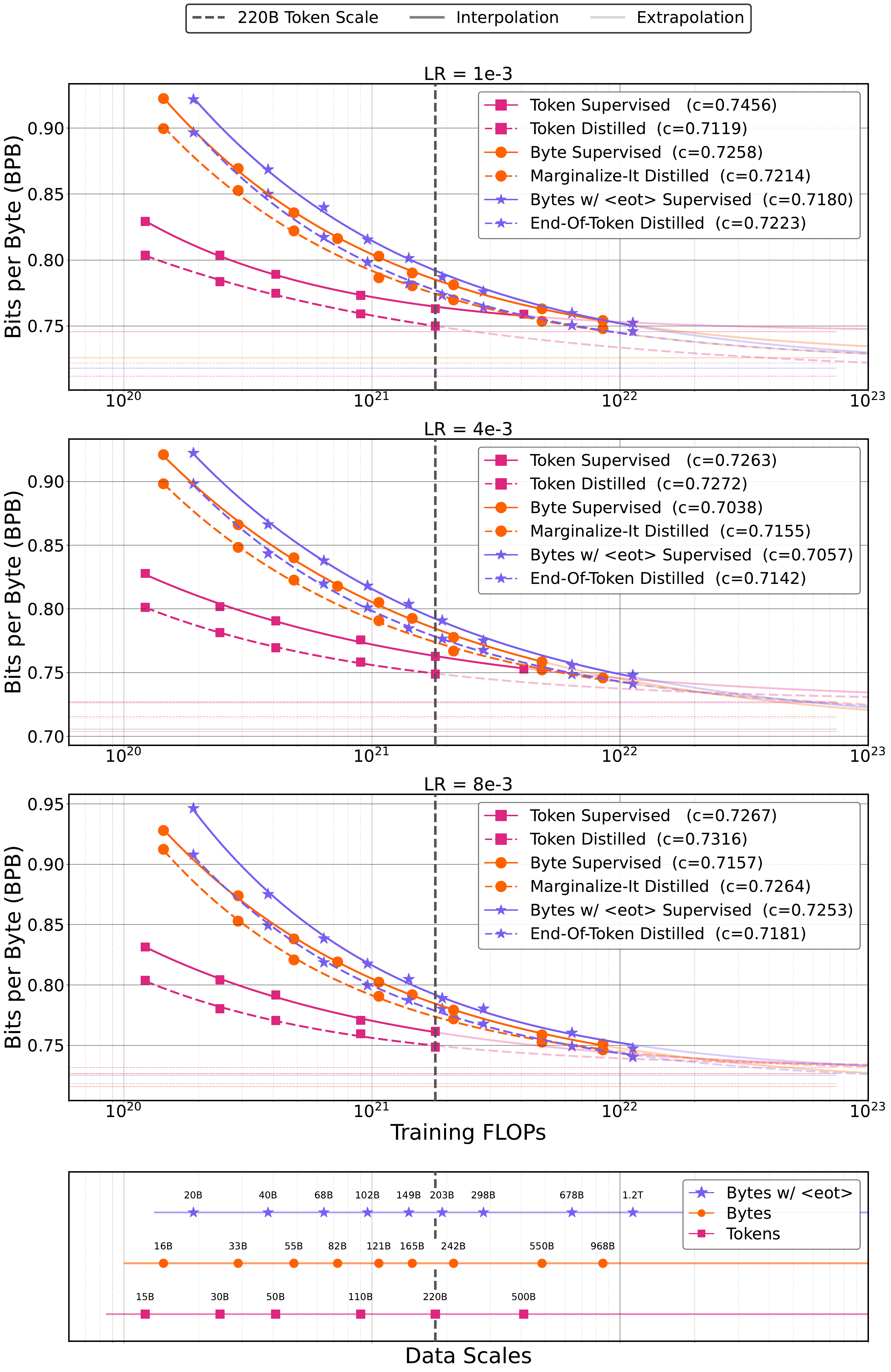}
    \caption{ BPB vs FLOPs for validation dataset 2. }
    \label{fig:news_bpb_vs_flops}

\end{figure}

\section{Additional Results using Validation Dataset 2}

\subsection{BPB vs Training FLOPs power law equations}

\begin{table}[H]
  \centering
  \caption{Fitted BPB-vs-FLOPs power laws of the form $y = b\, x^{a} + c$,
           where $x$ is training FLOPs and $y$ is bits-per-byte on validation dataset 2.}
  \label{tab:bpb-equations}
  \scalebox{1.0}{
  \begin{tabular}{lll}
  \toprule
  LR & Scenario & Equation \\
  \midrule
  \multirow{6}{*}{$ 1 \times 10^{-3} $} & \BPESupervised & $y = 9.993\times 10^{9}\, x^{-0.5514} + 0.7456$ \\
   & \BPEDistilled & $y = 3.251\times 10^{5}\, x^{-0.3261} + 0.7119$ \\
   & \BytesSupervised & $y = 7.753\times 10^{8}\, x^{-0.4759} + 0.7258$ \\
   & \MarginalizeDistilled & $y = 1.036\times 10^{9}\, x^{-0.4842} + 0.7214$ \\
   & \BytesEotSupervised & $y = 3.312\times 10^{8}\, x^{-0.4541} + 0.7180$ \\
   & \EndOfTokenDistilled & $y = 6.752\times 10^{9}\, x^{-0.5218} + 0.7223$ \\
  \midrule
  \multirow{6}{*}{$ 4 \times 10^{-3} $} & \BPESupervised & $y = 3.404\times 10^{6}\, x^{-0.3748} + 0.7263$ \\
   & \BPEDistilled & $y = 7.849\times 10^{7}\, x^{-0.4494} + 0.7272$ \\
   & \BytesSupervised & $y = 1.629\times 10^{7}\, x^{-0.3907} + 0.7038$ \\
   & \MarginalizeDistilled & $y = 2.437\times 10^{8}\, x^{-0.4526} + 0.7155$ \\
   & \BytesEotSupervised & $y = 3.929\times 10^{7}\, x^{-0.4073} + 0.7057$ \\
   & \EndOfTokenDistilled & $y = 5.648\times 10^{8}\, x^{-0.4679} + 0.7142$ \\
  \midrule
  \multirow{6}{*}{$ 8 \times 10^{-3} $} & \BPESupervised & $y = 2.413\times 10^{7}\, x^{-0.4163} + 0.7267$ \\
   & \BPEDistilled & $y = 1.237\times 10^{9}\, x^{-0.5097} + 0.7316$ \\
   & \BytesSupervised & $y = 2.294\times 10^{8}\, x^{-0.4480} + 0.7157$ \\
   & \MarginalizeDistilled & $y = 1.520\times 10^{10}\, x^{-0.5414} + 0.7264$ \\
   & \BytesEotSupervised & $y = 1.336\times 10^{10}\, x^{-0.5318} + 0.7253$ \\
   & \EndOfTokenDistilled & $y = 3.570\times 10^{9}\, x^{-0.5068} + 0.7181$ \\
  \bottomrule
  \end{tabular}}
\end{table}

\subsection{Average top-1 Error Scaling Laws}

\begin{table}[H]
  \centering
  \caption{Fitted downstream error scaling laws \DownstreamEquation for each model variant and learning rate, where $y$ is the validation BPB calibrated using Validation dataset 2 BPB values.}
  \label{tab:scaling_law_eqs}
  \scalebox{0.9}{
  \begin{tabular}{@{}llll@{}}
    \toprule
    \textbf{Model} & \multicolumn{1}{c}{LR$=1\times10^{-3}$} & \multicolumn{1}{c}{LR$=4\times10^{-3}$} & \multicolumn{1}{c}{LR$=8\times10^{-3}$} \\
    \midrule
    \BPESupervised & $0.8410 - 9.8211\,e^{-4.7851\,y}$ & $0.7793 - 55.2335\,e^{-7.3843\,y}$ & $0.7780 - 50.7658\,e^{-7.2916\,y}$ \\
    \BPEDistilled & $0.8419 - 11.1499\,e^{-4.9268\,y}$ & $0.7521 - 193.7935\,e^{-9.2079\,y}$ & $0.8640 - 8.9141\,e^{-4.5266\,y}$ \\
    \midrule
    \BytesSupervised & $0.7635 - 216.7748\,e^{-9.3897\,y}$ & $0.7641 - 103.2172\,e^{-8.4542\,y}$ & $0.7814 - 45.6536\,e^{-7.2404\,y}$ \\
    \MarginalizeDistilled & $0.7709 - 129.1220\,e^{-8.6465\,y}$ & $0.7705 - 95.2408\,e^{-8.2516\,y}$ & $0.7723 - 84.2661\,e^{-8.0843\,y}$ \\
    \midrule
    \BytesEotSupervised & $0.7843 - 47.9190\,e^{-7.2569\,y}$ & $0.7850 - 39.1641\,e^{-6.9986\,y}$ & $0.7729 - 69.5249\,e^{-7.8363\,y}$ \\
    \EndOfTokenDistilled & $0.7726 - 110.6994\,e^{-8.3872\,y}$ & $0.7673 - 133.6625\,e^{-8.6766\,y}$ & $0.7728 - 102.4737\,e^{-8.3038\,y}$ \\
    \bottomrule
  \end{tabular}}
\end{table}

  \begin{figure}[H]
    \centering
    \includegraphics[width=1.0\textwidth]{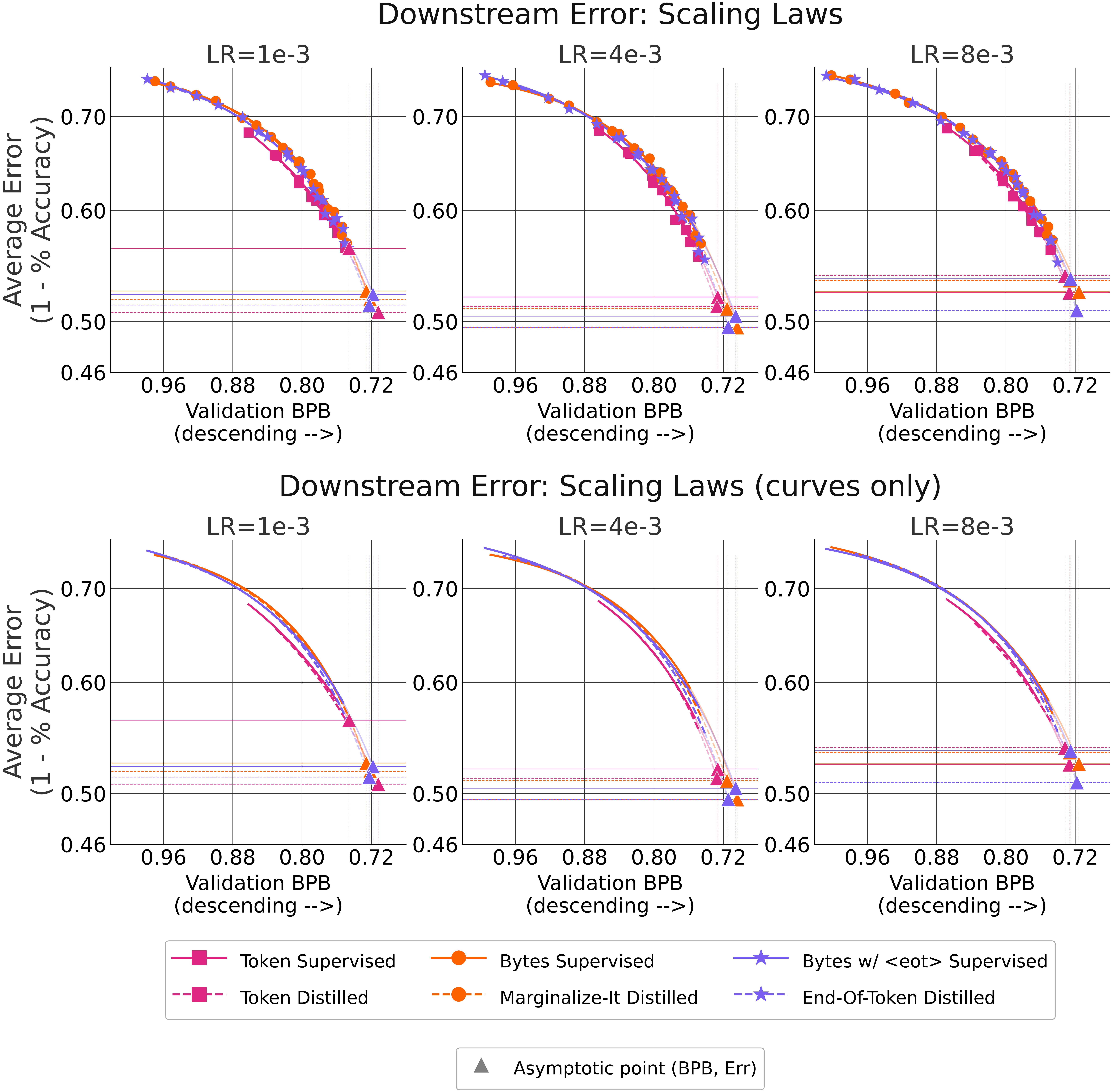}
    \caption{ Downstream error scaling laws calibrated using Validation dataset 2 BPB values. }
    \label{fig:news_downstream_error_scaling_laws}

\end{figure}

\begin{table}[H]
  \centering
  \caption{Asymptotic validation BPB ($\mathrm{BPB}^\star$) and the corresponding downstream average accuracy predicted by the scaling \DownstreamEquation at $y=c$, for each model variant and learning rate calibrated using Validation dataset 2 BPB values.}
  \label{tab:asymptotic}
  \scalebox{1.0}{
  \begin{tabular}{@{}lcccccc@{}}
    \toprule
     & \multicolumn{2}{c}{LR$=1\times10^{-3}$} & \multicolumn{2}{c}{LR$=4\times10^{-3}$} & \multicolumn{2}{c}{LR$=8\times10^{-3}$} \\
    \cmidrule(lr){2-3} \cmidrule(lr){4-5} \cmidrule(lr){6-7}
    \textbf{Model} & c & $a^\star$ & c & $a^\star$ & c & $a^\star$ \\
    \midrule
    \BPESupervised & 0.7456 & 43.6 & 0.7263 & 47.9 & 0.7267 & 47.6 \\
    \BPEDistilled & 0.7119 & 49.2 & 0.7272 & 48.7 & 0.7316 & 46.1 \\
    \midrule
    \BytesSupervised & 0.7258 & 47.4 & 0.7038 & 50.5 & 0.7157 & 47.5 \\
    \MarginalizeDistilled & 0.7214 & 48.1 & 0.7155 & 48.9 & 0.7264 & 46.5 \\
    \midrule
    \BytesEotSupervised & 0.7180 & 47.7 & 0.7057 & 49.6 & 0.7253 & 46.4 \\
    \EndOfTokenDistilled & 0.7223 & 48.6 & 0.7142 & 50.5 & 0.7181 & 49.1 \\
    \bottomrule
  \end{tabular}}
\end{table}

\subsection{\FeatherPlot: Comparisons using validation dataset 2}

  \begin{figure}[H]
    \centering
    \includegraphics[width=1.0\textwidth]{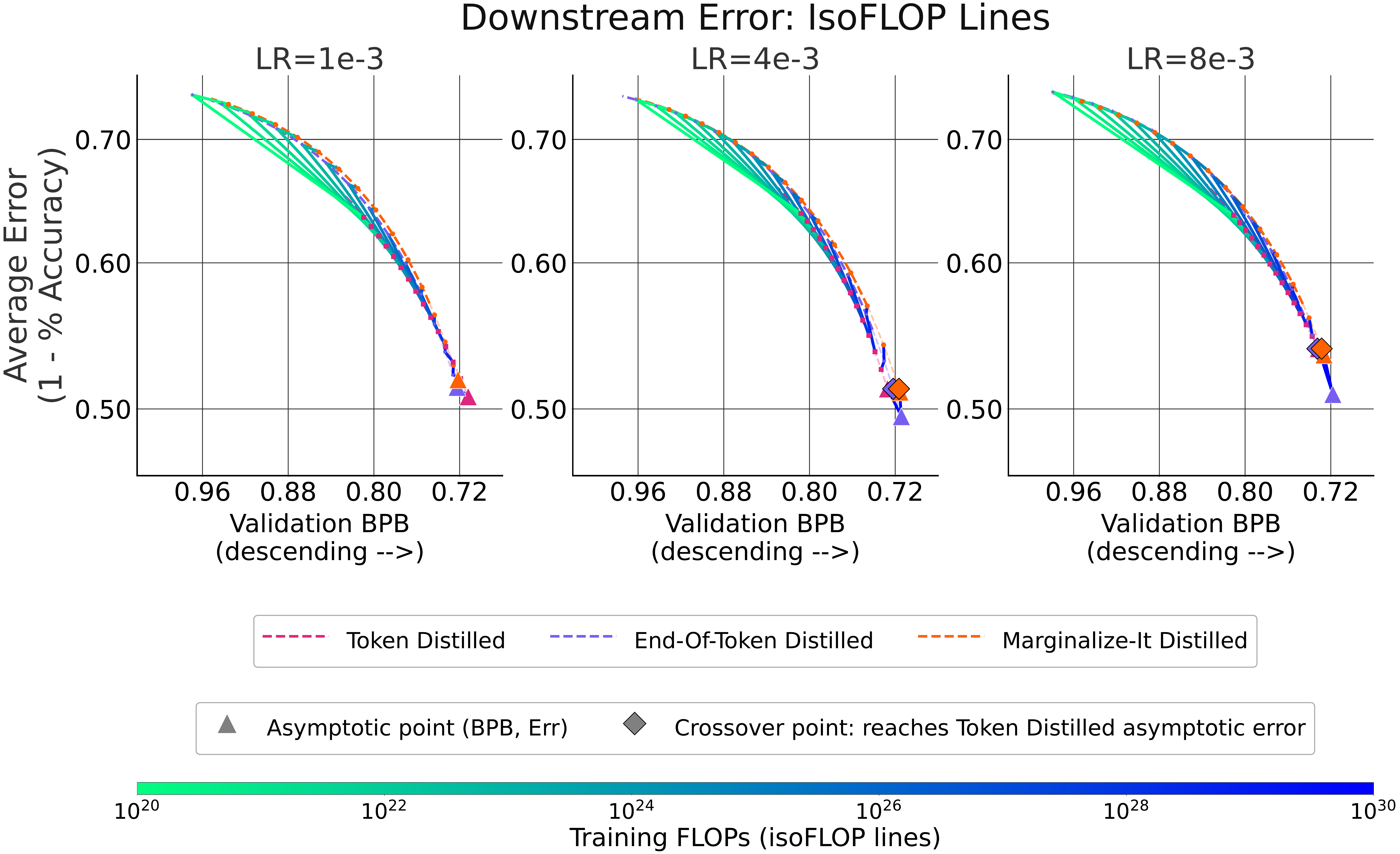}
    \caption{ Feather plots calibrated using Validation dataset 2 BPB values }
    \label{fig:feather_news}

\end{figure}

\subsection{Predicting the asymptotic Average Downstream task performance}

\begin{table}[H]
  \centering
  \caption{Training compute at which \MarginalizeDistilled{} and \EndOfTokenDistilled{} match the asymptotic downstream performance of Token Distilled using Validation Dataset 2. We take Token Distilled\'s asymptotic BPB (the $c$ term of its fitted BPB--FLOPs power law; column \textbf{Token Dist. BPB}) and map it through Token Distilled\'s benchmark scaling law to obtain its asymptotic downstream accuracy (column \textbf{Accuracy}). For each byte method we then invert its own benchmark scaling law at that accuracy to get the BPB it must attain (column \textbf{Method BPB}), and invert its BPB--FLOPs power law at that BPB to get the required training compute (column \textbf{Training FLOPs}). The accuracy column is shared within each learning rate because both methods are evaluated against the same Token Distilled asymptote.}
  \label{tab:crossover_compute}
  \renewcommand{\arraystretch}{1.3}
  \resizebox{\textwidth}{!}{
  \begin{tabular}{cccccc}
    \toprule
    \textbf{LR} & \textbf{Method} & \textbf{Token Dist. BPB} & \textbf{Accuracy (\%)} & \textbf{Method BPB} & \textbf{Method Training FLOPs} \\
    \midrule
    \multirow{2}{*}{$10^{-3}$} & \MarginalizeDistilled & \textemdash & \textemdash & \textemdash & \textemdash \\
     & \EndOfTokenDistilled & \textemdash & \textemdash & \textemdash & \textemdash \\
    \cmidrule{1-6}
    \multirow{2}{*}{$4\times10^{-3}$} & \MarginalizeDistilled & $0.7272$ & $48.7\%$ & $0.7165$ & $1.60 \times 10^{25}$ \\
     & \EndOfTokenDistilled & $0.7272$ & $48.7\%$ & $0.7219$ & $1.69 \times 10^{23}$ \\
    \cmidrule{1-6}
    \multirow{2}{*}{$8\times10^{-3}$} & \MarginalizeDistilled & $0.7316$ & $46.1\%$ & $0.7286$ & $5.17 \times 10^{23}$ \\
     & \EndOfTokenDistilled & $0.7316$ & $46.1\%$ & $0.7326$ & $3.01 \times 10^{22}$ \\
    \bottomrule
  \end{tabular}
  }
\end{table}

\begin{table}[H]
  \centering
  \caption{Effective training-data volume at the compute budget of Table~\ref{tab:crossover_compute}  using Validation Dataset 2. Here $\mathrm{FPU}$ ({FLOPs Per Unit}) is the training FLOPs consumed per sequence unit, where a unit is one token for Token Distilled and one byte for the byte-level methods; it is read directly from each run and used to convert a compute budget into a data count. \textbf{Byte Units} $=\text{FLOPs}/\mathrm{FPU}_{\text{bytes}}$ is the number of byte units the method processes at its crossover compute. \textbf{Bytes} $=\text{Byte Units}\times\text{Multiplier}$ converts byte units into actual UTF-8 bytes of unique text, accounting for the fraction of each unit that is genuine content (Multiplier $=1$ for \MarginalizeDistilled; $=4.5/5.5$ for \EndOfTokenDistilled, whose sequences carry an extra end-of-token symbol per $4.5$ content bytes). \textbf{Tokens} $=\text{FLOPs}/\mathrm{FPU}_{\text{tokens}}$ is the token count Token Distilled would process at the same compute. \textbf{Token to Byte} $=\text{Tokens}\times4.5$ expresses that token count in bytes using the average of $4.5$ bytes per token. \textbf{Data Savings} $=\text{Token to Byte}/\text{Bytes}$ is how much less unique text the byte method needs to reach the same performance. \textbf{Storage Savings} $=(\text{Tokens}\times4.5)/\text{Byte Units}$ compares the token pipeline\'s byte footprint against the raw byte-unit count.}
  \label{tab:crossover_data_volume}
  \renewcommand{\arraystretch}{1.3}
  \resizebox{\textwidth}{!}{
  \begin{tabular}{ccccccccccc}
    \toprule
    \textbf{LR} & \textbf{Method} & \textbf{Training FLOPs} & \textbf{FPU$_{\text{bytes}}$} & \textbf{FPU$_{\text{tokens}}$} & \textbf{Byte Units} & \textbf{Bytes} & \textbf{Tokens} & \textbf{Token to Byte} & \textbf{Data Savings} & \textbf{Storage Savings} \\
    \midrule
    \multirow{2}{*}{$10^{-3}$} & \MarginalizeDistilled & \textemdash & \textemdash & \textemdash & \textemdash & \textemdash & \textemdash & \textemdash & \textemdash & \textemdash \\
     & \EndOfTokenDistilled & \textemdash & \textemdash & \textemdash & \textemdash & \textemdash & \textemdash & \textemdash & \textemdash & \textemdash \\
    \cmidrule{1-11}
    \multirow{2}{*}{$4\times10^{-3}$} & \MarginalizeDistilled & $1.60 \times 10^{25}$ & $8.81 \times 10^{9}$ & $8.18 \times 10^{9}$ & $1.82 \times 10^{15}$ & $1.82 \times 10^{15}$ & $1.96 \times 10^{15}$ & $8.82 \times 10^{15}$ & $4.85\times$ & $4.85\times$ \\
     & \EndOfTokenDistilled & $1.69 \times 10^{23}$ & $9.44 \times 10^{9}$ & $8.18 \times 10^{9}$ & $1.79 \times 10^{13}$ & $1.47 \times 10^{13}$ & $2.07 \times 10^{13}$ & $9.30 \times 10^{13}$ & $6.35\times$ & $5.19\times$ \\
    \cmidrule{1-11}
    \multirow{2}{*}{$8\times10^{-3}$} & \MarginalizeDistilled & $5.17 \times 10^{23}$ & $8.81 \times 10^{9}$ & $8.18 \times 10^{9}$ & $5.87 \times 10^{13}$ & $5.87 \times 10^{13}$ & $6.32 \times 10^{13}$ & $2.84 \times 10^{14}$ & $4.85\times$ & $4.85\times$ \\
     & \EndOfTokenDistilled & $3.01 \times 10^{22}$ & $9.44 \times 10^{9}$ & $8.18 \times 10^{9}$ & $3.19 \times 10^{12}$ & $2.61 \times 10^{12}$ & $3.68 \times 10^{12}$ & $1.66 \times 10^{13}$ & $6.35\times$ & $5.19\times$ \\
    \bottomrule
  \end{tabular}
  }
\end{table}

\newpage
\section{Distillation Methods: Additional Details}
\subsection{\MarginalizeIt and \EndOfToken Pseudocode} \label{dist_pseudocode}

\begin{lstlisting}[
  language=Python,
  basicstyle=\ttfamily\small,
  showstringspaces=false,
  columns=fullflexible,
  keepspaces=true,
  tabsize=4
]
import numpy as np

EOT = 256  # end-of-token symbol (index 256 when use_eot=True)

def byte_probabilities(probs, token_to_bytes, lengths, gold, use_eot=False):
    """Byte distributions P(b | gold[:k], context) for k = 0..len(gold)-1.
    probs (vocab_size):                   token probabilities. 
    token_to_bytes (vocab_size, max_len): per-token bytes, -1 padded. 
    lengths (vocab_size):                 true byte length. 
    gold:                                 byte sequence to condition on. 
    use_eot:                              if True, alphabet is {0..255} U {EOT} (size 257) else 256.
    """
    n_bytes = 257 if use_eot else 256
    live = np.ones(len(probs), dtype=bool)
    dist = []
    for k in range(len(gold)):
        active = live & (lengths > k)
        bytes_at_k = token_to_bytes[:, k]
        accum = np.bincount(bytes_at_k[active], probs[active], minlength=n_bytes)
        total = accum.sum()
        dist.append(accum / total if total > 0 else accum)
        live &= (bytes_at_k == gold[k]) & (lengths > k + 1)
    return dist

\end{lstlisting}

\SetAlCapFnt{\ttfamily}
\SetAlFnt{\ttfamily}
\SetKwComment{tcp}{/* }{ */}

\subsection{Comparing  \MarginalizeIt method with \EndOfToken}\label{corner_cases}
In this section we discuss four special cases of byte logit conversion: \\
 
\textbf{Case I}: Predicting token's first byte distribution using token level teacher distribution (Figure \ref{fig:byte_dist_special_cases_skip_text_page_1}): 

\begin{figure}[H]
  \centering
  \includegraphics[width=1.0\textwidth]{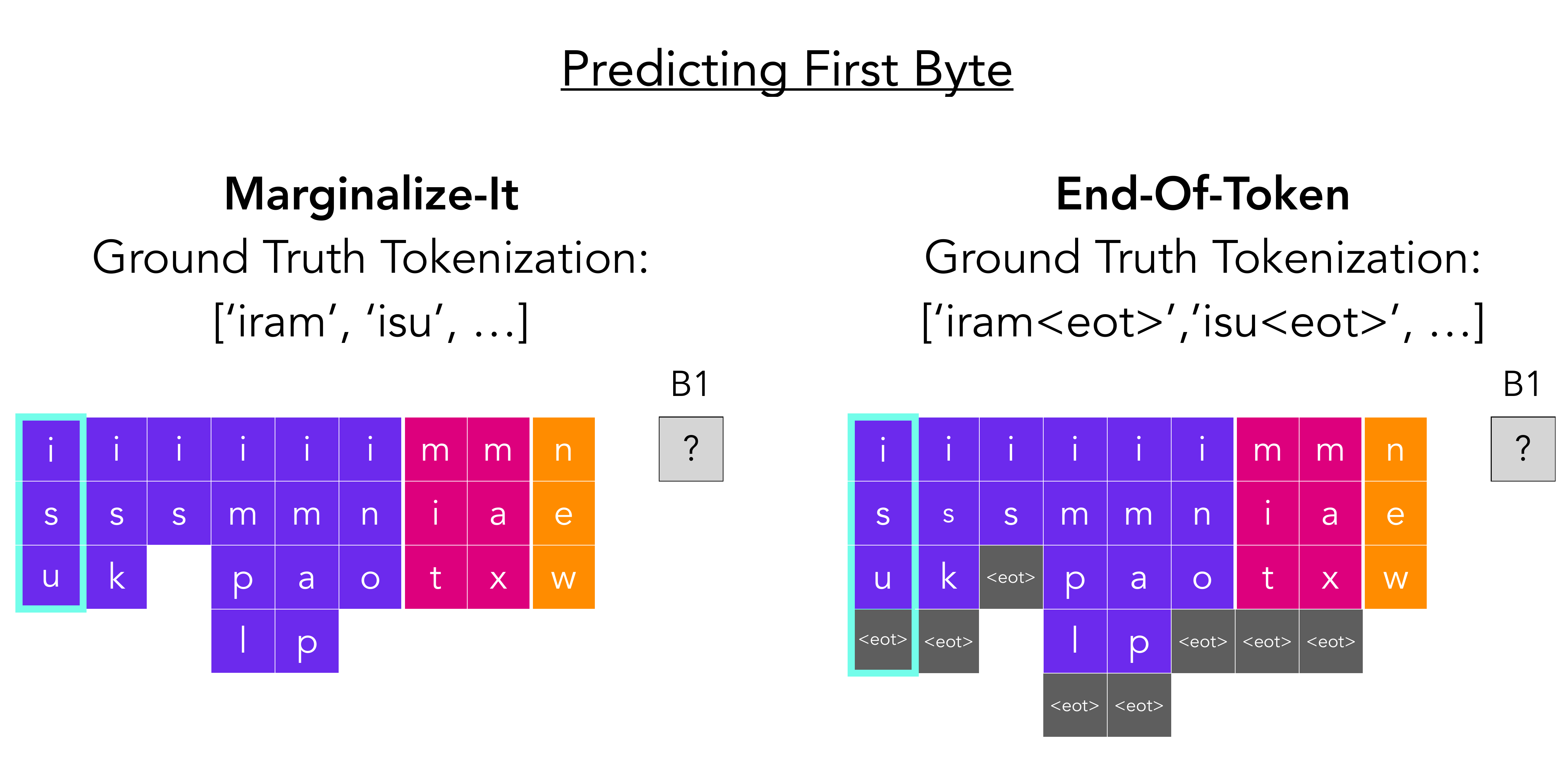}
  \\[2pt]
  \begin{subfigure}{0.49\textwidth}
    \centering
    \caption{B1 is predicted directly by marginalization using next
token distribution of token \texttt{iram} and is exact.}
    \label{fig:a}
  \end{subfigure}%
  \hfill
  \begin{subfigure}{0.49\textwidth}
    \centering
    \caption{B1 is predicted directly by marginalization using next
token distribution of token \texttt{iram<eot>} and is exact.}
    \label{fig:b}
  \end{subfigure}
  \caption{We demonstrate the first byte prediction of the next token. Both \MarginalizeIt and \EndOfToken methods exactly preserve teacher distributions in this case}
  \label{fig:byte_dist_special_cases_skip_text_page_1}
\end{figure}

\noindent\textbf{Case II}: Predicting token's intermediate byte distribution using token level teacher distribution when the ground truth token length matches with all other alternative tokens  with non-zero probability in the vocabulary (Figure \ref{fig:byte_dist_special_cases_skip_text_page_2}): :

\begin{figure}[H]
  \centering
  \includegraphics[width=1.0\textwidth]{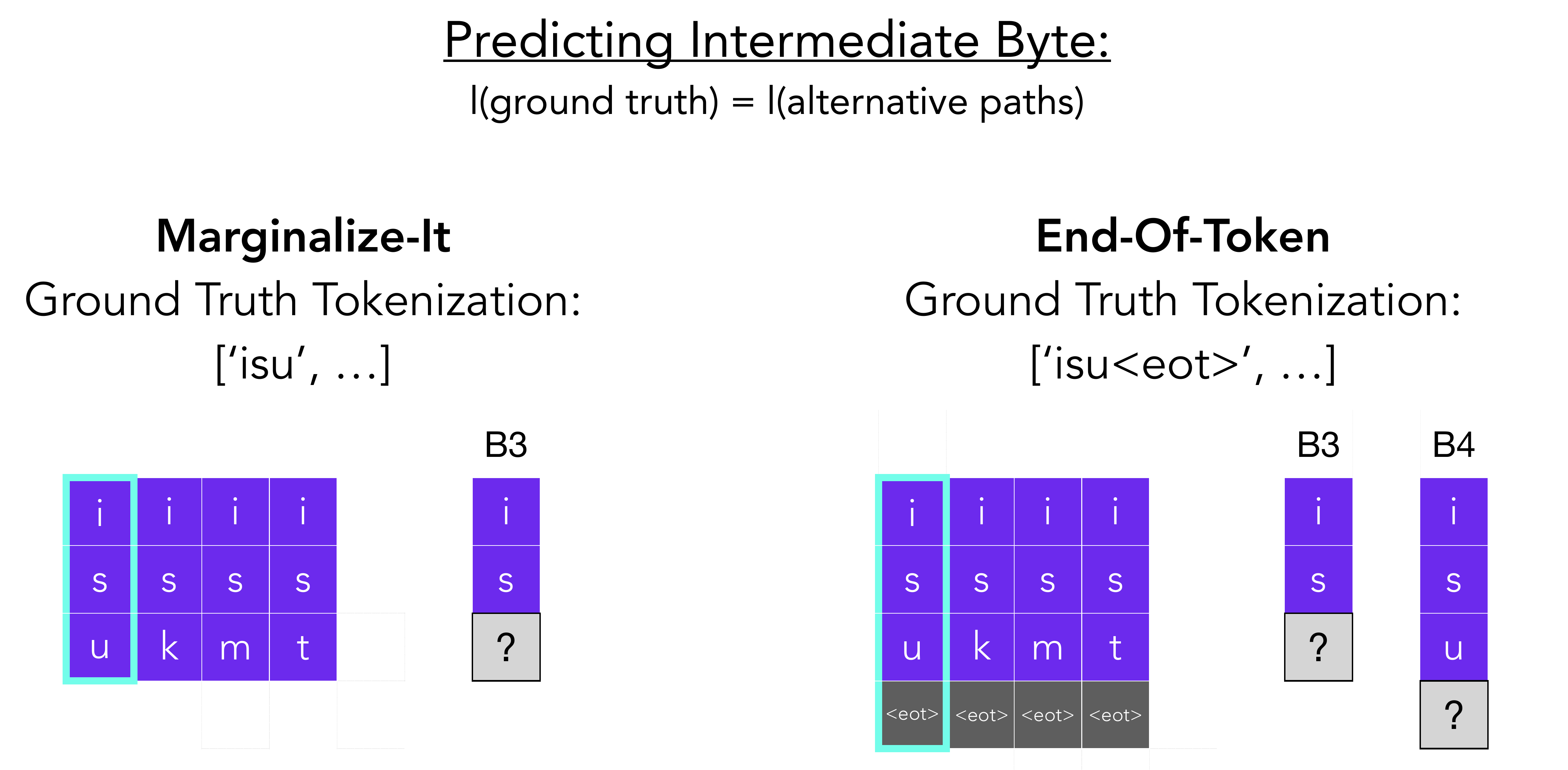}
  \\[2pt]
  \begin{subfigure}{0.49\textwidth}
    \centering
    \caption{B3 prediction is based on tokens with prefix \texttt{is} with a length of at least 3 bytes. Other alternative paths (with non zero probability) have exactly 3 bytes length and can be marginalized exactly. }
    \label{fig:a}
  \end{subfigure}%
  \hfill
  \begin{subfigure}{0.49\textwidth}
    \centering
    \caption{B3 prediction is based on tokens with the prefix \texttt{is} and a length of at least 3 bytes. All alternative paths with non zero probabilities have byte length 4 and can be marginalized
exactly. Similarly, for B4 prediction, all paths have length of 4 and can be
marginalized exactly.}
    \label{fig:b}
  \end{subfigure}
  \caption{We demonstrate the intermediate byte prediction case when all other alternative paths with non zero probability in the vocabulary have exactly the same length as that of the ground truth token. In this case, both \MarginalizeIt and \EndOfToken methods behave equally and give exact byte probabilities. }
  \label{fig:byte_dist_special_cases_skip_text_page_2}
\end{figure}

\begin{figure}[H]
  \centering
  \includegraphics[width=1.0\textwidth]{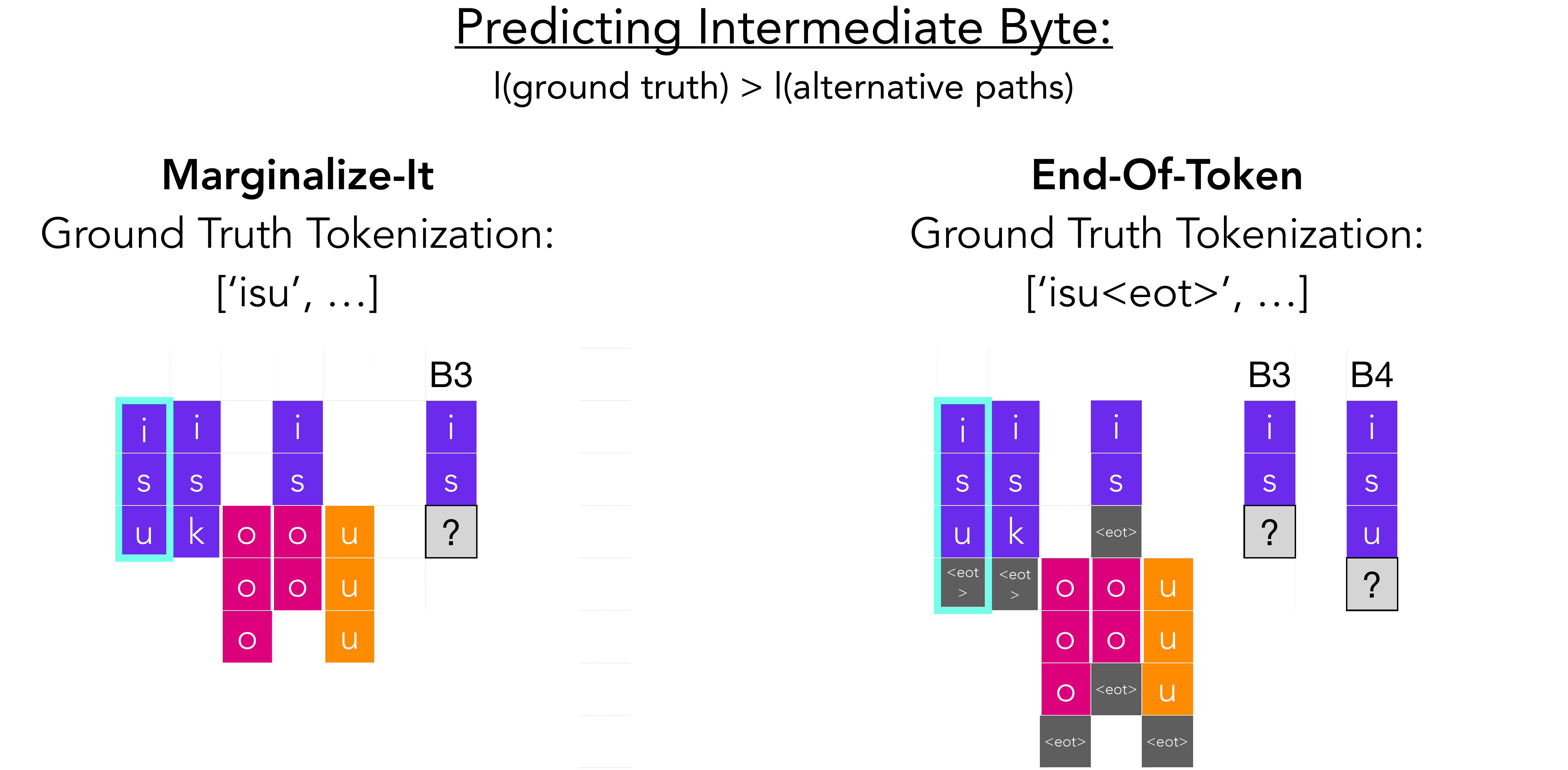}
  \\[2pt]
  \begin{subfigure}{0.49\textwidth}\caption{ B3 prediction is based on tokens with prefix \texttt{is} with a length
of at least 3 bytes. Therefore it misses other tokenization
paths: [\texttt{is}, \texttt{ooo}] , [\texttt{is}, \texttt{oo}], [\texttt{is}, \texttt{uuu}]}\label{fig:a}
\end{subfigure}
  \hfill
  \begin{subfigure}
  {0.49\textwidth}\caption{B3 prediction is based on tokens with the prefix \texttt{is} and a length of at least 3 bytes.
Adding \texttt{<eot>} preserves the distribution, as the byte length of \texttt{is<eot>} is now 3
and no alternative token paths are missed. B4 prediction is conditioned on tokens
starting with prefix \texttt{isu} and with a length of at least 4. Since only the token
\texttt{isu<eot>} satisfies that condition, the teacher distribution is preserved exactly.}\label{fig:b}\end{subfigure}
  \caption{We demonstrate the intermediate byte prediction when there exits other tokens in the vocabulary with shorter byte length with non zero probabilities. Since \MarginalizeIt computes probabilities based on the ground truth token it can miss these alternative paths. \EndOfToken method on the other hand circumvents this problem by adding the new \texttt{<eot>} token. }
  \label{fig:byte_dist_special_cases_skip_text_page_3}
\end{figure}

\noindent\textbf{Case III}: Predicting token’s intermediate byte distribution using token level teacher distribution when there exist other alternative tokens with non zero probabilities in the vocabulary sharing the same prefix but having shorter byte sequence length compared to the ground truth token (Figure \ref{fig:byte_dist_special_cases_skip_text_page_3}):.

\noindent\textbf{Case IV}: Predicting token’s intermediate byte distribution using token level teacher distribution when there exist other alternative tokens with non zero probabilities in the vocabulary sharing the same prefix but having longer byte sequence length compared to the ground truth token. (Figure \ref{fig:byte_dist_special_cases_skip_text_page_4}):  

\begin{figure}[H]
  \centering
  \includegraphics[width=0.99\textwidth]{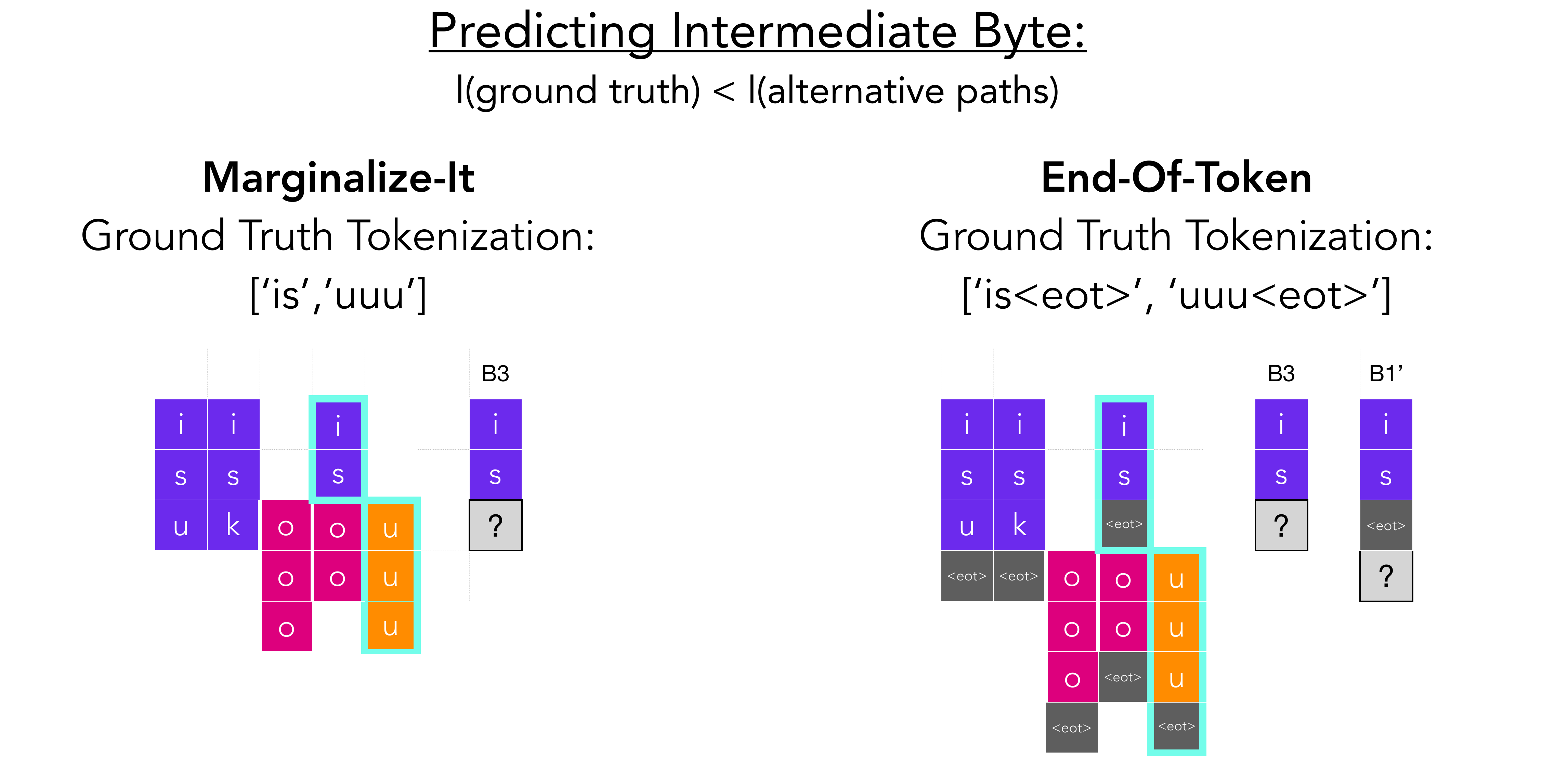}
  \\[2pt]
  \begin{subfigure}{0.49\textwidth}\caption{B3 prediction is based on the next token distribution of \texttt{is} and therefore it misses contributions from other
tokenization paths formed by tokens \texttt{isu} and \texttt{isk}.}\label{fig:a}\end{subfigure}%
\hfill
  \begin{subfigure}{0.49\textwidth}\caption{ B3 prediction is based on tokens with prefix \texttt{is} and has a length of at least 3
bytes. Adding \texttt{<eot>} preserves the distribution as now byte length of
\texttt{is<eot>} is 3 and no alternative token paths are missed. B4 prediction is
based on the next token distribution of \texttt{is<eot>} and preserves exact
distribution as no other token paths have prefix \texttt{is<eot>}.}\label{fig:b}
  \end{subfigure}
\caption{We demonstrate intermediate byte prediction when there exist other tokens in the vocabulary with longer byte lengths and non-zero probabilities. Since \MarginalizeIt computes probabilities based on the ground-truth token, it can miss the contributions from these alternative paths. The \EndOfToken method, on the other hand, circumvents this problem by adding the new \texttt{<eot>} token.}
  \label{fig:byte_dist_special_cases_skip_text_page_4}
\end{figure}

\section{On BPB vs Benchmark Performance Discrepancy}
\subsection{Why the same BPB may imply different downstream performance?}\label{different_downstream}

 BPB is a function of only {one number per position}: the probability the model
places on the target token, $p_{\text{correct}}$. The per-token loss is
$-\log_2 p_{\text{correct}}$, so once $p_{\text{correct}}$ is fixed the loss is
fixed---BPB is completely blind to how the remaining mass
$1-p_{\text{correct}}$ is arranged among the other vocabulary items. Downstream accuracy, in contrast, depend on whether the target is
the $\arg\max$, i.e.\ on the {ranking} of the full distribution. Since the
ranking is governed by the residual mass that BPB ignores, two models can share
an identical BPB while decoding completely different text. \\

\noindent \texttt{Example:} Concretely, take the target ``\textcolor{myblue}{\texttt{Where's my tiramisu?}}'', which the
Llama3-8b tokenizer segments into $7$ tokens spanning $20$ bytes:
\begin{center}
\texttt{Where} \textbar{} \texttt{'s} \textbar{}
\texttt{\textvisiblespace my} \textbar{} \texttt{\textvisiblespace tir}
\textbar{} \texttt{am} \textbar{} \texttt{isu} \textbar{} \texttt{?}
\end{center}
where \texttt{\textvisiblespace} denotes a leading space. Suppose both models
place probability $p_{\text{correct}}=0.4$ on the correct token at {every}
position. Then each per-token loss is $-\log_2 0.4 = 1.32$ bits, the total is
$7 \times 1.32 = 9.24$ bits, and
\begin{equation}
\text{BPB} \;=\; \frac{9.24}{20} \;=\; 0.462
\end{equation}
for {both} models. The models differ only in where they place the
remaining $0.6$ of mass---which BPB never observes:

\begin{table}
\small
\caption{An illustration of how the relative ranking of tokens in the vocabulary
can lead to worse generation performance while maintaining the same BPB value.}
\begin{tabular}{clccclc}
\toprule
Pos & Target & $p_{\text{correct}}$ & \multicolumn{2}{c}{Model A} & \multicolumn{2}{c}{Model B} \\
\cmidrule(lr){4-5}\cmidrule(lr){6-7}
    &        &        & distractors & $\arg\max$ & distractors & $\arg\max$ \\
\midrule
1 & \texttt{Where}            & $0.4$ & \texttt{There}$=0.35$, \texttt{When}$=0.25$   & \textbf{\texttt{Where}} \ding{51} & \texttt{There}$=0.5$, \texttt{When}$=0.1$   & \texttt{There} \ding{55} \\
2 & \texttt{'s}               & $0.4$ & \texttt{'re}$=0.3$, \texttt{'d}$=0.3$          & \textbf{\texttt{'s}} \ding{51}    & \texttt{'d}$=0.5$, \texttt{'re}$=0.1$        & \texttt{'d} \ding{55} \\
3 & \texttt{\textvisiblespace my}  & $0.4$ & \texttt{\textvisiblespace your}$=0.35$, \texttt{\textvisiblespace the}$=0.25$ & \textbf{\texttt{\textvisiblespace my}} \ding{51} & \texttt{\textvisiblespace be}$=0.5$, \texttt{\textvisiblespace the}$=0.1$ & \texttt{\textvisiblespace be} \ding{55} \\
4 & \texttt{\textvisiblespace tir} & $0.4$ & \texttt{\textvisiblespace bur}$=0.35$, \texttt{\textvisiblespace piz}$=0.25$ & \textbf{\texttt{\textvisiblespace tir}} \ding{51} & \texttt{\textvisiblespace cal}$=0.5$, \texttt{\textvisiblespace bur}$=0.1$ & \texttt{\textvisiblespace cal} \ding{55} \\
5 & \texttt{am}               & $0.4$ & \texttt{av}$=0.3$, \texttt{ig}$=0.3$          & \textbf{\texttt{am}} \ding{51}    & \texttt{z}$=0.5$, \texttt{av}$=0.1$          & \texttt{z} \ding{55} \\
6 & \texttt{isu}              & $0.4$ & \texttt{ami}$=0.35$, \texttt{osa}$=0.25$      & \textbf{\texttt{isu}} \ding{51}   & \texttt{one}$=0.5$, \texttt{osa}$=0.1$       & \texttt{one} \ding{55} \\
7 & \texttt{?}                & $0.4$ & \texttt{.}$=0.35$, \texttt{!}$=0.25$          & \textbf{\texttt{?}} \ding{51}     & \texttt{!}$=0.5$, \texttt{.}$=0.1$           & \texttt{!} \ding{55} \\
\bottomrule
\end{tabular}
\end{table}

\noindent In Model~A the target is rank~1 at every step, so greedy decoding produces ``\texttt{Where's my tiramisu?}''---a $7/7$ exact match. In Model~B a single
distractor sits at $0.5$, just above the target's $0.4$, so the target is rank~2
at every step and the concatenated $\arg\max$ tokens spell
``\texttt{There'd be calzone!}''---$0/7$. The loss never observes the $0.5$
distractor greater the $0.4$ target; it considers only the $0.4$. Both models
therefore have an identical BPB of $0.463$ while achieving $100\%$ versus
$0\%$ token accuracy.Equal BPB constrains the probability of the target but does not guarantee a rank 1.

\bibliographystyle{johd}
\bibliography{bib} 

\end{document}